\documentclass[lettersize,journal]{IEEEtran}
\usepackage{amsmath,amsfonts}
\usepackage{array}
\usepackage{textcomp}
\usepackage{stfloats}
\usepackage{url}
\usepackage{verbatim}
\usepackage{graphicx}
\usepackage{cite}
\usepackage{hyperref}
\usepackage{multirow}
\usepackage{cleveref}
\usepackage{threeparttable}
\usepackage{xcolor}
\usepackage{amsmath}
\usepackage{tabularx}
\usepackage{booktabs}
\usepackage{float}
\usepackage{subfigure}
\usepackage{subcaption}
\usepackage[linesnumbered,ruled,vlined]{algorithm2e}
\usepackage{amsmath}
\usepackage{amssymb}
\usepackage{tabularx}
\usepackage{wrapfig}
\usepackage{tikz}
\usepackage{tabularx}
\usepackage{adjustbox}

\hypersetup{
    colorlinks=true, 
    linkcolor=blue,  
    filecolor=blue,  
    urlcolor=black,   
    citecolor=blue,  
    pdfborder={0 0 0} 
}

\Crefname{figure}{Fig.}{Figs.}
\Crefname{equation}{Eq.}{Eqs.}

\def\eg{e.g.}

\def\ie{{i.e.}}

\newcommand{\rev}[1]{\textcolor{black}{{#1}}}

\begin{document}

\title{
Generalizable Autonomous Driving System across Diverse Adverse Weather Conditions
}

\author{~~Wei-Bin Kou, Guangxu Zhu, Rongguang Ye, Qingfeng Lin, \\ Zeyi Ren, Ming Tang*, Yik-Chung Wu*
\thanks{Wei-Bin Kou, Qingfeng Lin, Zeyi Ren and Yik-Chung Wu are with the Department of Electrical and Electronic Engineering, The University of Hong Kong, Hong Kong, China.}
\thanks{Ming Tang and Rongguang Ye are with the Department of Computer Science and Engineering, Southern University of Science and Technology, Shenzhen, China.}
\thanks{Guangxu Zhu is with Shenzhen Research Institute of Big Data, Shenzhen, China.}
\thanks{\textit{(Corresponding author: Ming Tang and Yik-Chung Wu.)}}
}

\markboth{IEEE TRANSACTIONS ON INTELLIGENT TRANSPORTATION SYSTEMS,~Vol.~xx, No.~x, August~2024}%
{Shell \MakeLowercase{\textit{et al.}}: A Sample Article Using IEEEtran.cls for IEEE Journals}

\maketitle

\begin{abstract}
Various adverse weather conditions pose a significant challenge to autonomous driving (AD) street scene semantic understanding (segmentation). A common strategy is to minimize the disparity between images captured in clear and adverse weather conditions. However, this technique typically relies on utilizing clear image as a reference, which is challenging to obtain in practice. Furthermore, this method typically targets a single adverse condition, and thus perform poorly when confronting \rev{a mixture} of multiple adverse weather conditions. To address these issues, we introduce a reference-free and \underline{Adv}erse weather-\underline{Immu}ne scheme (called AdvImmu) \rev{that leverages} the invariance of weather conditions over short periods (seconds). Specifically, AdvImmu includes three components: Locally Sequential Mechanism (LSM), Globally Shuffled Mechanism (GSM), and Unfolded Regularizers (URs). LSM leverages temporal correlations between adjacent frames to enhance model performance. GSM is proposed to shuffle LSM segments to prevent overfitting \rev{of} temporal patterns. URs are the deep unfolding implementation of two proposed regularizers to penalize the model complexity to enhance across-weather generalization. In addition, to overcome the over-reliance on consecutive frame-wise annotations in the training of AdvImmu (typically unavailable in AD scenarios), we incorporate \rev{a foundation model named} Segment Anything Model (SAM) to assist to annotate frames, and additionally propose a cluster algorithm (denoted as SBICAC) to surmount SAM's category-agnostic issue to generate pseudo-labels. Extensive experiments demonstrate that the proposed AdvImmu outperforms existing state-of-the-art methods by 88.56\% in mean Intersection over Union (mIoU).
\end{abstract}


\section{Introduction}
\label{sec1}
Autonomous driving (AD) perception is critical \cite{10461035,natan2022towards,muhammad2022vision} for AD vehicles. In general, AD perception \rev{consists of} a set of tasks, such as semantic segmentation \cite{kou2024fast,9913352}, object detection \cite{song2024robustness}, object tracking \cite{karle2023multi}, etc. This work stays focused on semantic segmentation which aims to classify each pixel into predefined categories. Currently, semantic segmentation models are typically built using Deep Learning (DL) techniques \cite{xiao2020multimodal,10337777,10494721,10323218,10372140}, particularly Convolutional Neural Networks (CNNs) \cite{yu2021bisenet,10414408} and Transformers \cite{hoyer2023domain}. These models are generally trained on large datasets with labels that are manually annotated at the pixel level. Once trained, these models can generally predict the category of each pixel of given images properly. 

\begin{figure}[tp]
    \centering
    \includegraphics[width=\linewidth, height=0.56\linewidth]{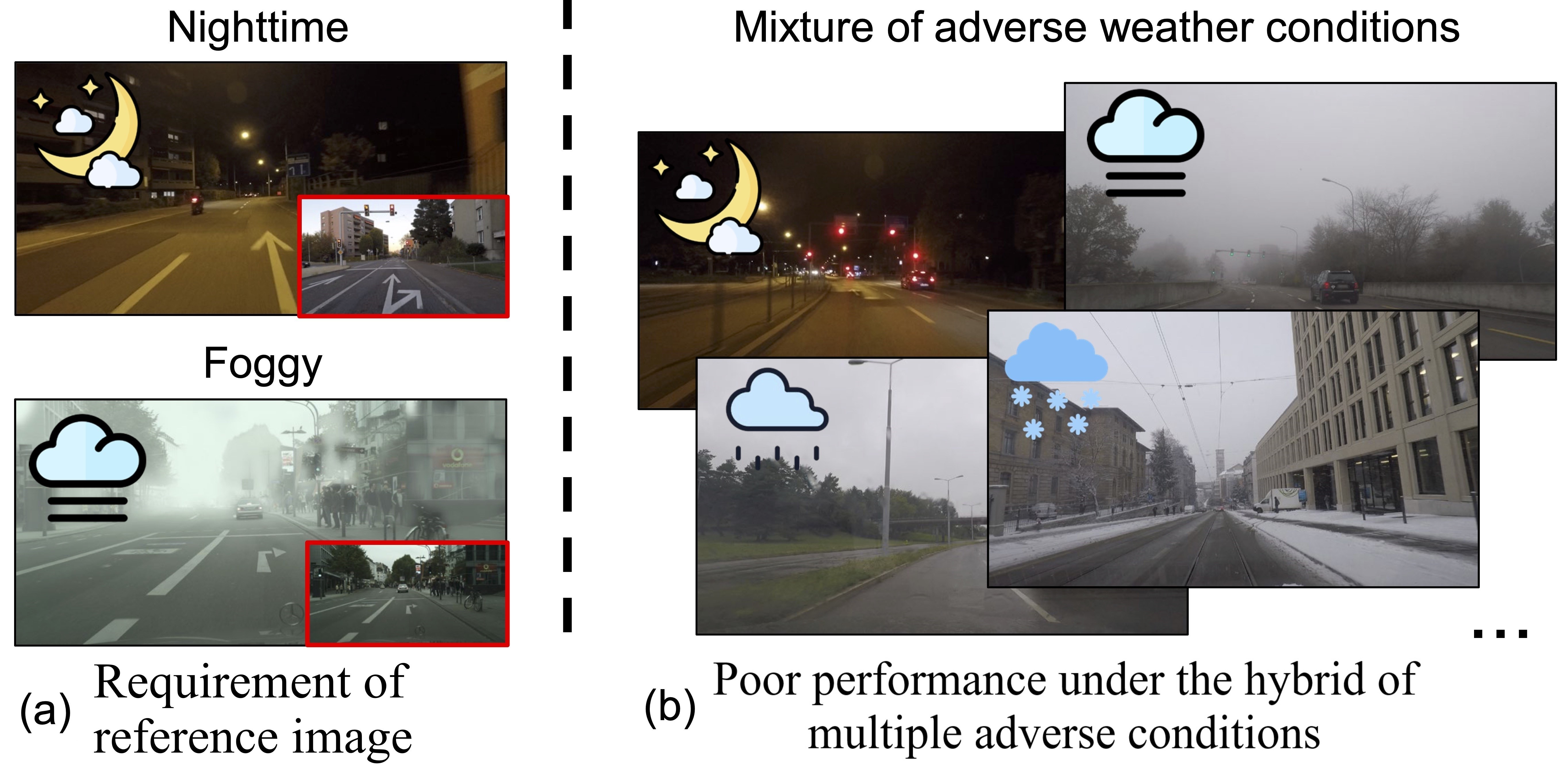}
    \caption{{Major challenges in the realm of domain adaption \cite{li2023vblc}.} (a) Requirement of reference images. Reference images, highlighted in the red box located at the bottom right corner, depict clear-weather-condition scenes which are almost same with those under adverse weather conditions (e.g., nighttime or fog). Generally, it is difficult to collect such reference images in practice. (b) Poor performance under the hybrid of multiple adverse conditions. In general, domain adaption models always work well in adapted target domains and face significant performance decline in unadapted domains, which results in poor performance when dealing with the case of hybrid of multiple adverse weather conditions.}
    \label{Fig:motivation}
    \vspace{-0.5cm}
\end{figure}

Although such DL-based models achieve advanced performance for semantic segmentation task, they generally face  significant performance drop in various adverse weather conditions \cite{10341550}. For example, a semantic segmentation model trained on a dataset predominantly composed of clear, sunny weather conditions, will \rev{decay} significantly when the vehicle encounters a foggy environment. This is because the fog causes a reduction in visibility and alters the appearance of objects, leading to incorrect prediction of each pixel's category. Existing domain adaption approach to address this challenge involves reducing the disparity between images captured in clear weather as a reference and those taken in adverse conditions \cite{li2023vblc}. However, obtaining such a reference image can be challenging in reality. Furthermore, these methods are generally developed to handle specific single adverse condition, such as foggy weather. This specialization often leads to a significant decline in performance when vehicles encounter other adverse weather conditions where the model has not been specifically tuned. \rev{The mentioned} challenges are summarized in \Cref{Fig:motivation}.

\begin{figure*}[tp]
\hspace{-0.4cm}
\includegraphics[width=1.03\linewidth, height=0.65\linewidth]{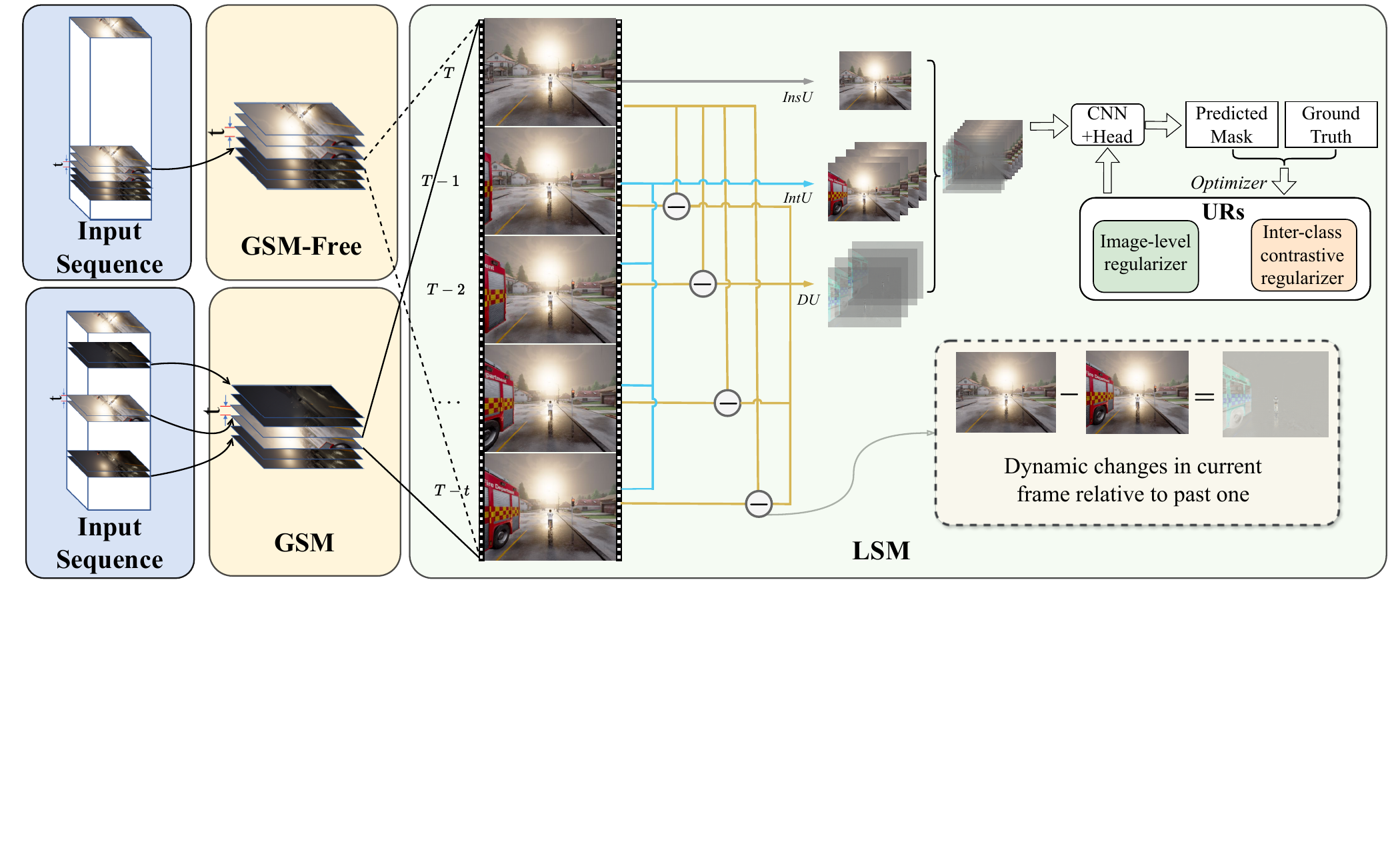}
\vspace{-4.1cm}
\caption{Overview of the proposed AdvImmu. In the context of LSM, Instant Unit (InsU), Integral Unit (IntU), and Derivative Unit (DU) capture instantaneous information, stable and shared background information, and dynamic changes from the input, respectively. These three units work together to enhance model performance by leveraging temporal correlation of intra-weather condition. Notably, $T$ denotes the current frame number in the input sequence, while $t$ (\ie, LSM depth) refers to the count of consecutive frames prior to frame $T$. GSM shuffles and groups the LSM segments to avoid the overfitting to specific temporal patterns, whereas GSM-Free does not. GSM is proposed to enable the model to learn common patterns across multiple adverse weather conditions. URs include two unfolded regularizers and are proposed to penalize the model complexity to avoid overfitting to specific patterns as well, therefore, enhancing the model performance across multiple adverse weather conditions.}
\vspace{-0.3cm}
\label{Fig.LSM_with_EF}
\end{figure*}

To effectively tackle these challenges, we propose a novel scheme (named AdvImmu). AdvImmu benefits from being independent of reference images and is robust against \rev{a mixture} of various adverse weather conditions. \rev{The proposed method} is materialized by considering the local temporal correlation, global randomness, and unfolded regularization. \rev{Specifically}, AdvImmu consists of Locally Sequential Mechanism (LSM), Globally Shuffled Mechanism (GSM), and Unfolded Regularizers (URs) modules. 

\rev{In particular}, LSM exploits the temporal correlations between consecutive frames via three distinct units: Instant Unit (InsU), which is designed to detect instantaneous information in the input; Integral Unit (IntU), which focuses on capturing stable and shared background information from the scenes; Derivative Unit (DU), which excels in sensing dynamic changes in the environment. \rev{However}, since LSM integrates temporal patterns from historical consecutive frames, it is susceptible to overfitting specific input sequence order. To mitigate LSM's temporality-related overfitting, GSM is introduced to rearrange the segments processed by the LSM. This strategic shuffling ensures that the model does not merely memorize a particular data order but instead learns a generalizable pattern across \rev{various} data sequences. On top of LSM and GSM, URs are proposed to improve the model's generalization across diverse adverse weather conditions by penalizing the model complexity. Concretely, in addition to the conventional pixel-level cross entropy loss used in semantic understanding, we propose to use the deep unfolding technique \cite{10529194} to unfold an image-level regularizer and an inter-class contrastive regularizer into layers. Deep unfolding can avoid heuristic or exhaustive searches on the weights of both regularizers, which prevents instability and suboptimal performance. The proposed AdvImmu is summarized in \Cref{Fig.LSM_with_EF}.

\begin{figure}[tp]
\vspace{-0.1cm}
\hspace{-0.8cm}
\subfigure[Raw image]{
\label{Fig.row_image}
\includegraphics[width=0.55\linewidth]{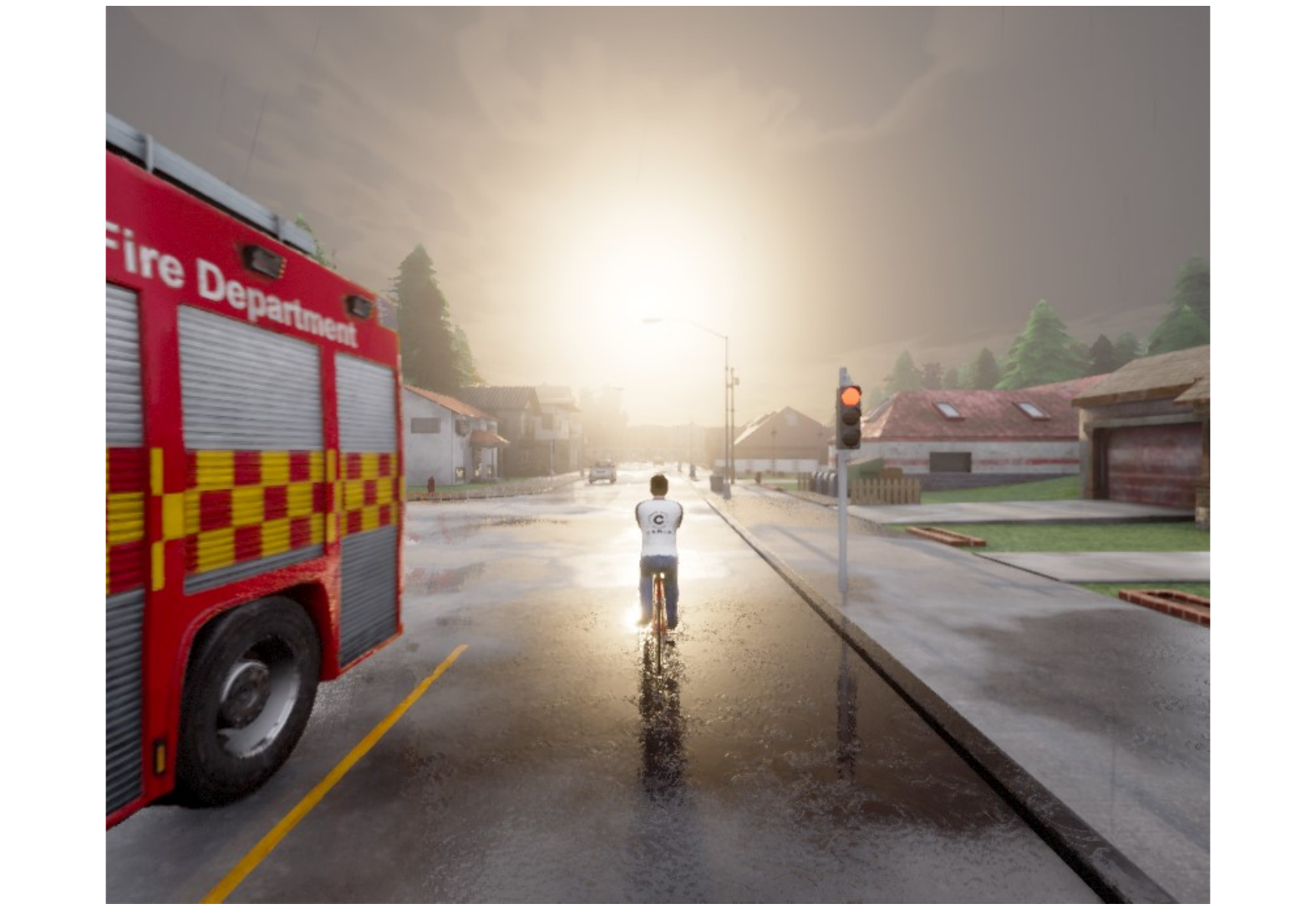}}
\hspace{-0.9cm}
\subfigure[SAM-generated mask]{
\label{Fig.SAM_generated_mask}
\includegraphics[width=0.55\linewidth]{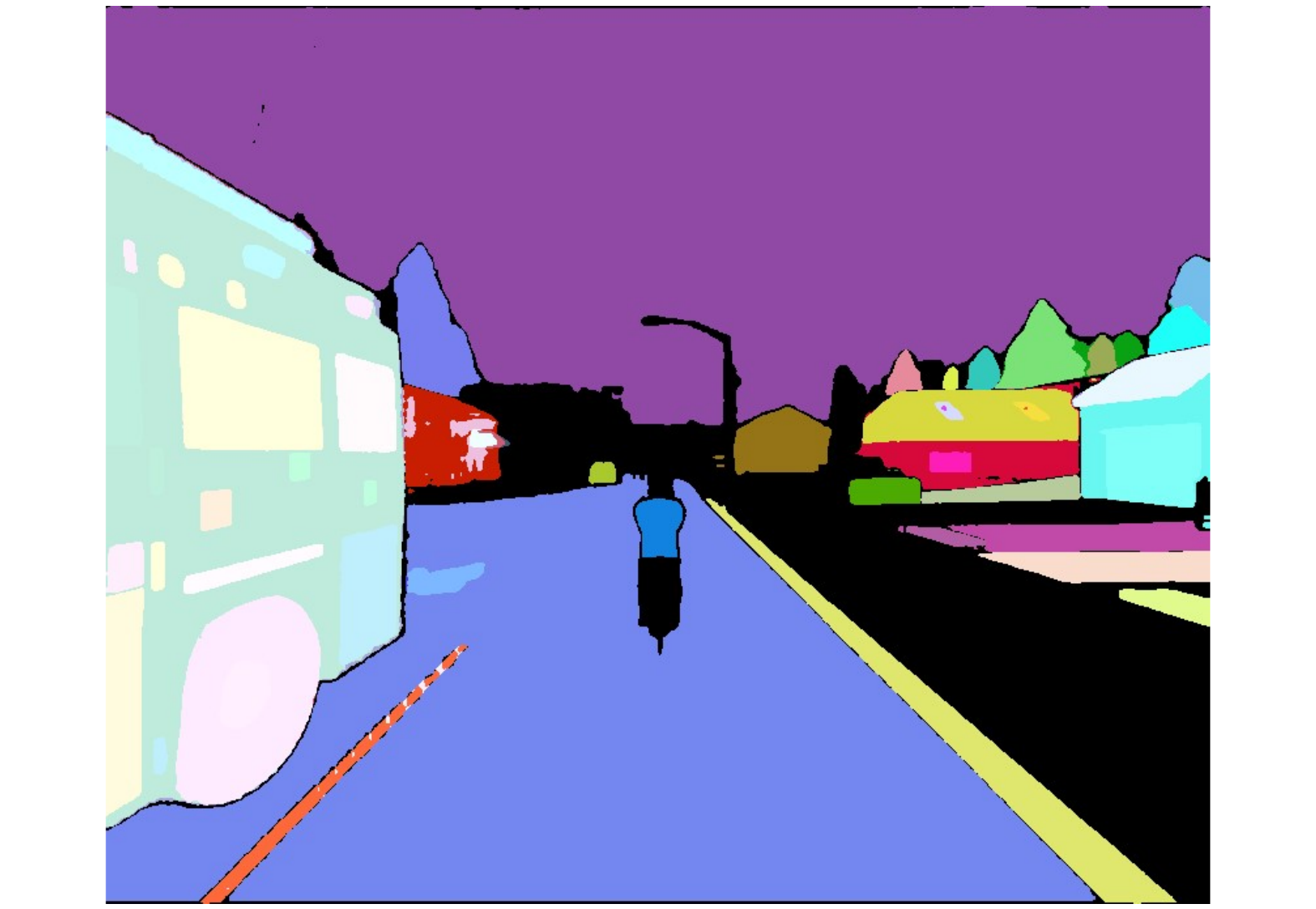}}
\vspace{-0.2cm}
\caption{Illustration of the category-agnostic issue in SAM-generated mask. For example, trees are assigned to different semantic IDs (colors) in SAM-generated mask.}
\label{Fig.SAM_reason_to_cluster}
\vspace{-0.5cm}
\end{figure}

\rev{In general}, the training of the proposed AdvImmu requires consecutive frame-wise annotations for all images in the training sequences. \rev{As it} is impractical to annotate manually in the realm of AD scenarios, we propose to use \rev{a foundation model named} Segment Anything Model (SAM) \cite{kirillov2023segment} to assist \rev{annotating} images in the training sequences, thanks to SAM's powerful capabilities to segment objects \cite{xiong2024efficientsam}. However, SAM-generated masks lack category information (demonstrated in \Cref{Fig.SAM_reason_to_cluster}). To overcome this category-agnostic issue, \cite{jiang2023segment} propose to use prompts to assist SAM to generate precise class masks. This method is difficult to automatically process batches of images owing to the requirement of articulated prompts. \cite{chen2023segment} utilizes Class Activation Maps (CAM) pseudo-labels as cues to select and combine SAM masks to generate class-aware pseudo-labels, which, however, needs extra CAM-related classification model and post-processing techniques. Therefore, this method is also impractical for generating consecutive frame-wise annotations.

In this work, we propose \rev{a novel} clustering algorithm to supplement SAM to generate category-aware pseudo-labels. Specifically, we propose a clustering algorithm named Similarity-based Iterative Cluster Assignments and Centroids Refinement (SBICAC) to overcome SAM's category-agnostic issue and further assign unique ID to each semantic class as frame-wise labels. SBICAC performs clustering on SAM-extracted latent feature maps. The process begins by initializing cluster centroids randomly. It then iterates through a predefined maximum number of iterations. In each iteration, the proposed SBICAC: \textbf{(I) Computes Similarity:} It calculates the similarity between each pixel's latent feature and all centroids using a similarity metric (\eg, inner product or cosine similarity). This operation results in a similarity matrix. \textbf{(II) Assigns Clusters:} Each pixel is assigned to the cluster with the highest similarity, thus updating the labels for each pixel based on maximum similarity. \textbf{(III) Updates Centroids:} For each cluster, the new centroid is recalculated as the mean features of all pixels assigned to that cluster. \textbf{(IV) Checks for Convergence:} The loop checks if the centroids have stabilized (i.e., the old centroids and new centroids are sufficiently close to each other). If they have converged, the loop breaks, ending the iterations early. The output of the algorithm is the cluster assignments (labels) for all pixels. In summary, SBICAC iteratively refines pixels' cluster assignments and centroids based on similarity until the convergence is achieved. Based on the high-quality pseudo-labels generated by SBICAC and SAM, we can train the proposed AdvImmu in an unsupervised manner.

The main contributions \rev{of this paper} are highlighted as follows:
\begin{itemize}
    \item This work proposes AdvImmu that is independent of reference images and resilient to various adverse weather conditions. It considers both local temporal correlations to enhance model performance and global randomness to prevent overfitting to specific temporal patterns.
    \item AdvImmu also includes an image-level regularizer and an inter-class contrastive regularizer to enhance the model's across-weather generalization, and additionally unfolds both regularizers into layers by deep unfolding to avoid instability and suboptimal performance.
    \item The training of AdvImmu needs to overcome SAM's category-agnostic issue in the output. To this end, an unsupervised clustering algorithm (\ie, SBICAC) is proposed to generate category IDs serving as training labels based on SAM's output. 
    \item Extensive experiments demonstrate that the proposed AdvImmu can generalize well under various adverse weather conditions. It outperforms existing state-of-the-art (SOTA) method by 88.56\% in mean Intersection over Union (mIoU). 
\end{itemize}

The remainder of this paper is organized as follows. \Cref{related_work} provides an overview of the related work\rev{s}. \Cref{methodology} elaborates on the proposed methodologies. \Cref{experiments} presents a comprehensive set of experiments along with empirical analyses. \Cref{conclusion} concludes this paper.

\section{Related Work}
\label{related_work}
\subsection{Domain Adaption for Autonomous Driving Perception under Adverse Weather Conditions}
AD street scene semantic understanding under various adverse conditions has recently attracted increasing attention due to the strict safety demand in various outdoor perception \cite{9811870,wu2024hierarchical} and navigation \cite{10342134} tasks. Domain adaption \cite{9981603} is the major scheme in enhancing AD capabilities of street scene semantic understanding under various adverse conditions. Studies in domain adaption \cite{10342102} aim to learn a generalizable model from single or multiple related but distinct source domains where target data is inaccessible during model learning. Most domain adaption approaches can be grouped into adversarial-based domain adaption, curriculum learning-based domain adaption, and self-training-based domain adaption. 

\rev{Specifically}, adversarial training-based methods for domain adaptation aim to minimize the domain discrepancy through style transfer or by learning representations that are indistinguishable across domains \cite{kim2020learning}. Typically, these methods work within a GAN framework \cite{goodfellow2014generative} to align the distributions of the source and target domains at various levels: input \cite{gong2021dlow}, feature \cite{tsai2018learning}, or output \cite{vu2019advent}. \rev{On the other hand}, curriculum learning-based domain adaptation achieves gradual adaptation to a more distant target domain \cite{sakaridis2018model} by utilizing extensive data from multiple domains. \rev{Finally}, self-training-based domain adaptation involves adapting the model using high-confidence pseudo-labels generated from the target domain data. To regulate training and prevent drift in pseudo-labels, strategies such as confidence thresholding \cite{zou2018unsupervised}, pseudo-label prototypes \cite{zhang2021prototypical}, and consistency regularization through data augmentation \cite{araslanov2021self} and varying contexts \cite{zhou2022context} are employed. Moreover, when extending domain adaptation to encompass multiple target domains, domain transfer techniques are applied \cite{lee2022adas} to improve generalization across these domains.

Domain datasets frequently exhibit a long-tail distribution, leading to a model bias towards more prevalent classes \cite{wang2017learning}. In the field of domain adaptation, strategies such as loss re-weighting \cite{zou2018unsupervised} and class-balanced sampling tailored for image classification \cite{prabhu2021sentry} have been implemented to address this issue. HRDA \cite{hoyer2023domain}  introduces Rare Class Sampling (RCS) to extend these techniques from classification tasks to semantic segmentation, which tackles the challenge of both rare and common classes appearing within each semantic segmentation category. HRDA \cite{hoyer2023domain} has also expanded the application of scale attention \cite{tao2020hierarchical} to domain adaptation, highlighting its critical role in enhancing domain robustness by improving the generalization capabilities across varying object sizes.

Some semantic segmentation models often incorporate multi-scale \emph{features}, a practice popularized by the widespread use of DeepLabV2 in domain adaptation \cite{chen2017deeplab}. Subsequent studies have explored techniques such as multi-scale average pooling during inference \cite{araslanov2021self} and enforcing scale consistency \cite{subhani2020learning,iqbal2020mlsl} for low-resolution inputs to facilitate multi-scale feature fusion. In addition, previous research \cite{li2017learning} has demonstrated that using knowledge distillation from prior tasks can serve as a regularization mechanism for new tasks, thereby enhancing the performance of domain adaptation \cite{chen2018road}. 

In this work, unlike domain adaption, the proposed AdvImmu can sense instantaneous, stable background, and dynamic information by considering temporal correlations, which is robust to arbitrary adverse conditions.

\subsection{Large Models (LMs) in AD}
Large Models (LMs) have gained significant interest in the context of AD due to their proficiency in analyzing images and contexts \cite{10295972}. These advancements have greatly improved zero-shot and few-shot image classification \cite{palanisamy2023proto}, segmentation \cite{kou2024pfedlvm}, object detection \cite{10465607}, etc. For example, CLIP \cite{radford2021learning} have shown that training to match images with captions can effectively create image representations from scratch; LLaMa \cite{liu2024visual} combines a vision encoder with an Large Language Model to enhance the understanding of both visual and linguistic concepts. 

Furthermore, researchers have explored the vectorized visual embeddings to equip LMs with environmental perception capabilities, particularly in AD scenarios. For example, DriveGPT4 \cite{xu2023drivegpt4} interprets video inputs to generate driving-related textual responses; Talk2BEV \cite{dewangan2023talk2bev} leverages pre-trained image-language models to combine Bird's Eye View maps with linguistic context, enabling visuo-linguistic reasoning in autonomous vehicles. 

Generative models in AD are also gaining popularity. For instance, GAIA-1 \cite{hu2023gaia} generates realistic driving scenarios by integrating video, text, and action inputs and showcases generative models in adapting to the changing dynamics of the real world. In this work, we propose to use SAM \cite{kirillov2023segment} to assist to generate frame-wise annotation. However, SAM's output lacks of category information. To mitigate this limitation, we additionally propose SBICAC clustering algorithm to supplement SAM to generate category-aware annotations.

\subsection{Knowledge Distillation}
In the domain of DL, knowledge distillation is defined as the technique where knowledge is transferred from a large, intricate model (\ie, teacher model) to a smaller, more streamlined model (\ie,  student model) \cite{gou2021knowledge}. This method is crucial for addressing the computational and resource limitations associated with the deployment of large-scale models in real-world settings.

Prior to the advent of LMs, knowledge distillation primarily focused on the transfer of knowledge from a sophisticated and often large teacher model to smaller and more efficient student models \cite{sanh2019distilbert}. The drive behind this process was the necessity to implement machine learning models in environments constrained by computational and memory resources, such as mobile or edge computing devices. The emphasis was mainly on selecting neural architectures and designing training objectives specifically for individual tasks. Earlier approaches included training a compact student model to replicate the output the teacher model. Generally, employing methods like soft target training to enable the student model to learn from the softened softmax outputs of the teacher model. 

With the emergence of LMs, the focus of knowledge distillation has significantly evolved. Unlike previous approaches that primarily targeted architectural compression, the modern approach in LM era emphasizes extracting and transferring knowledge \cite{codealpaca}. This shift is driven by the rich and intricate knowledge embedded within LMs such as SAM \cite{kirillov2023segment}. Moreover, the vast and often unmanageable number of parameters in LMs complicates the use of traditional size-reduction techniques like pruning \cite{han2015deep_compression} or quantization \cite{liu2023llm}. Currently, the objective of knowledge distillation in the context of LMs is not merely to replicate the teacher model’s output or to reduce the model size, but rather to harness and transfer the distinct knowledge contained within these powerful LMs.

The modern approach to knowledge distillation within LMs centers around the use of crafted prompts. These prompts are designed to specifically elicit targeted knowledge \cite{ding2023enhancing} or capabilities \cite{codealpaca} from LMs. They effectively access the LMs' comprehension and skills across a spectrum of areas, from natural language understanding \cite{he2023annollm} to more sophisticated cognitive functions such as reasoning \cite{hsieh2023distilling} and problem-solving \cite{qiao2024autoact}. The strategic use of prompts introduces a effective and flexible method of knowledge extraction, enabling precise targeting of skills or areas of interest. This technique proves especially beneficial in leveraging the emergent abilities of LMs. 

Additionally, the new paradigm in knowledge distillation extends mere output replication to transfer of more nuanced attributes like reasoning patterns \cite{mitra2023orca2}, preference alignment \cite{cui2023ultrafeedback}, and value alignment \cite{sun2023principledriven}. This shift represents a move towards a broader and more in-depth transfer of cognitive skills. Current practices not only replicate outputs but also emulate the cognitive processes \cite{mitra2023orca2} and decision-making approaches \cite{asai2023self} of the teacher model. Techniques such as chain-of-thought (CoT) prompting are employed, where the student model is trained to internalize and replicate the teacher's reasoning pathway, thus enhancing its capabilities in problem-solving and decision-making. In this paper, to generate consecutive frame-wise annotations, we propose to distill knowledge from SAM \cite{kirillov2023segment} and then use the distilled knowledge along with the proposed unsupervised SBICAC clustering algorithm to generate pseudo-labels for consecutive frames.

\section{Methodology}
\label{methodology}
To depict the proposed AdvImmu clearly, we denote the combination of LSM and GSM as LSaGS hereafter. We will introduce LSaGS in \Cref{LSaGS_method}. Subsequently, we will detail URs in \Cref{URs_method}, including the definition of both regularizers and the proposed unfolding method. Finally, we will present how the proposed SBICAC algorithm along with SAM to generate frame-wise pseudo-labels in \Cref{SBICAC_method}.

\subsection{LSaGS} \label{LSaGS_method}
LSM is proposed to process local temporal dependency to enhance the model performance. Specifically, LSM extracts immediate, stable background, and dynamic features through InsU, IntU, and DU, respectively. Concretely, InsU is designed to process the current image as \rev{the} vehicle drives. It extracts instantaneous information from this image, which is crucial for the vehicle to accurately understand immediate surroundings. IntU focuses on processing stable background information from the environment, achieved by analyzing past consecutive images prior to the current frame. This enables IntU to understand what remains constant over time, which is important for consistent understanding of the scene. DU analyzes dynamic changes in the environment by comparing the current frame against previous ones, which helps to highlight the dynamic objects, such as pedestrians and vehicles. In this work, we propose to use ResNet50 \cite{he2016deep} to serve as the network architecture of InsU, IntU and DU. Notably, LSM contains a hyper-parameter named LSM depth, which is used to indicate the count of the considered consecutive frames prior to current frame. LSM depth definitely influences IntU and DU.

\begin{figure}[tp]
\centering 
\includegraphics[width=1.4\linewidth, height=0.99\linewidth]{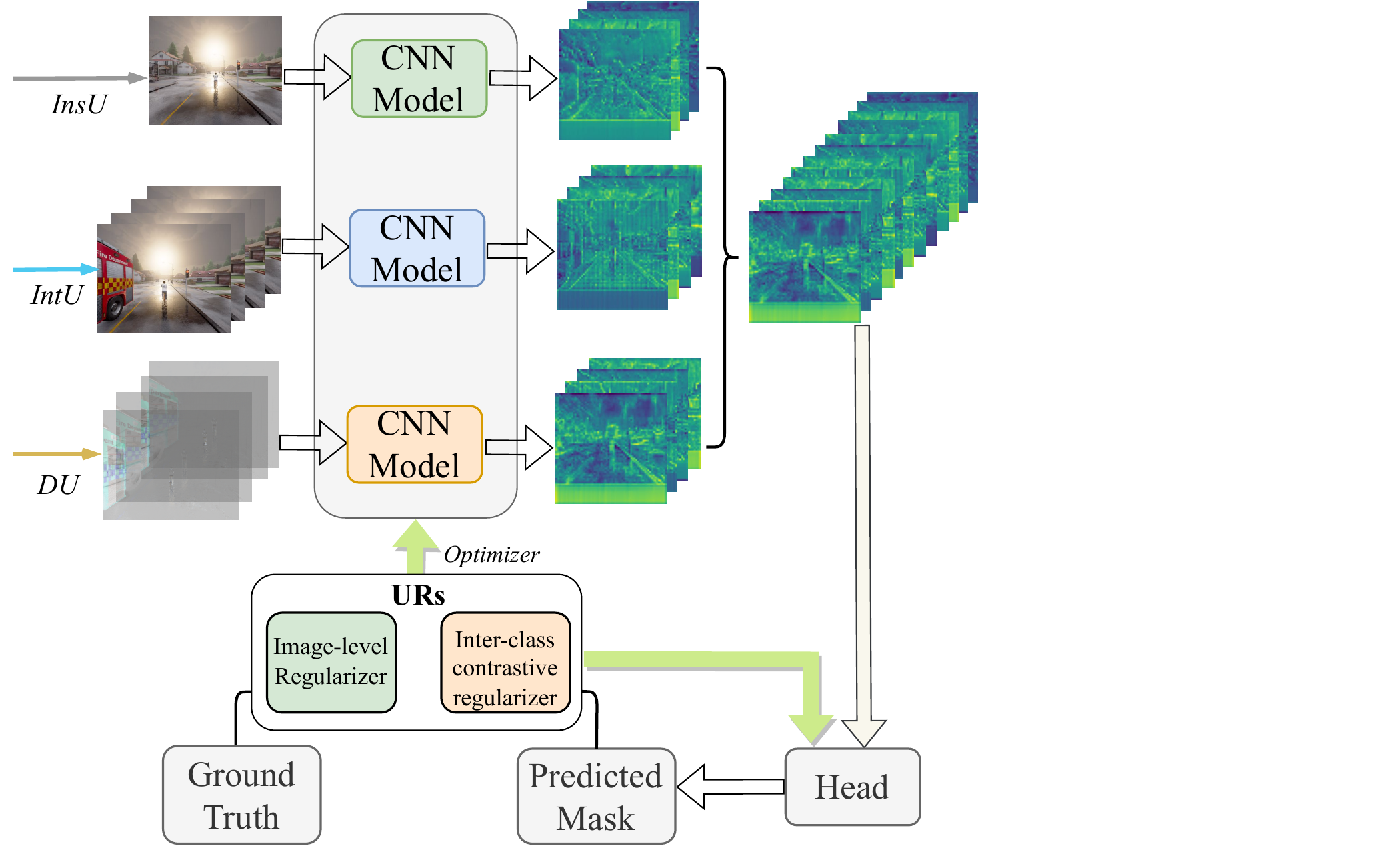}
\vspace{-0.2cm}
\caption{Illustration of LF.}
\label{Fig.LF}
\vspace{0.5cm}
\end{figure}

Early Fusion (EF) and Late Fusion (LF) are two policies for fusing features from InsU, IntU, and DU. In EF strategy, features from InsU, IntU, and DU are first merged. Then the merged features are processed by a shared CNN model. This early integration allows the model to utilize the correlations between these different types of features, enhancing the model for understanding street scenes. On the other hand, in LF strategy, InsU, IntU, and DU process their own features by three separate CNN models. This allows each type of features to be finely analyzed on its own. After processing, these different types of features are combined. EF is summarized in the right part of \Cref{Fig.LSM_with_EF} and LF is outlined in \Cref{Fig.LF}.

GSM is introduced to shuffle and group different LSM segments into mini-batches. In this way, each mini-batch can reflect the diversity of the dataset instead of following a strict sequence. This enables the model learn more general and robust patterns, enhances its performance and generalization. Based on the shuffled mini-batches by GSM, the involved CNN models are then optimized by back propagation.

After the introduction of the philosophy of LSM and GSM, we then formulate LSaGS in Algorithm \ref{Algo:LSaGS} \rev{as} a whole picture of LSaGS. Specifically, LSaGS iterates multiple rounds. In each round, it follows below steps: \textbf{(I) Extract features:} LSaGS takes video sequence $F$ as input to process. For frame $F_t \in F$, InsU (with parameters $\phi_{\text{InsU}}$) processes immediate features $InsU_t$ from $F_t$. IntU (with parameters $\phi_{\text{IntU}}$) analyzes the background by looking at the past $d$ frames (i.e., $F_{t-i}$ for $i=1$ to $d$) to gather historical context features $IntU_t$. DU (with parameters $\phi_{\text{DU}}$) examines changes in the current frame $F_t$ relative to the past $d$ frames, capturing dynamic change features $DU_t$. \textbf{(II) Merge features:} After extracting features by InsU, IntU, and DU, LSaGS performs feature fusion based on the specified fusion method. For EF, $InsU_t$, $IntU_t$, and $DU_t$ are merged into $MF_t$, which is then processed by a shared CNN (with parameters $\text{CNN}_{\text{shd}}$) to output the processed features $PF_t$. For LF, $InsU_t$, $IntU_t$, and $DU_t$ are processed separately by their respective CNNs (with parameters $\text{CNN}_{\text{InsU}}$, $\text{CNN}_{\text{IntU}}$, $\text{CNN}_{\text{DU}}$, respectively) to extract features $PInsU_t$, $PIntU_t$ and $PDU_t$. Such features are then merged to form the final processed features $PF_t$. \textbf{(III) Group and Shuffle Mini-Batches:} After processing all frames, LSaGS shuffles and groups $PF_t$ into mini-batches. This step helps in preparing the data for training. \textbf{(IV) Optimization via back propagation:} LSaGS is optimized by back propagation in a batch-by-batch way.  

\setlength{\textfloatsep}{0pt}
\begin{algorithm}[tp]
\caption{LSaGS}
\label{Algo:LSaGS}
\SetAlgoLined
\KwIn{Input sequence $\{F\}$, LSM depth $d$, Fusion method $FusionType$, Epochs $epochs$}
\KwOut{$\text{Head},\text{CNN}_{\text{shd}}\ (or\ \text{CNN}_{\text{InsU/IntU/DU}})$}
\BlankLine
\For{$i \gets 1$ \KwTo $epochs$}{
\tcp{LSM}
\For{frame $F_t$ in $\{F\}$}{
    $InsU_t \gets \phi_{\text{InsU}}(F_t)$ \tcp*{Process immediate features from $F_t$}
    
    $IntU_t \gets \phi_{\text{IntU}}(\{F_{t-i}\}_{i=1}^d)$ \tcp*{Analyze background from past $d$ frames}
    
    $DU_t \gets \phi_{\text{DU}}(\{F_t\hspace{-0.1cm}-\hspace{-0.1cm}F_{t-i}\}_{i=1}^d)$ \tcp*{Analyze changes of $F_t$ to past $d$ frames}
    
    \BlankLine
    \tcp{Fusion of InsU, IntU, and DU}
    \If{$FusionType == \text{EF}$}{
        $MF_t \gets \text{Merge}(InsU_t, IntU_t, DU_t)$; \\
        $PF_t \gets \text{CNN}_{\text{shd}}(MF_t)$; \\
    }
    \ElseIf{$FusionType == \text{LF}$}{
        $PInsU_t \gets \text{CNN}_{\text{InsU}}(InsU_t)$; \\
        $PIntU_t \gets \text{CNN}_{\text{IntU}}(IntU_t)$; \\
        $PDU_t \gets \text{CNN}_{\text{DU}}(DU_t)$; \\
        $PF_t \gets \text{Merge}(PInsU_t, PIntU_t, PDU_t)$; \\
    }
}
\BlankLine
\tcp{GSM}
$MiniBatches \gets \text{Shuffle\&Group } \{PF_t\}$ \\
\For{mini-batch $B$ in $MiniBatches$}{
    $\text{CNN}_{\text{shd}}\ (or\ \text{CNN}_{\text{InsU/IntU/DU}}) \gets B$; \\
}
}
\end{algorithm}

\subsection{Unfolded Regularization} \label{URs_method}
On top of the conventional pixel-level cross entropy loss (denoted as $L_{CE}$) in semantic understanding, we propose an image-level regularizer and an inter-class contrastive regularizer to enhance the model's across-weather generalization. URs are the deep unfolding implementation of such regularizers, which aims at avoiding the heuristic and exhaustive searches on weights of them.

\subsubsection{Definition of the proposed Regularizers}
To formulate the definition clearly, we denote each input image as $F$, \rev{the} LSaGS's predicted masks as $X$ and the corresponding ground truth as $Y$. \rev{Furthermore}, we term each pixel's prediction within $X$ as $x$, and denote the distribution of $X$ and $Y$ as $P_X$ and $P_Y$, respectively. 

On the one hand, we propose to use Bhattacharyya distance (BD) as the image-level regularizer to quantify the distribution similarity between $P_X$ and $P_Y$. The definition of the BD regularizer over input image $X$ is given by:
\begin{align}
L_{BD} = D_B(P_X, P_Y) = -\ln(\sum\nolimits_{x \in X} \sqrt{P_X(x) P_Y(x)}),
\end{align}

On the other hand, we propose to use InfoNCE as an inter-class contrastive regularizer that can pull pixles of same class closer while push pixles of different classes apart. The InfoNCE regularizer is expressed as:
\begin{align}
\hspace{-0.2cm}L_{con}\!=\!-\!\log\!\frac{\sum_{m=1}^{|\mathcal{U}|}\exp(\frac{\sigma(x, u_{m})}{\tau})}{\sum_{m=1}^{|\mathcal{U}|}\!\exp(\frac{\sigma(x, u_{m})}{\tau})\!+\!\sum_{n=1}^{|\mathcal{V}|}\exp(\frac{\sigma(x, v_n)}{\tau})},
\end{align}
where \(x\) is an anchor pixel, \(u_m\) is any pixel from the same-class pixle set \(\mathcal{U}\), \(v_n\) is any pixel from the different-class pixel set \(\mathcal{V}\), \(\sigma(\cdot, \cdot)\) denotes the inner product, and \(\tau\) controls the separation sharpness.

\begin{figure*}[tp]
\centering
\includegraphics[width=\linewidth,height=0.3\linewidth]{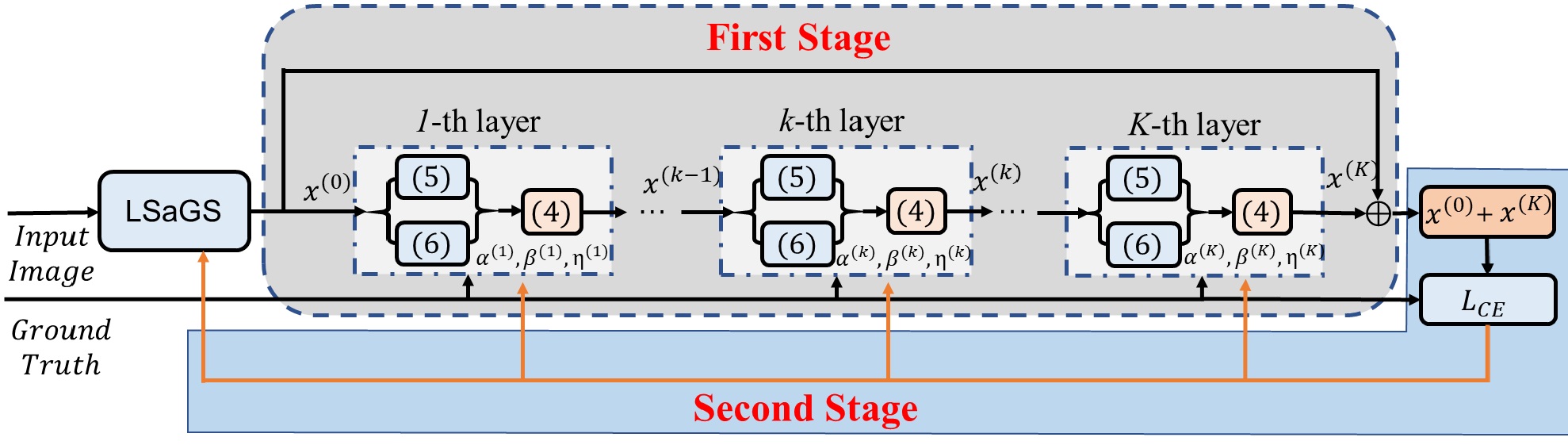}
\vspace{-0.3cm}
\caption{Illustration of AdvImmu architecture, which contains LSaGS and URs. (4), (5), (6) represent equation number.}
\label{Fig.RDU}
\vspace{-0.3cm}
\end{figure*}

\subsubsection{Unfolded Regularizers (URs)}
On top of the conventional cross entropy loss $L_{CE}$, considering the proposed $L_{BD}$ and $L_{con}$, the total loss \( L \) for semantic segmentation task is
\begin{align}
    L = L_{CE} + \alpha L_{BD} + \beta L_{con},
\end{align}
where \( \alpha \) and \( \beta \) are the weights assigned to \( L_{BD} \) and \( L_{con} \), respectively. \rev{Traditionally, $\alpha$ and $\beta$} are tuned based on heuristic or exhaustive searches, and then the total loss $L$ \rev{is used} to optimize the model in training phase. This conventional optimization strategy generally results in instability and suboptimal performance because there is no guarantee optimal \( \alpha \) and \( \gamma \) by \rev{a finite} number of searches. To tackle this issue, we propose a two-stage optimization policy based on above discussed $L_{BD}$ and $L_{con}$ regularizers. 

For the first stage, we propose to convert weights $\alpha$ and $\beta$ as trainable parameters rather than searched hyper-parameters. To this end, inspired by \cite{10529194,lin2023communication}, we propose to treat each optimization iteration of the proposed $L_{BD}$ and $L_{con}$ as one layer, and treat $\alpha$ and $\beta$ as learnable parameters within each layer. Based on this, both regularizers can be unfolded into $K$ layers (illustrated in the right part of \Cref{Fig.RDU}) by using deep unfolding technique. Then such $K$ unfolded layers are appended at the end of the LSaGS model, serving as the part of AdvImmu model architecture. 

To formulate the update rule of URs, for layer $k, \text{where}\ k \in \{1, 2, \cdots, K\}$, the output $x^{(k)}$ corresponding to pixel $x \in X$ is updated as follow:
\begin{align}
    x^{(k)} =\hspace{0.15cm} &x^{(k-1)}-\eta^{(k)} (\alpha^{(k)}\nabla_{x^{(k-1)}} L_{BD}(x^{(k-1)}) \hspace{0.15cm}+\hspace{0.15cm}
    \nonumber \\
    &\beta^{(k)}\nabla_{x^{(k-1)}} L_{con}(x^{(k-1)})),
    \label{Eq:RDU_update}
\end{align}
where $x^{(0)}$ is the input of URs and it exactly equals to $x \in X$, $\alpha^{(k)}$ and $\beta^{(k)}$ are the learnable weights, $\eta^{(k)}$ is the learnable step size of update, $\nabla_{x^{(k-1)}} L_{BD}(x^{(k-1)})$ and $\nabla_{x^{(k-1)}} L_{con}(x^{(k-1)})$ are the gradients of $L_{BD}$ and $L_{con}$ relative to pixel $x^{(k-1)}$, respectively, and they can be formulated as follows (denoting $x^{(k-1)}$ as $x$ for short):
\begin{align}
\nabla_x L_{BD}(x) &= - \frac{1}{2}\frac{1}{\sum_{x} \sqrt{P_X(x) P_Y(x)}}  \frac{\sqrt{P_Y(x)}}{\sqrt{P_X(x)}},
\label{Eq:grad_L_BD}
\\
\nabla_x L_{\text{con}}(x) &= \nabla_x B / B - \nabla_x A / A,
\label{Eq:grad_L_con}
\vspace{-0.9cm}
\end{align}
where
\begin{align}
A &= \sum\nolimits_{m=1}^{|\mathcal{U}|} \exp(\sigma(x, u_{m}) / \tau),
\label{Eq:L_con_A}
\\
B &= A + \sum\nolimits_{n=1}^{|\mathcal{V}|} \exp(\sigma(x, v_{n}) / \tau),
\label{Eq:L_con_B}
\\
\nabla_x A &= \frac{1}{\tau} \sum\nolimits_{m=1}^{|\mathcal{U}|} \exp(\sigma(x, u_{m}) / \tau) \cdot u_{m}, 
\label{Eq:grad_L_con_A}
\\
\!\nabla_x B \! &=\! \nabla_x A\! +\! \frac{1}{\tau}  \sum\nolimits_{n=1}^{|\mathcal{V}|}\! \exp(\sigma(x, v_{n}) / \tau) \cdot v_{n}.
\label{Eq:grad_L_con_B}
\end{align}
After multiple gradient descent updates by $K$ sequentially-connected layers, the output of the $K$-th layer is optimized by both regurlarizers. 

For the second stage, the optimized output of the $K$-th unfolded layer is added to the output of LSaGS model.
Then we use the summation results serving as predicted masks and the ground truth to calculate the cross entropy loss $L_{CE}$ to optimize LSaGS and \{$\alpha^{(k)}$, $\beta^{(k)}$, $\eta^{(k)}\}, \text{where}\ k \in \{1, 2, \cdots K\}$ by back propagation. Notably, this stage \rev{follows} the traditional model optimization.

In conclusion, the proposed AdvImmu contains LSaGS and URs, which are collectively illustrated in \Cref{Fig.RDU}. In addition, AdvImmu is further outlined in Algorithm \ref{Algo:DUN}, where $K$ is a hyper-parameter and can be determined according to \rev{practical} needs and conditions. 

\begin{algorithm}[tp]
\caption{AdvImmu}
\label{Algo:DUN}
\SetAlgoLined
\KwIn{$F$ (Input image from sequence), $Y$ (One-hot ground truth), $K$ (Layer number)}
\KwOut{Model $LSaGS$, Learnable variables $\alpha^{(k)}$, $\beta^{(k)}$, $\eta^{(k)}, \text{where}\ k \in \{1, 2, \cdots K\}$}
\BlankLine
Initialize model $LSaGS \gets \mathcal{W}_0$, and  $\alpha^{(k)}$, $\beta^{(k)}$, $\eta^{(k)} \gets \alpha_0, \gamma_0, \eta_0, \text{where}\ k \in \{1, 2, \cdots K\}$; \\
\BlankLine
\For {$epoch\ i \gets 1$ \KwTo $max\_epochs$}{
\tcp{Forward propagation}
$x^{(0)} \gets LSaGS(F)$; \\
\For{$layer\ k \gets 1$ \KwTo $K$}{
    \tcp{Compute gradients}
    $\nabla_x L_{BD}(x^{(k-1)}) \gets \Cref{Eq:grad_L_BD}$; \\
    $\nabla_x L_{\text{con}}(x^{(k-1)}) \gets \Cref{Eq:grad_L_con}$; \\
    \BlankLine
    \tcp{Update rule}
    $x^{(k)} \gets \Cref{Eq:RDU_update}$; \\
}
\BlankLine
$\hat{x} \gets x^{(0)} + x^{(K)}$; \\
\BlankLine
\tcp{Back propagation}
$L_{CE} = CrossEntropy (\hat{x}, Y)$; \\
$LSaGS, \alpha^{(k)}$, $\beta^{(k)}$, $\eta^{(k)}, \text{where}\ k \in \{1, 2, \cdots K\} \gets L_{CE}.Backward()$; 
}
\end{algorithm}

\subsection{SBICAC's Annotation based on SAM-distilled Knowledge} \label{SBICAC_method}
\begin{figure}[tp]
\vspace{-2.13cm}
\hspace{0.03cm}
\includegraphics[width=1.08\linewidth,height=0.9\linewidth]{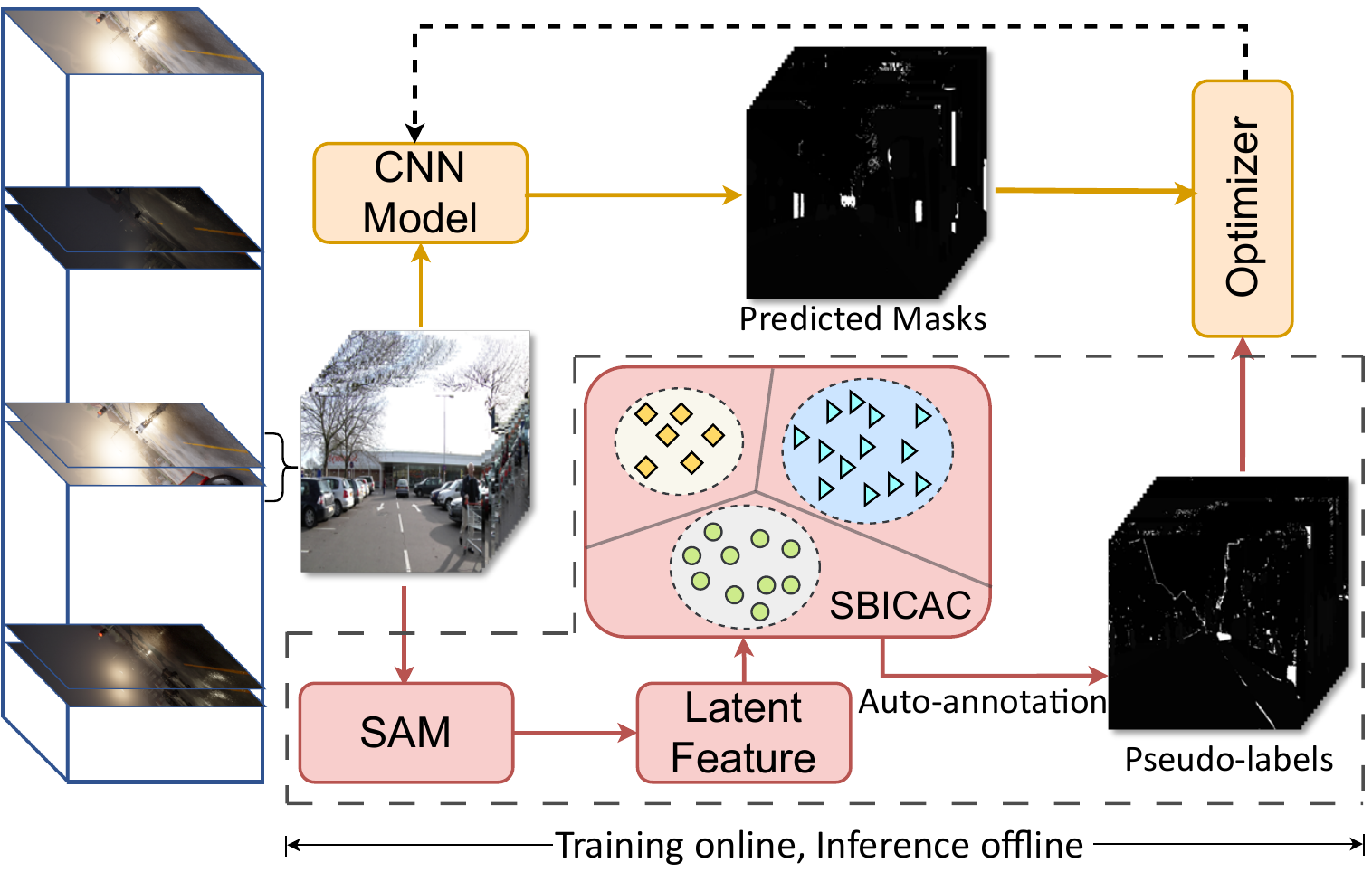}
\caption{Overview of SAM-aided auto-annotation system. SAM and SBICAC in the dashed box just operate in training phase.
}
\label{Fig.SAM_labelling}
\end{figure}

The training of the proposed AdvImmu requires frame-wise annotation for consecutive frames, which is impractical to be \rev{done} manually. To generate the required frame-wise annotation automatically, we propose an unsupervised clustering algorithm based on the distilled knowledge from a large vision model SAM.

SAM, a vision foundation model known for its strong zero-shot inference abilities, is used to \rev{assist generating} masks for images in our research. It should be noted that SAM operates online during the training phase and offline during the inference phase, as illustrated in the dashed box in \Cref{Fig.SAM_labelling}. However, SAM does not support \rev{generating} category information (as demonstrated in \Cref{Fig.SAM_reason_to_cluster}). To overcome this limitation, SBICAC algorithm is introduced to assign each semantic class a unique ID based on SAM-distilled latent features, which are then served as labels for training the model. SBICAC follows an iterative refinement manner and each iteration is described as follows:

\begin{figure*}[tp]
\hspace{-0.6cm}
\subfigure[\textbf{Apolloscapes image}]{
\label{Fig.row_image}
\includegraphics[width=0.227\linewidth,height=0.16\linewidth]{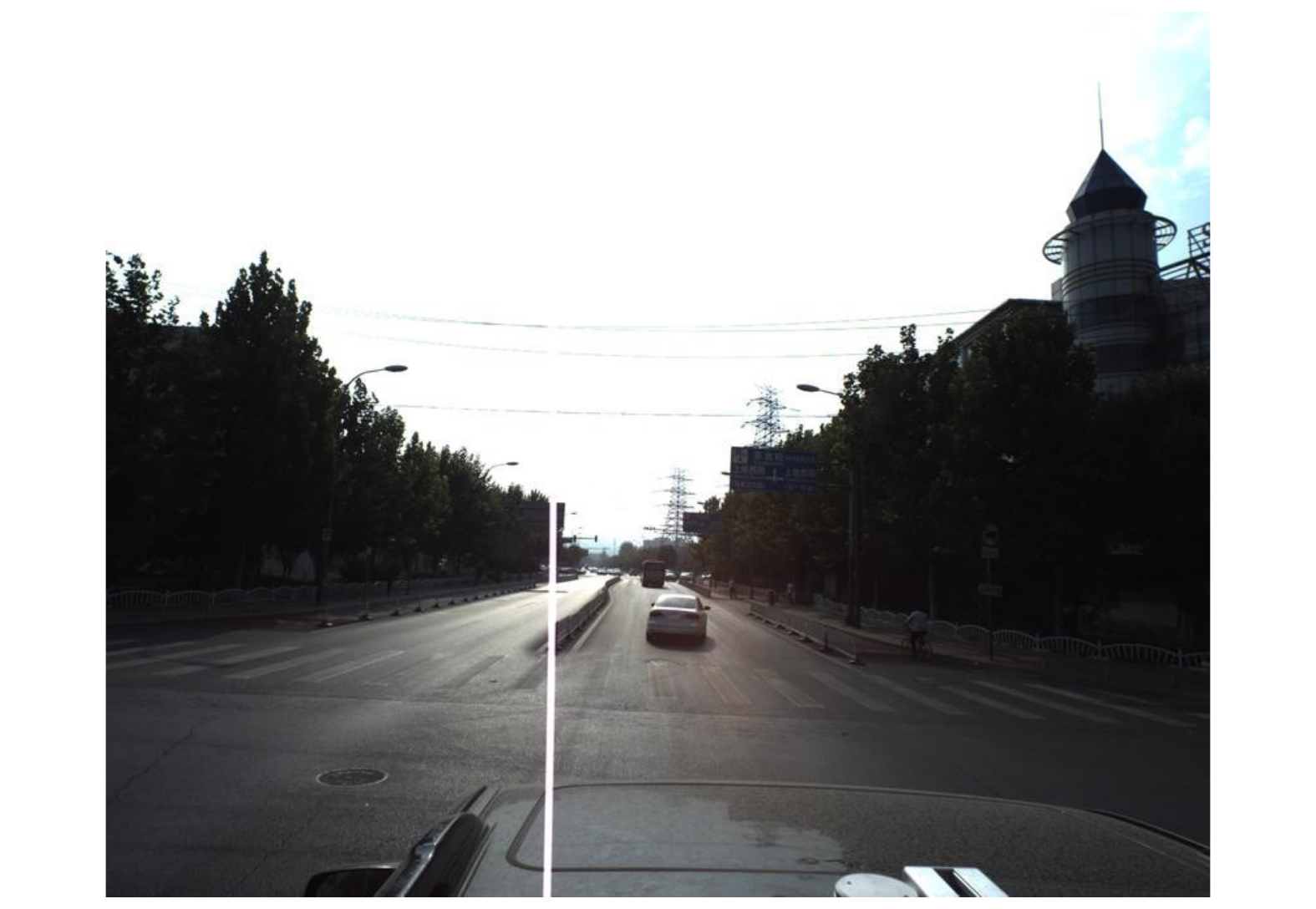}}
\hspace{-1.0cm}
\subfigure[\textbf{AgglomerativeClustering}]{
\label{Fig.SAM_generated_mask}
\includegraphics[width=0.227\linewidth,height=0.16\linewidth]{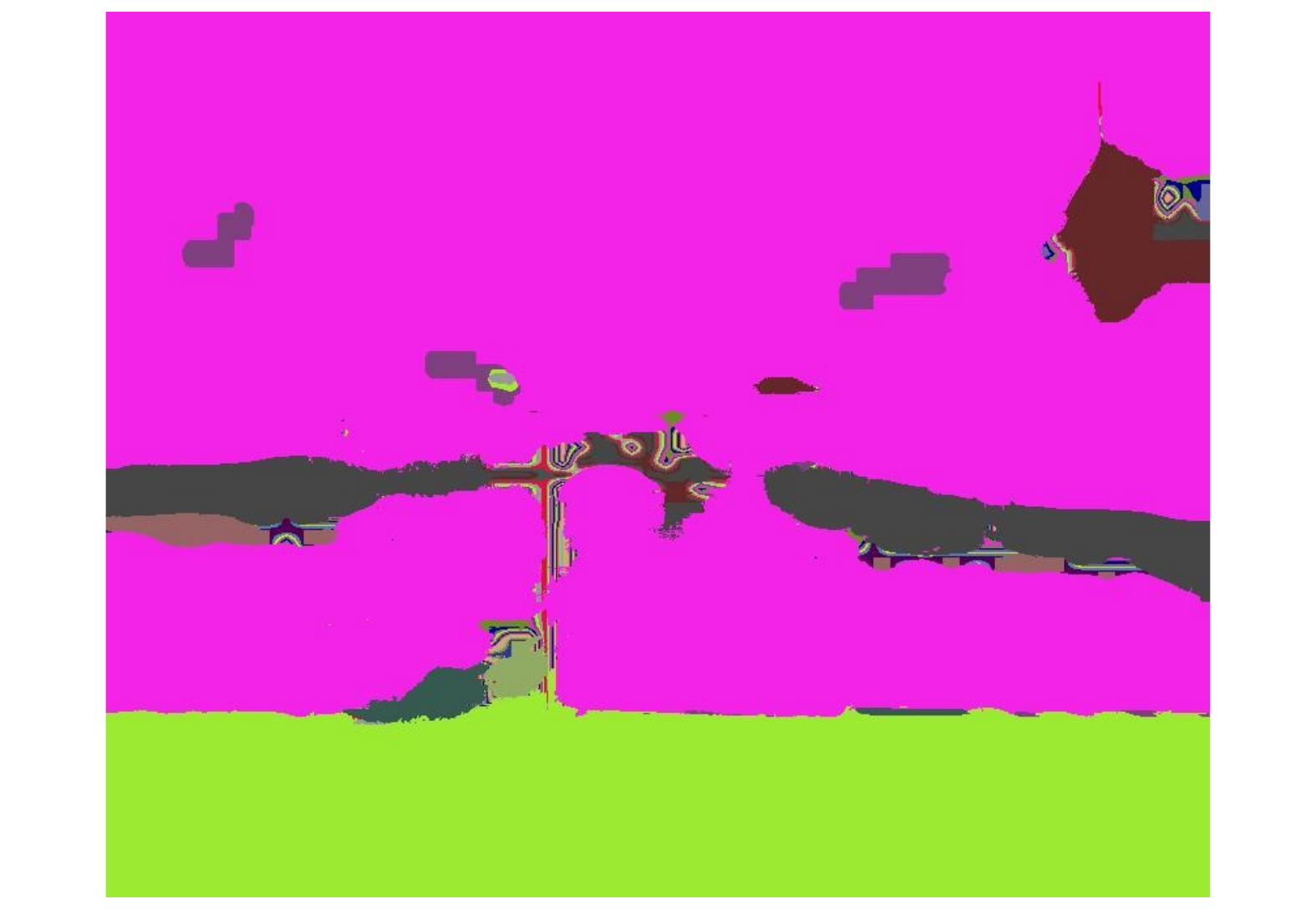}}
\hspace{-1.0cm}
\subfigure[\textbf{KMeans}]{
\label{Fig.row_image}
\includegraphics[width=0.227\linewidth,height=0.16\linewidth]{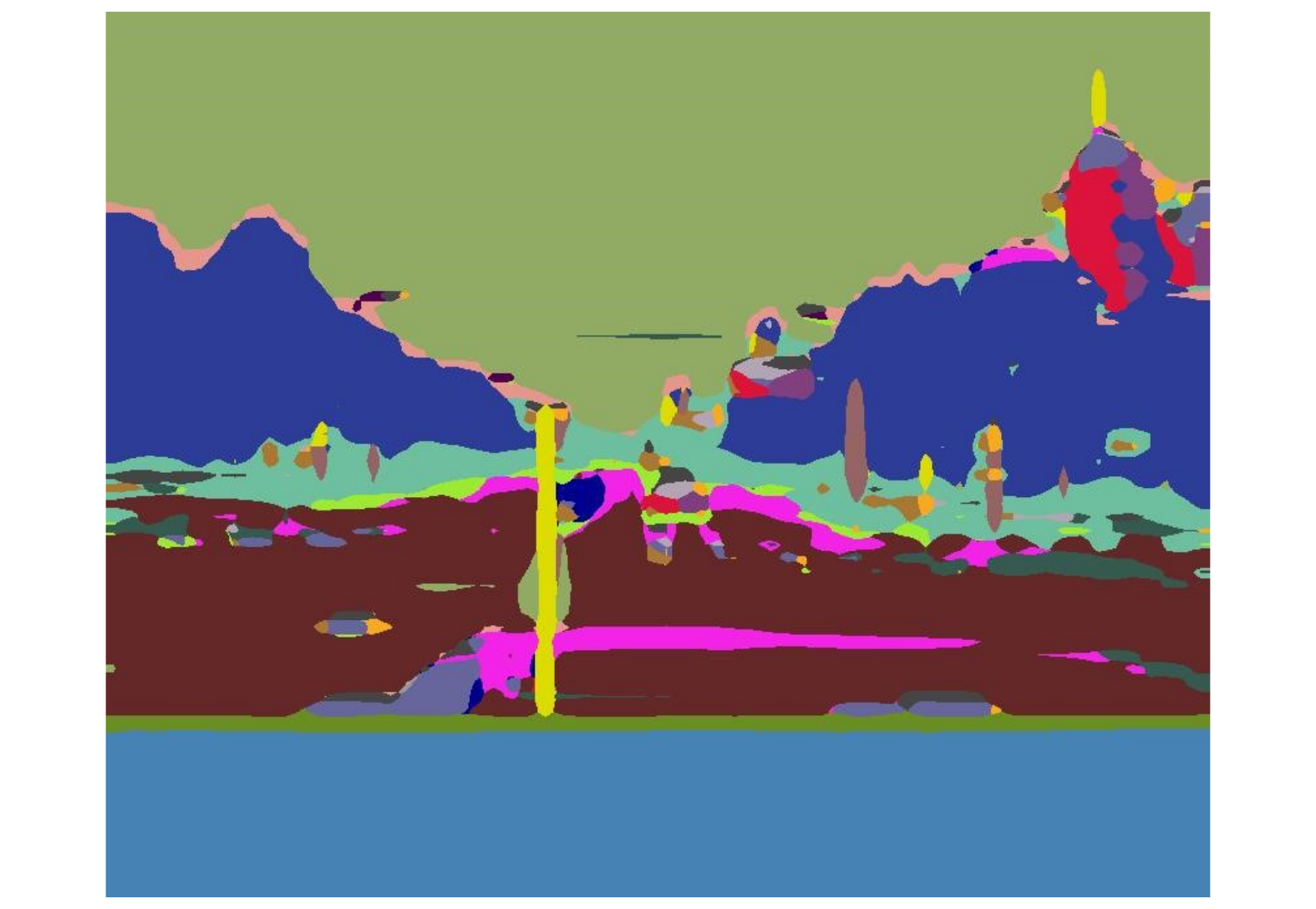}}
\hspace{-1.0cm}
\subfigure[\textbf{SpectralClustering}]{
\label{Fig.SAM_generated_mask}
\includegraphics[width=0.227\linewidth,height=0.16\linewidth]{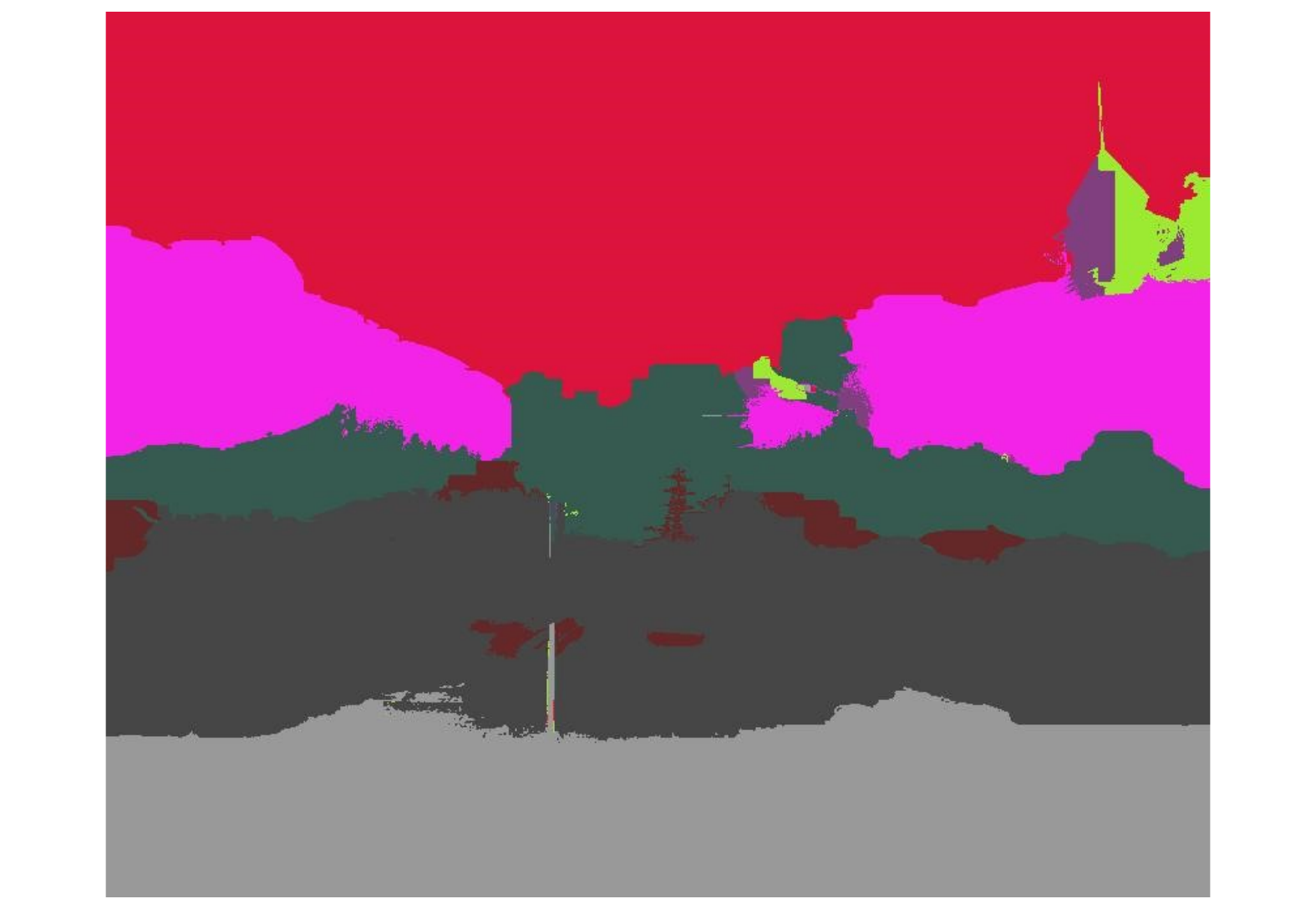}}
\hspace{-1.0cm}
\subfigure[\textbf{SBICAC (Ours)}]{
\label{Fig.SAM_generated_mask}
\includegraphics[width=0.227\linewidth,height=0.16\linewidth]{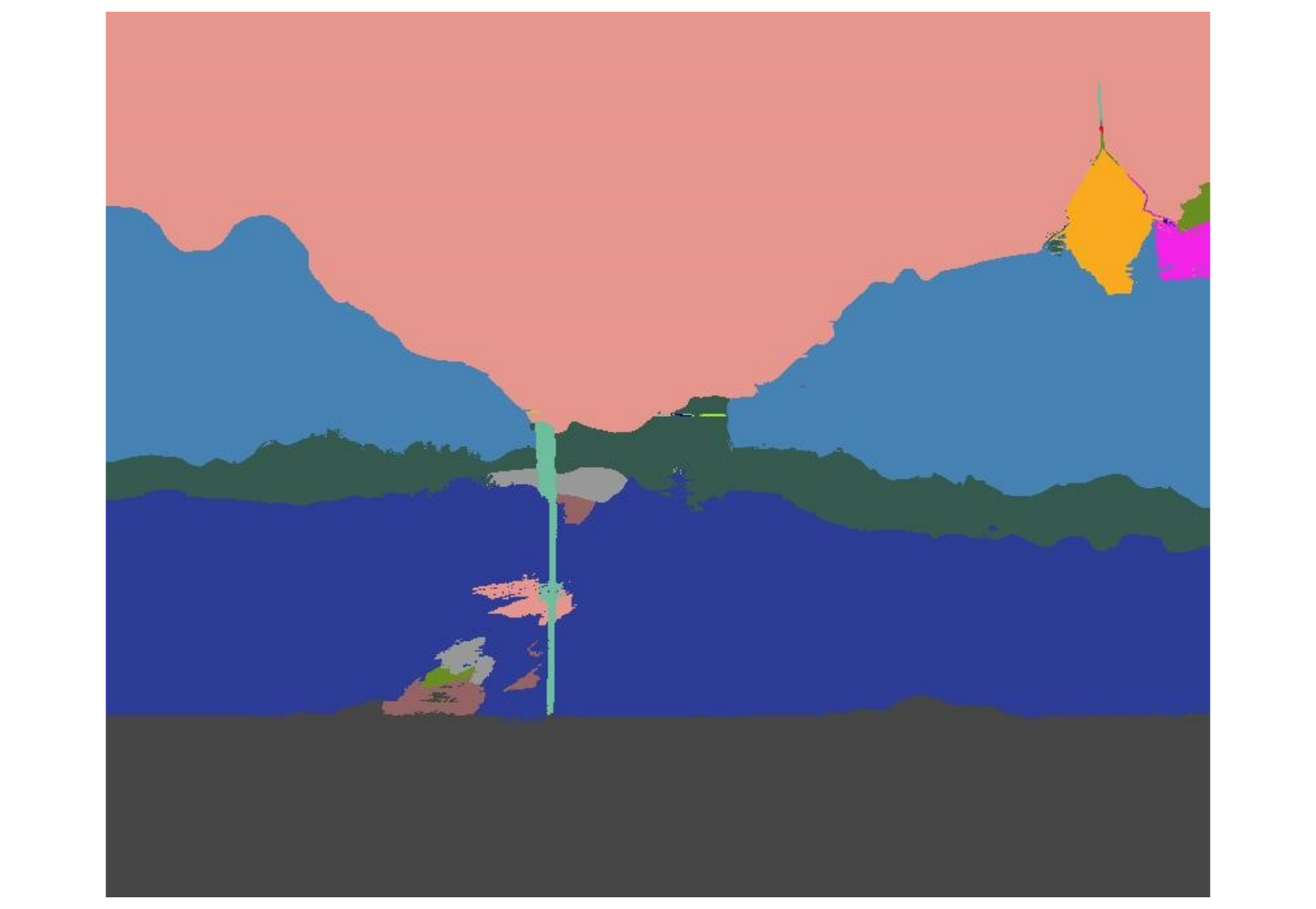}}
\vspace{-0.1cm}
\caption{The intuitive comparison of SBICAC against other clustering algorithms. (Just focus on the shape rather than the color in semantic masks, as different algorithms may use different colors for the same category.)}
\label{Fig.cluster_comp}
\vspace{-0.2cm}
\end{figure*}

\begin{enumerate}
    \item \textbf{Similarity Computation}:
    For each pixel in $X$, the similarity to each cluster $centroid$ is computed using the inner product. The results store in a similarity matrix where each element represents the similarity between a pixel and a cluster $centroid$. 

    \item \textbf{Cluster Assignment}:
    Each pixel is assigned to the cluster with which it has the highest similarity. This is achieved by finding the index of  maximum value in the similarity matrix for each pixel.

    \item \textbf{Cluster Centroid Calculation}:
    The centroids are then updated based on the new cluster assignments. Specifically, for cluster $j$, the $new\_centroid$ is computed by averaging all pixel representation of that cluster:
    \[
    new\_centroids[j] = \frac{1}{|X_j|} \sum_{k=1}^{|X_j|} X_j^{(k)},
    \]
    where $X_j$ is the set of pixels assigned to cluster $j$.
    
    \item \textbf{Convergence Check and Centroid Update}:
    SBICAC checks for convergence by comparing $new\_centroids$ to $centroids$. If $new\_centroids$ have not changed significantly (measured by a function like $np.allclose$), SBICAC terminates earlier; otherwise, $centroids$ are updated by $new\_centroids$ and next iteration begins.
\end{enumerate}

In summary, SBICAC leverages inner product similarity to iteratively assign pixels to clusters and update clusters' centroids until the cluster assignments stabilize. SBICAC ensures that the clusters are formed based on maximizing the similarity between pixels and their respective centroids. Upon completion of the iterative process, SBICAC outputs the cluster assignments $labels$ for all pixels. The whole pipeline of using SBICAC+SAM to generate frame-wise annotation is illustrated in \Cref{Fig.SAM_labelling}, and SBICAC clustering algorithm is outlined in Algorithm \ref{Algo:SBICCR}. 

In addition, \Cref{Fig.cluster_comp} exemplifies the annotation of SBICAC against other off-the-shelf clustering algorithms for a specific image. We can observe that KMeans exists the issue of small object error, while AgglomerativeClustering and SpectralClustering face the problem of larger boundary uncertainty. Obviously, the proposed SBICAC achieves relatively better performance compared to those off-the-shelf competitors.

\setlength{\textfloatsep}{0pt}
\begin{algorithm}[tp]
\caption{Similarity-based Iterative Cluster Assignments and Centroids Refinement (SBICAC)}
\label{Algo:SBICCR}
\SetAlgoLined
\KwIn{$X$ (Latent feature maps), $n\_clusters$ (Number of clusters), $init$ (Initial random centroids), $max\_iter$ (Maximum iterations)}
\KwOut{Cluster assignments $labels$}
\BlankLine
Initialize centroids as $centroids \gets init$; \\
\For{$i \gets 1$ \KwTo $max\_iter$}{
    Compute similarity between $X$ and $centroids$: $similarity \gets X \cdot centroids^T$; \\
    Assign clusters based on maximum similarity: $labels \gets \arg\max(similarity, axis=1)$;\\
    Cluster Centroid Calculation: \\
    \For{$j \gets 1$ \KwTo $n\_clusters$}{
        $cluster\ j$: $new\_centroids[j] \gets \frac{1}{|X_j|} \sum_{k=1}^{|X_j|} X_j^{(k)}$;
    }
    Check for convergence: \\
    \If{np.allclose$(centroids, new\_centroids)$}{
        \textbf{break}; \\
    }
    Update centroids: $centroids \gets new\_centroids$; \\
}
\end{algorithm}

\section{Experiments}
\label{experiments}
In this section, we firstly introduce the experimental settings in \Cref{datasets_metrics_imp}. Based on such settings, we will evaluate the proposed AdvImmu's performance, convergence, complexity and inference delay in \Cref{AdvImmu_evaluation}. Subsequently, we conduct ablation studies about AdvImmu in \Cref{abl_study}. Finally, we will evaluate SBICAC in \Cref{SBICAC_eval}.

\subsection{Datasets, Metrics and Implementation}
\label{datasets_metrics_imp}
\subsubsection{Datasets}
In the evaluation, we utilize six datasets: 
\begin{itemize}
    \item  The Apolloscapes dataset \cite{wang2019apolloscape} is a comprehensive AD dataset featuring a variety of driving scenarios captured under various weather conditions. For our purposes, we use a subset of this dataset, which includes 854 training images and 400 test images, each annotated with pixel-level labels across 23 classes such as vehicle, pedestrian, etc. 
    \item  The CARLA\_ADV dataset derives from the Carla simulator (version 0.9.13) \cite{dosovitskiy2017carla}. CARLA\_ADV dataset is specifically designed to encompass a range of adverse weather conditions including fog, clouds, rain, darkness, and combinations thereof. It contains 2,764 training images and 1,921 test images, annotated at the pixel level for 23 classes, including vehicle, building, tree, etc.
    \item The Cityscapes dataset \cite{Cordts2016Cityscapes}, captured across multiple cities, comprises 2,975 training images and 500 test images. The dataset features pixel-level labels for 20 classes, such as vehicles, pedestrians, etc. Notably, we disable LSM in AdvImmu in case of the test on Cityscapes dataset because Cityscapes dataset does not contain consecutive frames.
    \item The GTA5 dataset \cite{richter2016playing} comprises 24,966 synthetic images, each with pixel-level semantic annotations. These images are rendered from the video game Grand Theft Auto 5, capturing scenes from ego-vehicle's perspective in American-style virtual cities. We use a subset of this dataset, including 2,000 training images and 500 test images. The dataset includes 20 semantic classes compatible with those in the Cityscapes dataset.
    \item The Shift dataset \cite{shift2022} consists of images captured across various adverse weather conditions and different time slot of the day. This dataset is particularly valuable for developing and testing algorithms that must perform consistently in diverse and dynamically changing real-world settings. The training and test datasets contain 7,218 and 2,875 images, respectively, where each image is annotated with pixel-level labels for 23 semantic classes.
    \item The SynthiaSF dataset \cite{ros2016synthia} features a collection of synthetic but photorealistic images that simulate urban driving scenarios. In our experiments, the training dataset and the test dataset include 1,596 and 628 images, respectively. This dataset includes pixel-level annotations for 23 semantic classes.
\end{itemize}

\subsubsection{Metrics}
We evaluate the proposed AdvImmu on street scene understanding task by employing four metrics: Mean Intersection over Union (\textbf{mIoU}), which measures the overlap between predicted mask and ground truth; Mean Precision (\textbf{mPre}), which assesses the accuracy of positive predictions; Mean Recall (\textbf{mRec}), evaluating how well the model identifies all relevant instances; and Mean F1 (\textbf{mF1}), which provides a balance between precision and recall. 
These metrics are formulated as follows: 
\begin{align}
    &mIoU = \frac{1}{\mathcal{C}}\sum_{c=1}^{\mathcal{C}}IoU_c = \frac{1}{\mathcal{C} \mathcal{N}}\sum_{c=1}^{\mathcal{C}} \sum_{n=1}^{\mathcal{N}} \frac{TP_{n, c}}{FP_{n, c}\hspace{-0.1cm}+\hspace{-0.1cm} TP_{n, c}\hspace{-0.1cm} +\hspace{-0.1cm} FN_{n, c}},
    \nonumber
    \\
    \vspace{-0.2cm}
    &mPre = \frac{1}{\mathcal{C}}\sum_{c=1}^{\mathcal{C}}Pre_c = \frac{1}{\mathcal{C} \mathcal{N}}\sum_{c=1}^{\mathcal{C}} \sum_{n=1}^{\mathcal{N}} \frac{TP_{n, c}}{FP_{n, c} + TP_{n, c}},
    \nonumber
    \\
    \vspace{-0.2cm}
    &mRec = \frac{1}{\mathcal{C}}\sum_{c=1}^{\mathcal{C}}Rec_c = \frac{1}{\mathcal{C} \mathcal{N}}\sum_{c=1}^{\mathcal{C}} \sum_{n=1}^{\mathcal{N}} \frac{TP_{n, c}}{TP_{n, c} + FN_{n, c}},
    \nonumber
    \\
    \vspace{-0.2cm}
    &mF1 = \frac{1}{\mathcal{C}}\sum_{c=1}^{\mathcal{C}}F1_c = \frac{1}{\mathcal{C}}\sum_{c=1}^{\mathcal{C}} \frac{2 * Pre_c * Rec_c}{Pre_c + Rec_c},
    \label{Eq:mF1}
\end{align}
where $TP$, $FP$, $TN$ and $FN$ stand for True Positive, False Positive, True Negative and False Negative, respectively. $\mathcal{C}$ denotes the number of semantic classes within the dataset and $\mathcal{N}$ signifies the size of the test dataset. For example, with respect to SynthiaSF dataset, $\mathcal{C}$ is 23 and $\mathcal{N}$ is 628.

\begin{table*}[tp]
\centering
\caption{The performance comparison of AdvImmu against other baselines across multiple datasets}
\setlength{\tabcolsep}{7.0pt}
\begin{tabularx}{\linewidth}{cccccccccc}
\hline
\multirow{2}{*}{Methods} & \multicolumn{4}{c}{\textbf{\emph{Apolloscape Dataset}}}                   &  & \multicolumn{4}{c}{\textbf{\emph{CARLA\_ADV Dataset}}}                    \\ \cline{2-5} \cline{7-10} 
                         & mIoU       & mPre       & mRec       & mF1        &  & mIoU       & mPre       & mRec       & mF1        \\ \hline
DeepLabv3+  & \underline{26.58$\pm$0.49} & 30.23$\pm$0.89 & \underline{31.14$\pm$0.57} & \underline{32.32$\pm$0.59} &  & \underline{36.90$\pm$1.00} & 46.96$\pm$0.97 & 42.44$\pm$0.79 & 43.78$\pm$0.93 \\
BiSeNetV2  & 22.92$\pm$0.86 & 27.85$\pm$1.20 & 27.10$\pm$1.11 & 27.12$\pm$1.57 &  & 28.80$\pm$1.93 & 34.47$\pm$2.23 & 33.46$\pm$2.19 & 33.59$\pm$2.30 \\
SegNet  & 21.01$\pm$0.51 & 24.80$\pm$1.11 & 24.96$\pm$0.54 & 24.60$\pm$0.70 &  & 31.67$\pm$2.11 & 38.45$\pm$3.09 & 36.55$\pm$2.44 & 37.15$\pm$2.70 \\
AttaNet  & 20.59$\pm$0.18 & 25.88$\pm$0.31 & 25.38$\pm$0.22 & 25.55$\pm$0.19 &  & 27.12$\pm$1.40 & 33.41$\pm$1.68 & 32.90$\pm$1.65 & 32.60$\pm$1.57 \\
BASeg  & 20.14$\pm$0.18 & 31.72$\pm$0.29 & 25.26$\pm$0.20 & 25.75$\pm$0.31 &  & 36.36$\pm$0.76 & \underline{57.16$\pm$0.43} & \underline{43.86$\pm$0.82} & \underline{45.01$\pm$0.71} \\
HRDA  & 21.55$\pm$0.27 & \underline{34.13$\pm$0.31} & 26.09$\pm$0.34 & 26.40$\pm$0.31 &  & 33.66$\pm$0.14 & 50.87$\pm$1.04 & 39.23$\pm$1.30 & 41.57$\pm$1.23 \\
SeaFormer  & 20.34$\pm$0.16 & 25.33$\pm$0.21 & 25.03$\pm$0.13 & 24.26$\pm$0.16 &  & 27.85$\pm$0.64 & 35.80$\pm$0.55 & 32.58$\pm$0.58 & 33.48$\pm$0.56 \\
TopFormer  & 20.41$\pm$0.00 & 25.19$\pm$0.45 & 25.20$\pm$0.22 & 24.28$\pm$0.20 &  & 29.84$\pm$1.73 & 38.09$\pm$2.72 & 34.29$\pm$1.75 & 35.41$\pm$2.08 \\
\textbf{AdvImmu (Ours)} &  \textbf{59.35$\pm$0.57} & \textbf{81.38$\pm$2.22} & \textbf{60.76$\pm$0.49} & \textbf{65.28$\pm$0.55} &  & \textbf{69.58$\pm$1.79} & \textbf{85.71$\pm$2.01} & \textbf{70.78$\pm$1.56} & \textbf{75.01$\pm$1.56} \\ \hline \hline
\multirow{2}{*}{Methods} & \multicolumn{4}{c}{\textbf{\emph{Shift Dataset}}}                   &  & \multicolumn{4}{c}{\textbf{\emph{SynthiaSF Dataset}}}                    \\ \cline{2-5} \cline{7-10} 
                         & mIoU       & mPre       & mRec       & mF1        &  & mIoU       & mPre       & mRec       & mF1        \\ \hline
DeepLabv3+  & 15.51$\pm$0.26 & \underline{23.17$\pm$0.20} & 16.84$\pm$0.29 & 18.68$\pm$0.25 &  & 31.80$\pm$1.11 & 39.76$\pm$1.79 & 35.68$\pm$0.85 & 36.64$\pm$1.17 \\
BiSeNetV2  & 10.74$\pm$2.50 & 15.23$\pm$5.01 & 11.88$\pm$2.81 & 12.63$\pm$3.38 &  & 27.04$\pm$0.68 & 33.49$\pm$0.93 & 31.42$\pm$0.64 & 31.49$\pm$0.70 \\
SegNet  & 7.94$\pm$0.31 & 8.41$\pm$0.50 & 8.71$\pm$0.41 & 8.55$\pm$0.44 &  & 23.59$\pm$2.74 & 30.10$\pm$2.88 & 28.15$\pm$2.52 & 28.13$\pm$2.63 \\
AttaNet  & 14.24$\pm$0.98 & 21.19$\pm$1.62 & 15.73$\pm$0.96 & 17.34$\pm$1.15 &  & 24.87$\pm$1.00 & 32.13$\pm$1.86 & 29.94$\pm$0.78 & 29.75$\pm$0.91 \\
BASeg  & \underline{16.60$\pm$0.18} & 22.93$\pm$0.30 & \underline{17.87$\pm$0.18} & \underline{19.91$\pm$0.18} &  & \underline{33.70$\pm$0.54} & \underline{54.05$\pm$0.44} & \underline{36.61$\pm$0.57} & \underline{40.48$\pm$0.47} \\
HRDA  & 13.51$\pm$0.46 & 21.60$\pm$0.40 & 14.45$\pm$0.53 & 16.51$\pm$0.51 &  & 27.13$\pm$1.81 & 46.80$\pm$2.47 & 32.56$\pm$1.51 & 33.95$\pm$1.74 \\
SeaFormer  & 7.69$\pm$0.08 & 8.26$\pm$0.03 & 8.43$\pm$0.09 & 8.33$\pm$0.05 &  & 26.01$\pm$0.87 & 33.23$\pm$0.69 & 30.30$\pm$0.76 & 30.64$\pm$0.71 \\
TopFormer  & 13.53$\pm$1.52 & 20.00$\pm$2.83 & 15.01$\pm$1.69 & 16.44$\pm$2.04 &  & 25.45$\pm$0.76 & 32.35$\pm$0.60 & 30.20$\pm$0.58 & 30.31$\pm$0.63 \\
\textbf{AdvImmu (Ours)} &  \textbf{26.09$\pm$1.54} & \textbf{29.35$\pm$2.06} & \textbf{26.54$\pm$1.20} & \textbf{26.47$\pm$1.21} &  & \textbf{51.30$\pm$4.36} & \textbf{74.91$\pm$2.46} & \textbf{52.19$\pm$4.01} & \textbf{55.65$\pm$3.79} \\ \hline \hline
\multirow{2}{*}{Methods} & \multicolumn{4}{c}{\textbf{\emph{Cityscapes Dataset}}}                   &  & \multicolumn{4}{c}{\textbf{\emph{GTA5 Dataset}}}                    \\ \cline{2-5} \cline{7-10} 
                         & mIoU       & mPre       & mRec       & mF1        &  & mIoU       & mPre       & mRec       & mF1        \\ \hline
DeepLabv3+  & 30.37$\pm$0.62 & 34.95$\pm$0.84 & 41.78$\pm$0.32 & 37.18$\pm$0.56 &  & 25.51$\pm$1.94 & 28.17$\pm$2.54 & 39.07$\pm$2.45 & 31.62$\pm$2.52 \\
BiSeNetV2  & 19.33$\pm$1.20 & 20.61$\pm$1.66 & 25.47$\pm$1.30 & 22.34$\pm$1.52 &  & 16.33$\pm$0.43 & 17.29$\pm$0.43 & 28.10$\pm$0.26 & 20.30$\pm$0.42 \\
SegNet  & 18.66$\pm$1.06 & 20.06$\pm$1.74 & 24.82$\pm$0.98 & 21.63$\pm$1.34 &  & 15.81$\pm$0.71 & 17.05$\pm$0.81 & 26.66$\pm$0.43 & 19.74$\pm$0.77 \\
AttaNet  & 20.18$\pm$1.05 & 22.47$\pm$1.34 & 27.74$\pm$1.77 & 24.13$\pm$1.37 &  & 16.93$\pm$1.18 & 18.43$\pm$1.47 & 28.16$\pm$1.39 & 21.24$\pm$1.54 \\
BASeg  & \underline{64.56$\pm$0.60} & \underline{82.29$\pm$0.76} & \underline{73.15$\pm$0.62} & \underline{76.44$\pm$0.60} &  & \underline{58.09$\pm$0.76} & \textbf{79.08$\pm$0.57} & \underline{64.42$\pm$0.61} & \textbf{68.72$\pm$0.66} \\
HRDA  & 36.59$\pm$1.63 & 60.42$\pm$1.97 & 43.80$\pm$1.73 & 46.72$\pm$1.97 &  & 36.83$\pm$2.36 & 61.20$\pm$2.05 & 44.04$\pm$2.25 & 47.20$\pm$2.65 \\
SeaFormer  & 22.53$\pm$0.42 & 25.59$\pm$0.45 & 30.19$\pm$0.39 & 26.90$\pm$0..38 &  & 17.84$\pm$1.22 & 19.35$\pm$1.48 & 29.54$\pm$1.55 & 22.21$\pm$1.58 \\
TopFormer  & 20.81$\pm$1.06 & 23.07$\pm$1.32 & 28.50$\pm$1.77 & 24.88$\pm$1.46 &  & 19.33$\pm$1.42 & 21.47$\pm$1.81 & 31.29$\pm$1.75 & 24.64$\pm$1.91 \\
\textbf{AdvImmu (Ours)} &  \textbf{71.14$\pm$3.70} & \textbf{88.64$\pm$4.12} & \textbf{79.69$\pm$3.84} & \textbf{79.17$\pm$2.88} &  & \textbf{58.11$\pm$2.47} & \underline{78.89$\pm$2.16} & \textbf{72.15$\pm$2.00} & \underline{67.58$\pm$2.36} \vspace{0.02cm} \\ \hline
\end{tabularx}
\label{Tab.perf_comp}
\end{table*}

\begin{table*}[tp]
\centering
\renewcommand{\arraystretch}{0.24}
\addtolength{\tabcolsep}{-0.45pt}
\caption{Qualitative performance of AdvImmu against other SOTA baselines under various adverse weather conditions}
\begin{tabularx}{\linewidth}{|llllll|} 
\hline
\hspace{0.6cm}Raw Images &\hspace{0.3cm}Ground Truth &\hspace{0.6cm}BASeg &\hspace{0.6cm}HRDA &\hspace{0.3cm}DeepLabv3+ &\hspace{0.1cm}\textbf{AdvImmu (Ours)} \\
\hline
\includegraphics[width=0.162\linewidth, height=0.11\linewidth]{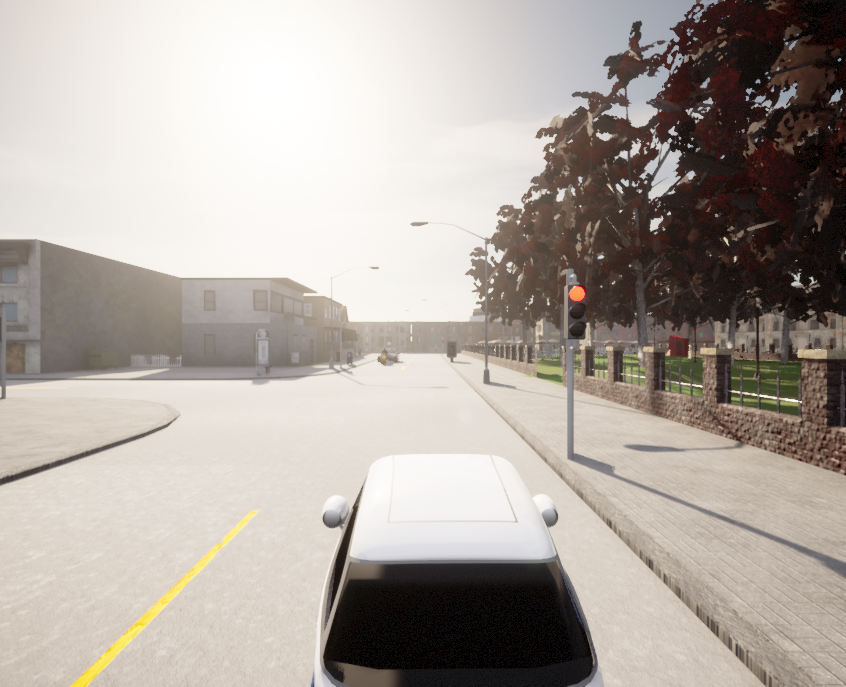} &\hspace{-0.47cm}
\includegraphics[width=0.162\linewidth, height=0.11\linewidth]{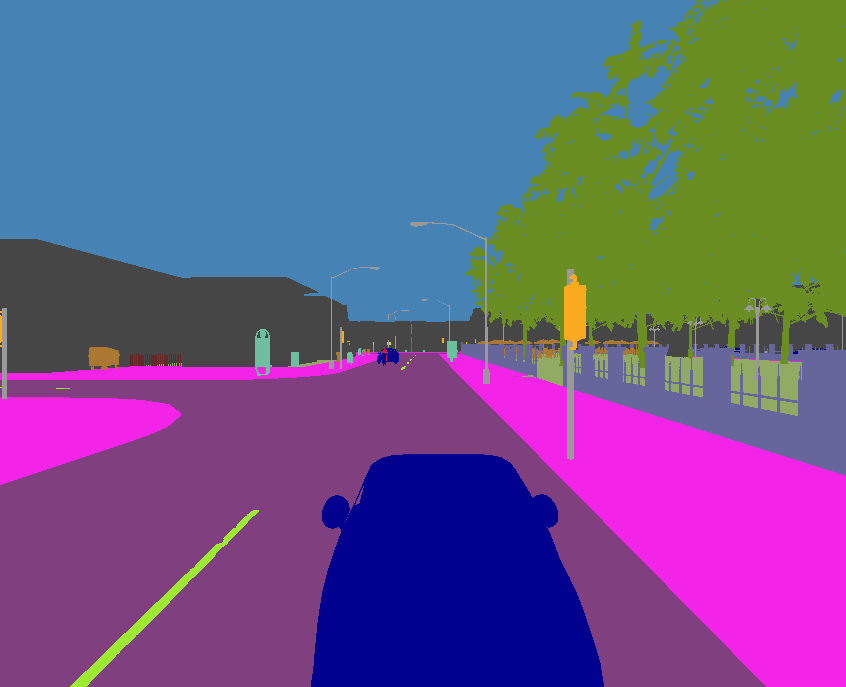} &\hspace{-0.47cm}
\includegraphics[width=0.162\linewidth, height=0.11\linewidth]{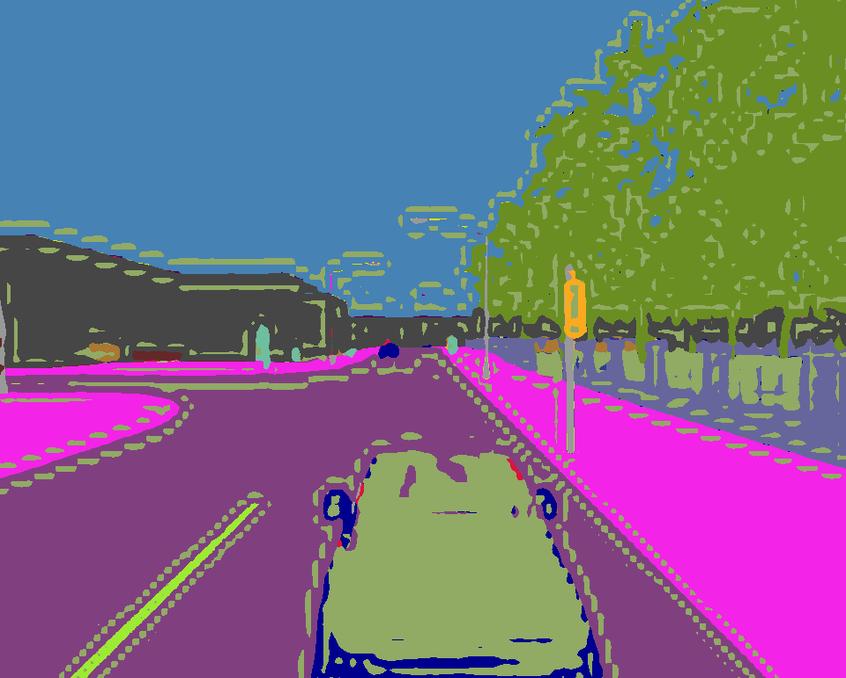} &\hspace{-0.47cm}
\includegraphics[width=0.162\linewidth, height=0.11\linewidth]{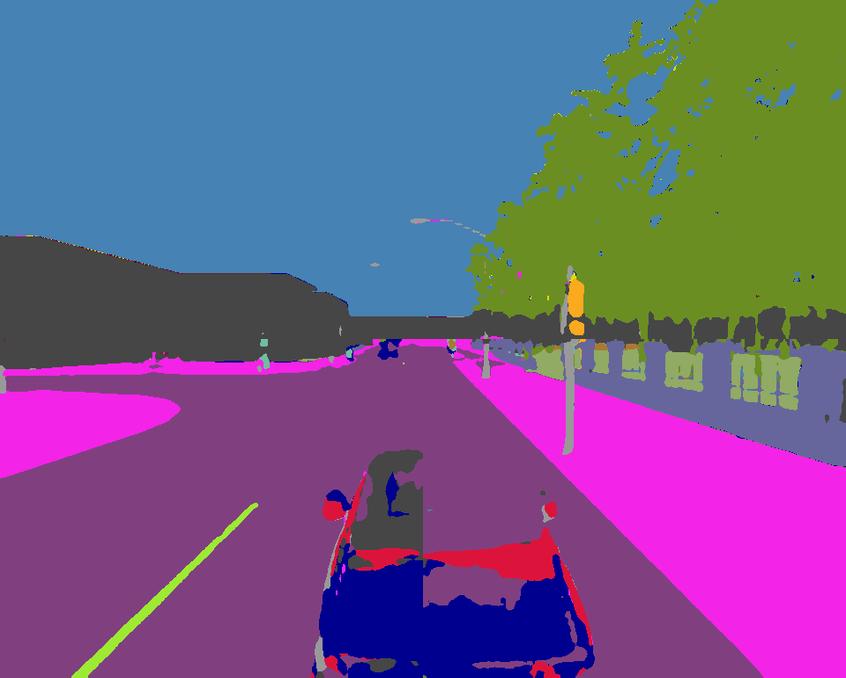} &\hspace{-0.47cm}
\includegraphics[width=0.162\linewidth, height=0.11\linewidth]{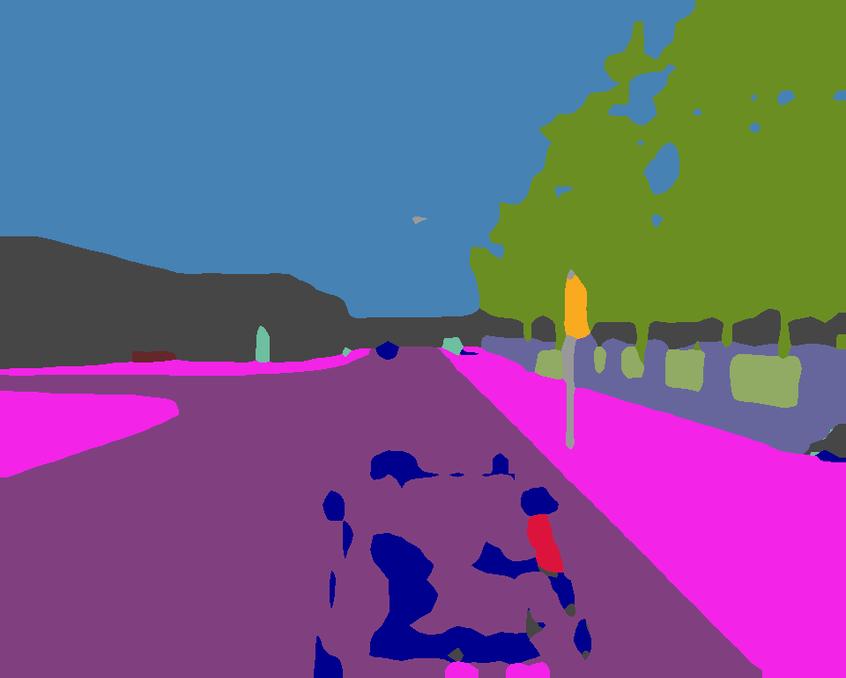} &\hspace{-0.47cm}
\includegraphics[width=0.162\linewidth, height=0.11\linewidth]{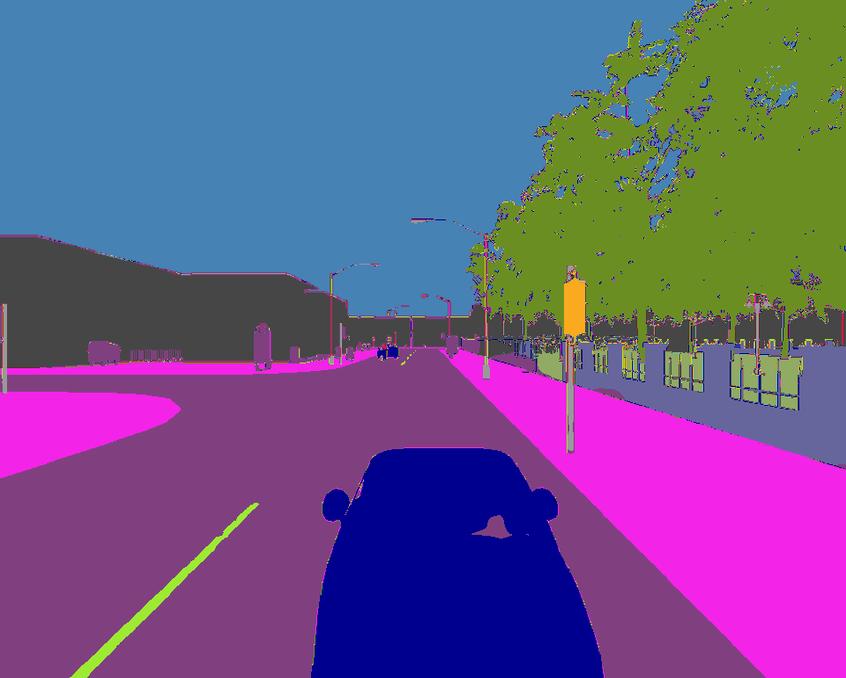}\\
\hline

\includegraphics[width=0.162\linewidth, height=0.11\linewidth]{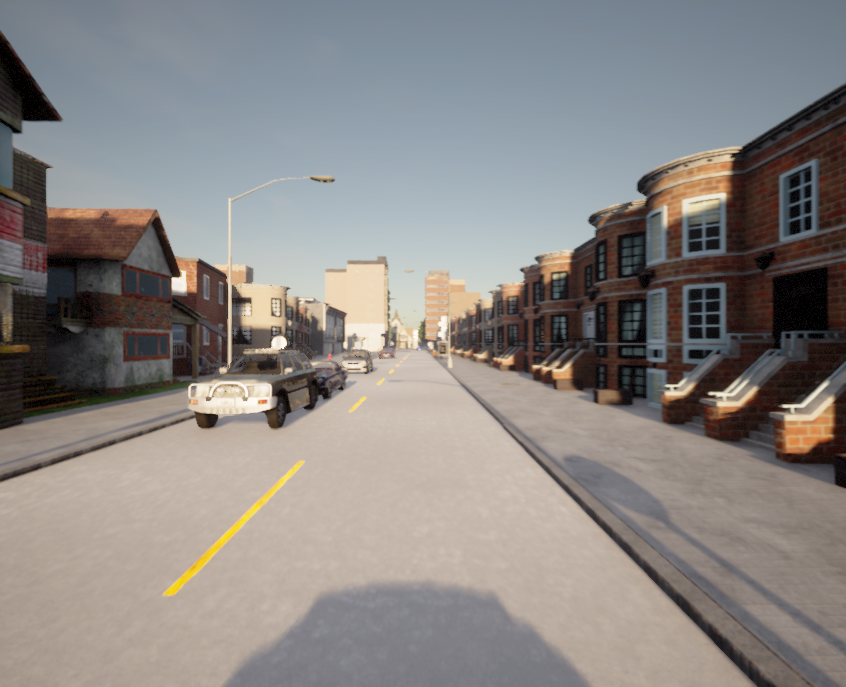} &\hspace{-0.47cm}
\includegraphics[width=0.162\linewidth, height=0.11\linewidth]{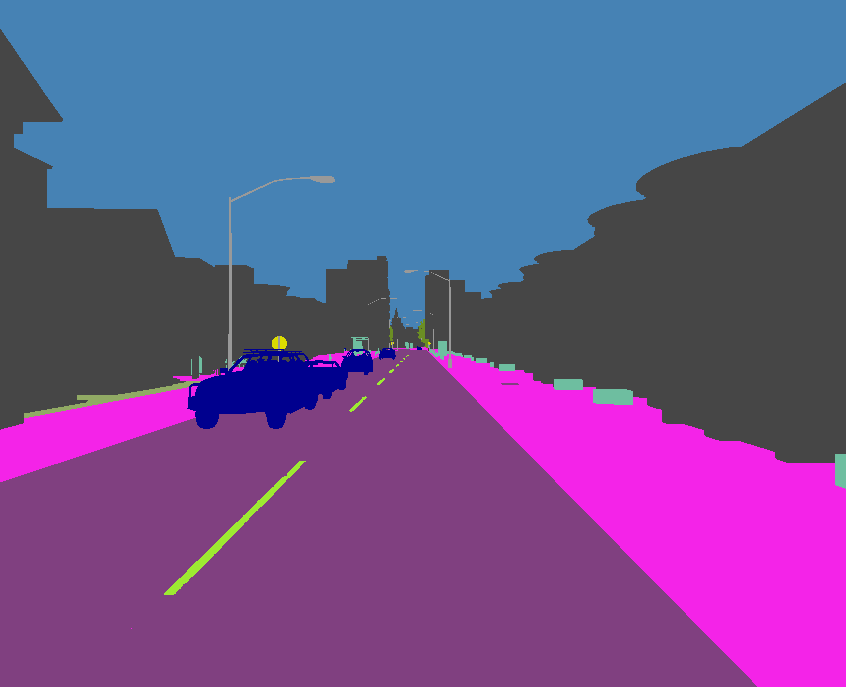} &\hspace{-0.47cm}
\includegraphics[width=0.162\linewidth, height=0.11\linewidth]{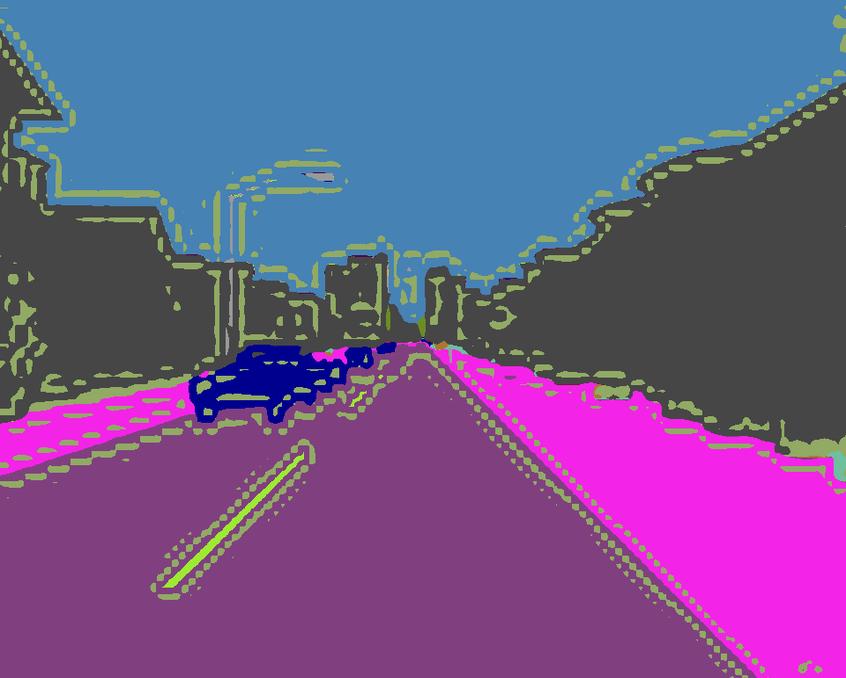} &\hspace{-0.47cm}
\includegraphics[width=0.162\linewidth, height=0.11\linewidth]{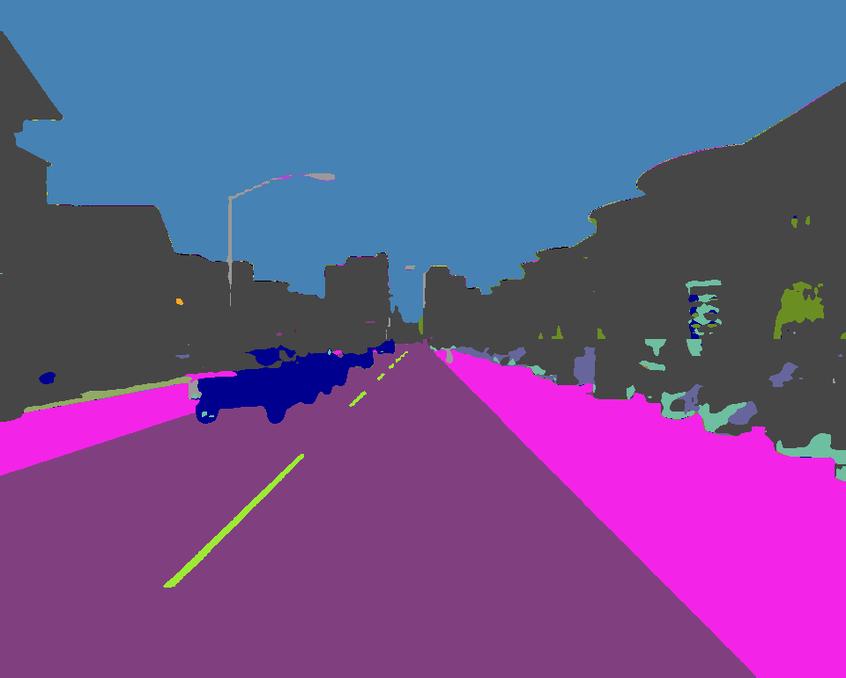} &\hspace{-0.47cm}
\includegraphics[width=0.162\linewidth, height=0.11\linewidth]{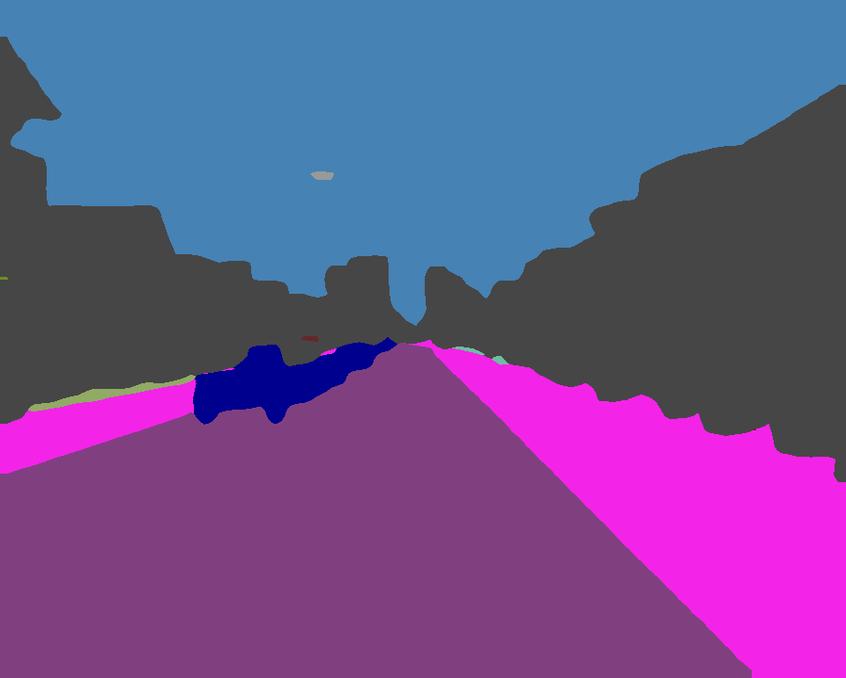} &\hspace{-0.47cm}
\includegraphics[width=0.162\linewidth, height=0.11\linewidth]{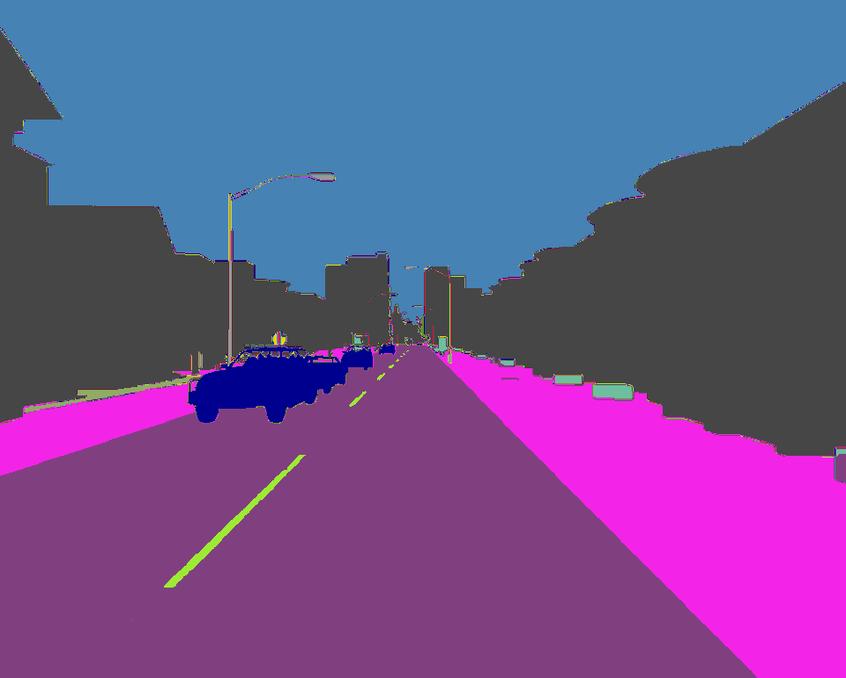}\\
\hline

\includegraphics[width=0.162\linewidth, height=0.11\linewidth]{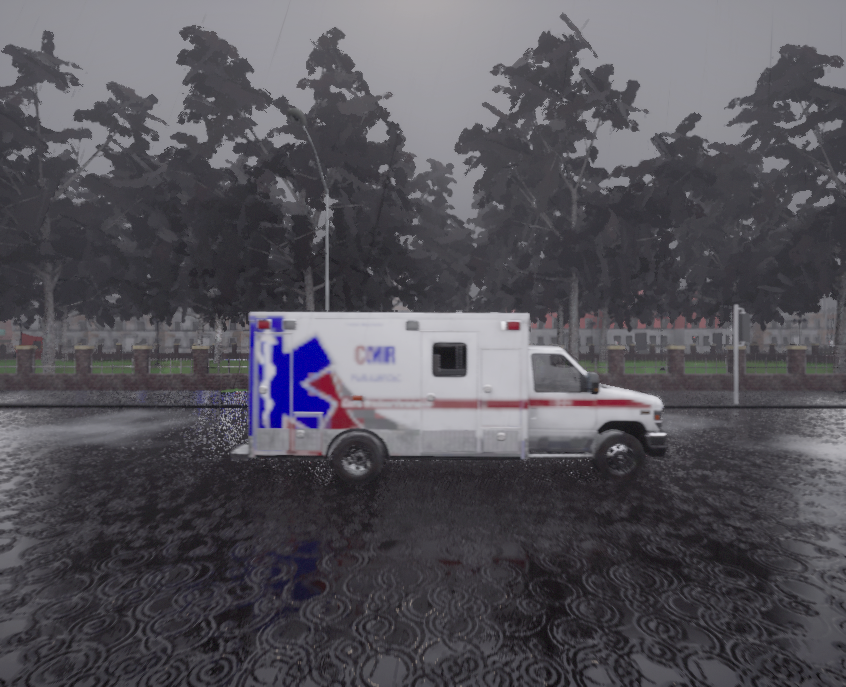} &\hspace{-0.47cm}
\includegraphics[width=0.162\linewidth, height=0.11\linewidth]{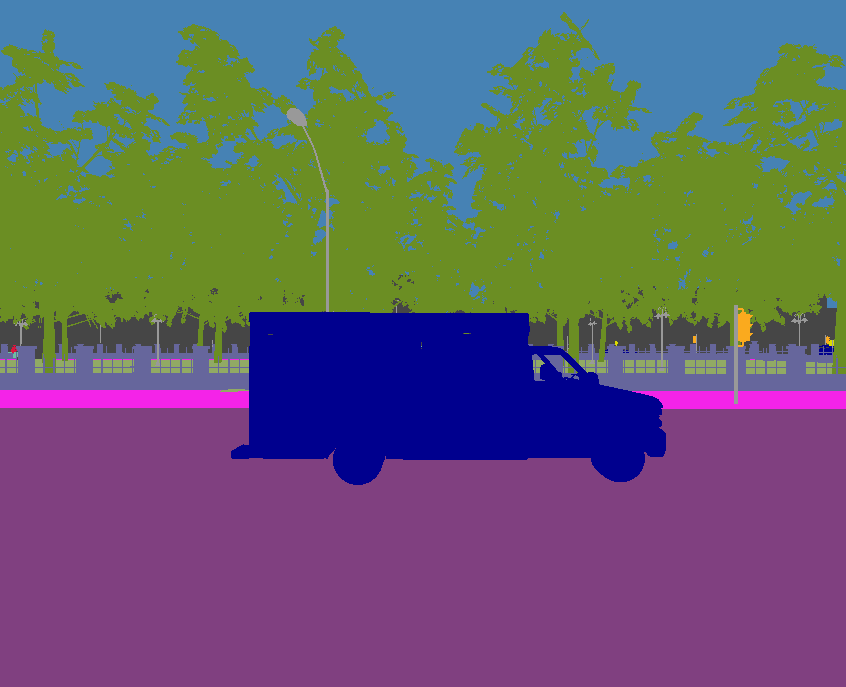} &\hspace{-0.47cm}
\includegraphics[width=0.162\linewidth, height=0.11\linewidth]{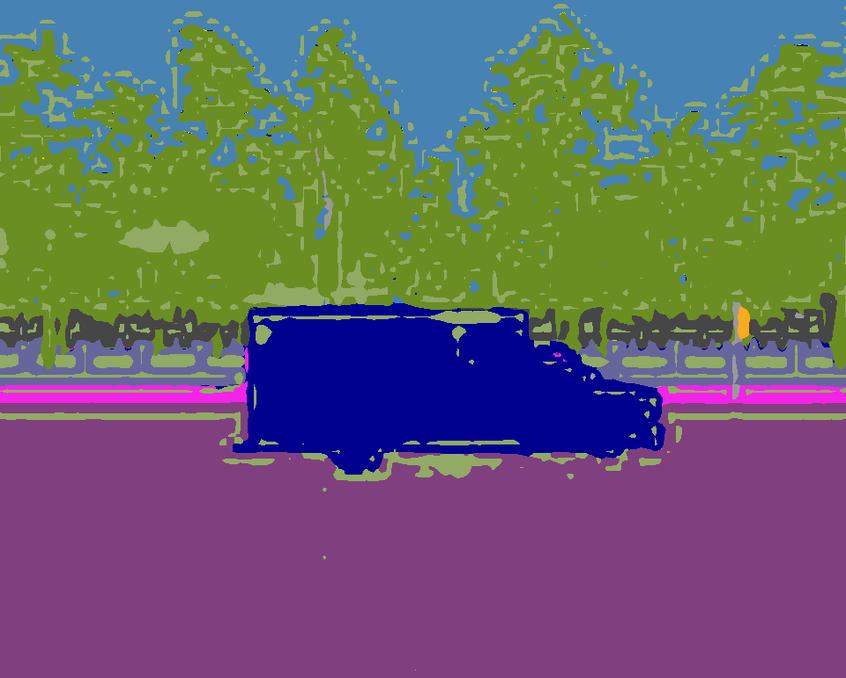} &\hspace{-0.47cm}
\includegraphics[width=0.162\linewidth, height=0.11\linewidth]{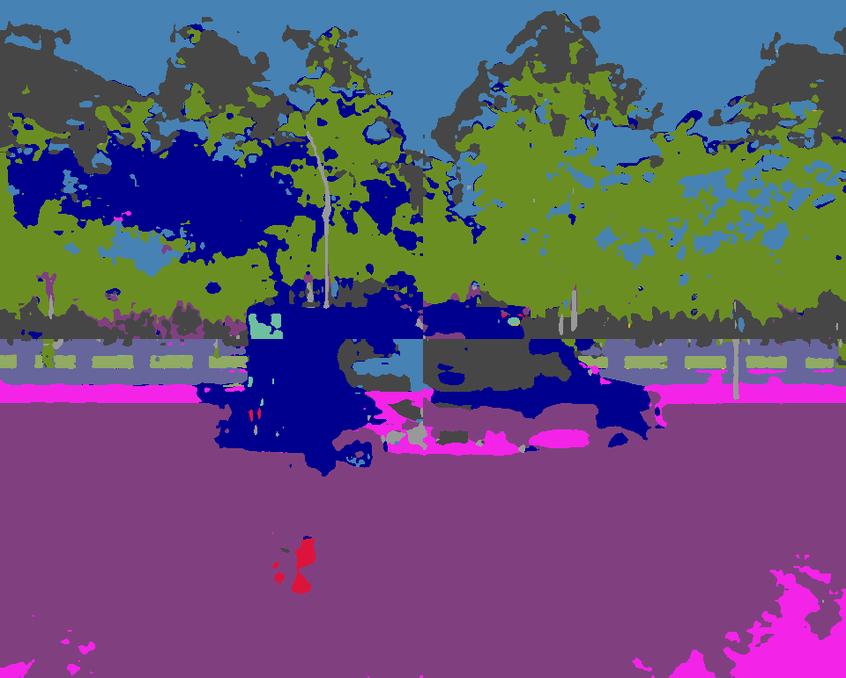} &\hspace{-0.47cm}
\includegraphics[width=0.162\linewidth, height=0.11\linewidth]{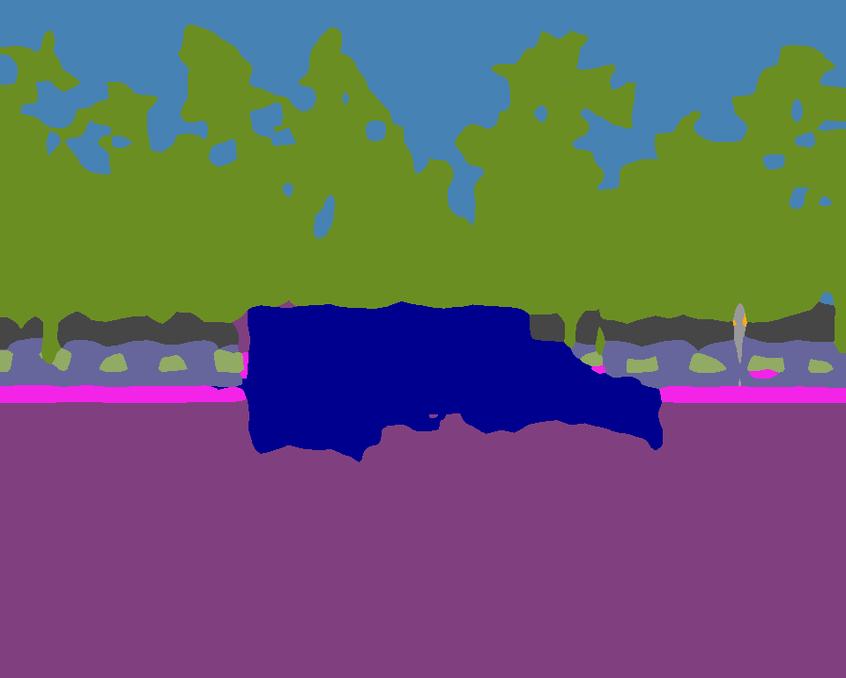} &\hspace{-0.47cm}
\includegraphics[width=0.162\linewidth, height=0.11\linewidth]{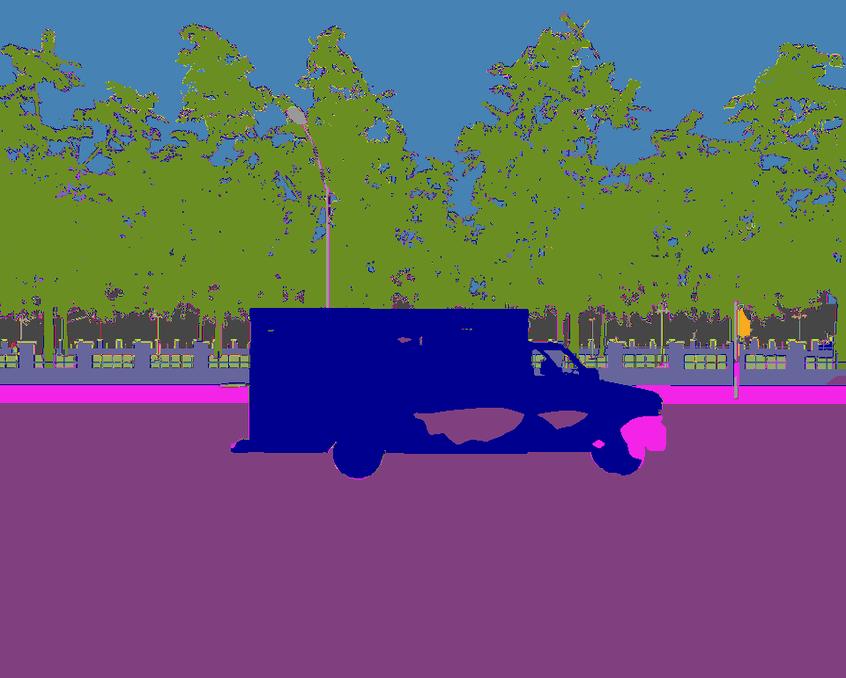}\\
\hline

\includegraphics[width=0.162\linewidth, height=0.11\linewidth]{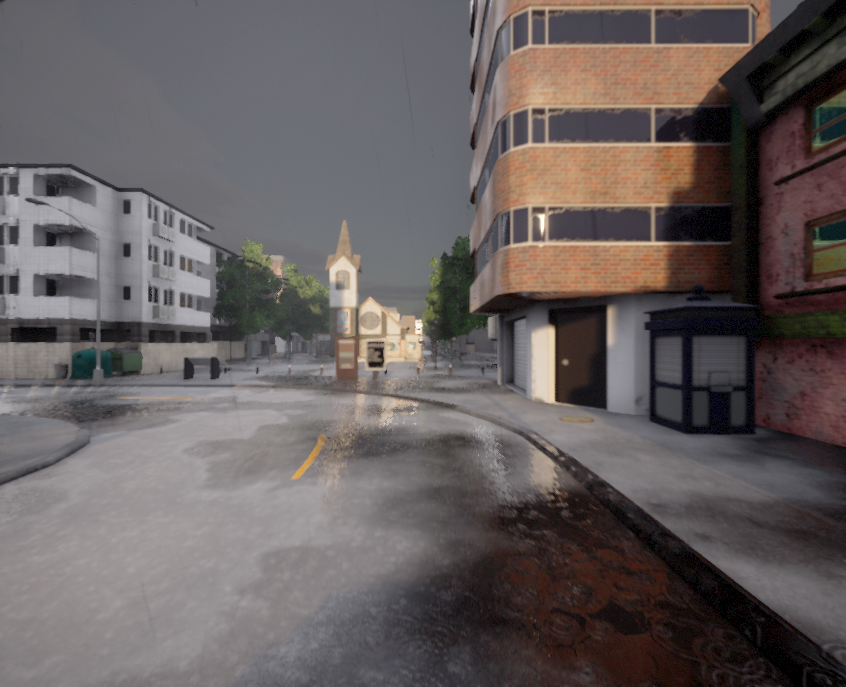} &\hspace{-0.47cm}
\includegraphics[width=0.162\linewidth, height=0.11\linewidth]{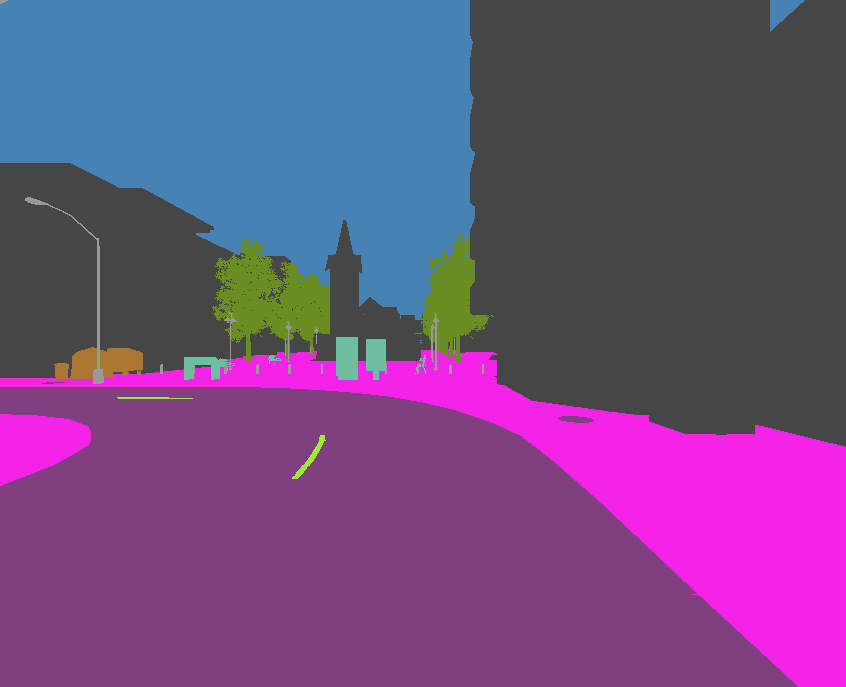} &\hspace{-0.47cm}
\includegraphics[width=0.162\linewidth, height=0.11\linewidth]{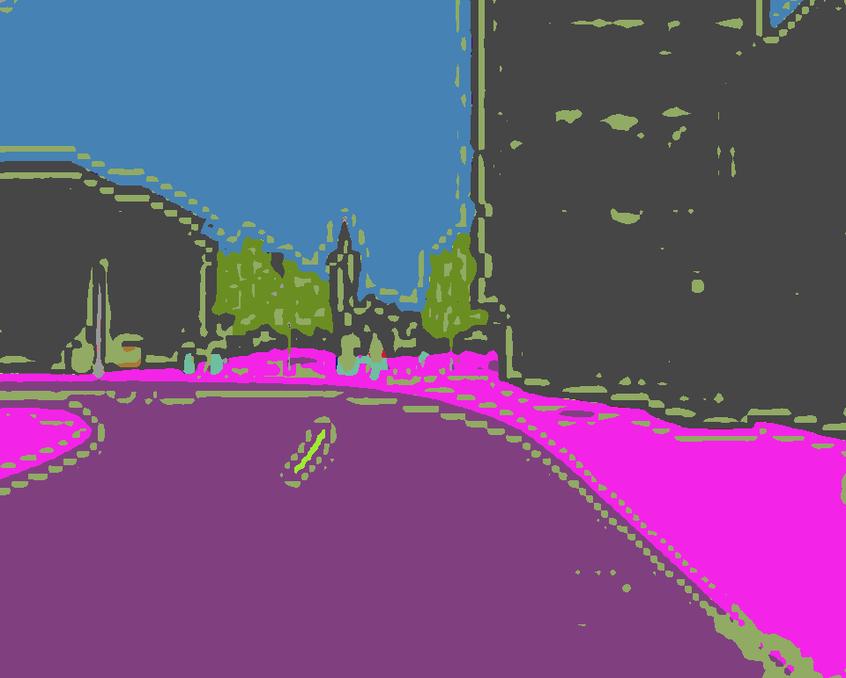} &\hspace{-0.47cm}
\includegraphics[width=0.162\linewidth, height=0.11\linewidth]{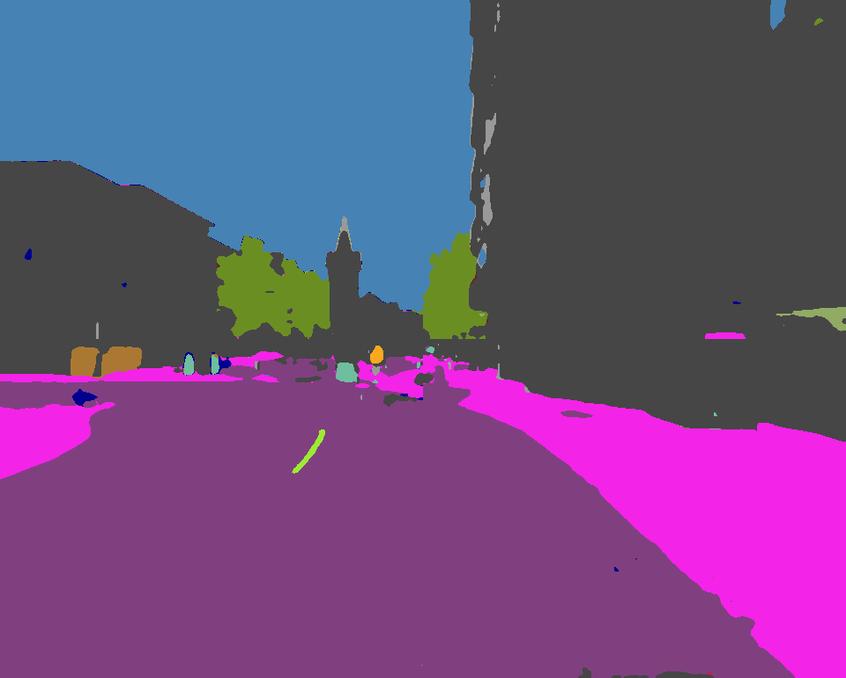} &\hspace{-0.47cm}
\includegraphics[width=0.162\linewidth, height=0.11\linewidth]{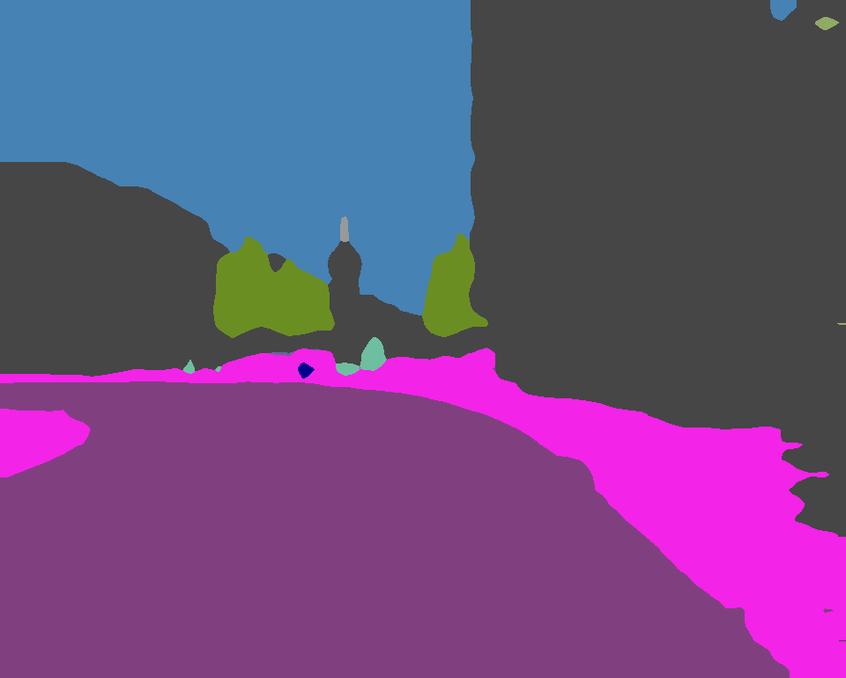} &\hspace{-0.47cm}
\includegraphics[width=0.162\linewidth, height=0.11\linewidth]{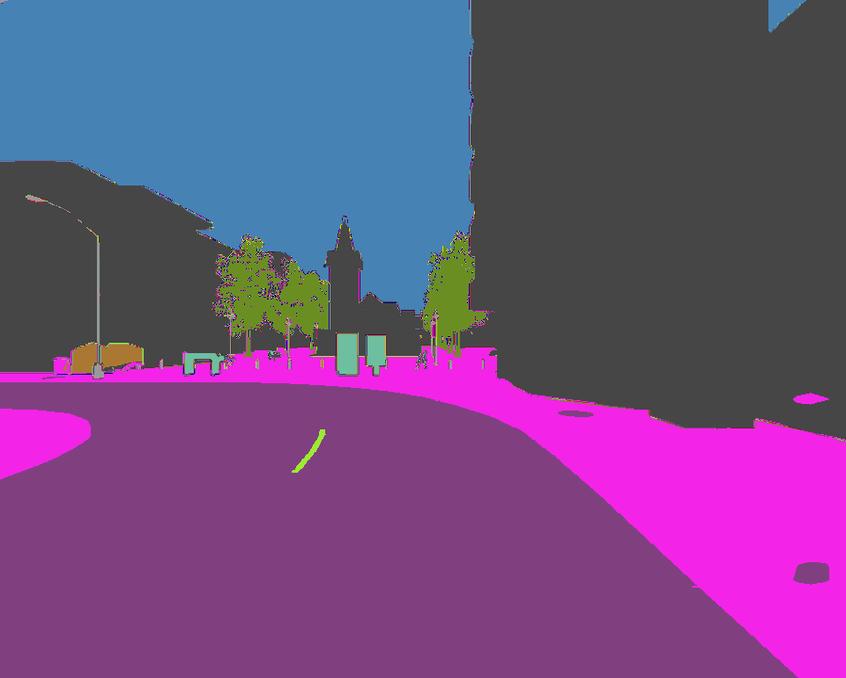}\\
\hline

\includegraphics[width=0.162\linewidth, height=0.11\linewidth]{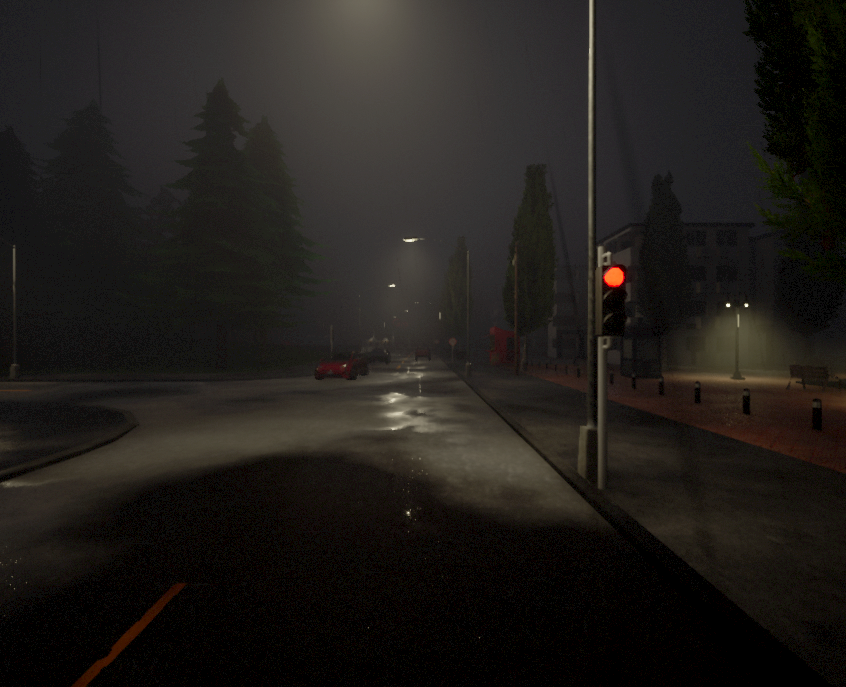} &\hspace{-0.47cm}
\includegraphics[width=0.162\linewidth, height=0.11\linewidth]{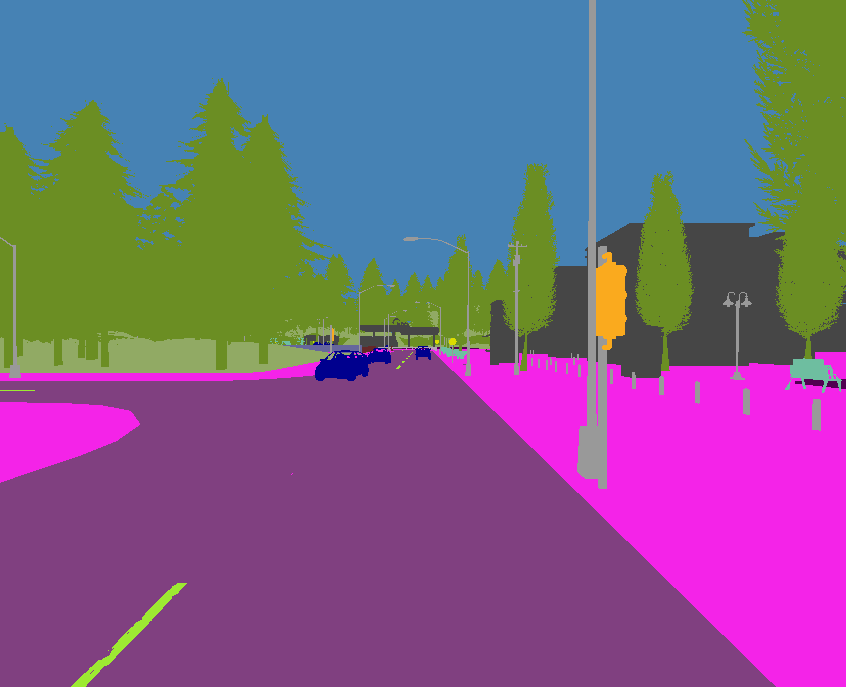} &\hspace{-0.47cm}
\includegraphics[width=0.162\linewidth, height=0.11\linewidth]{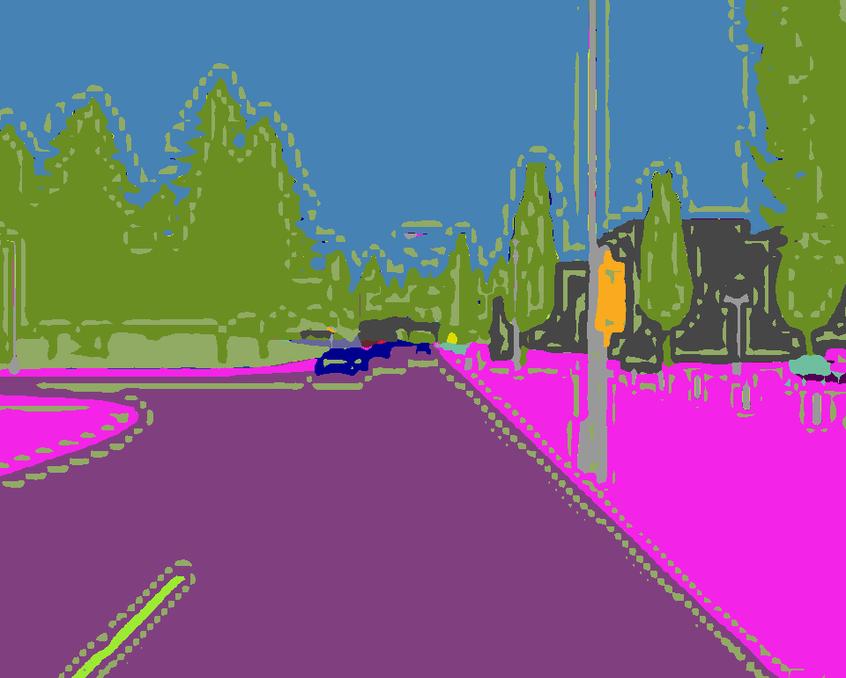} &\hspace{-0.47cm}
\includegraphics[width=0.162\linewidth, height=0.11\linewidth]{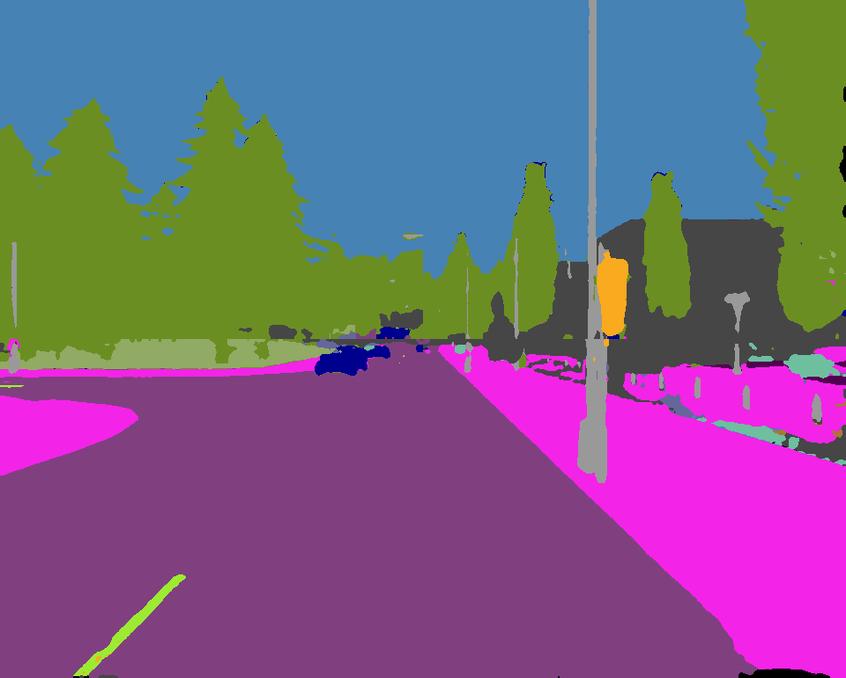} &\hspace{-0.47cm}
\includegraphics[width=0.162\linewidth, height=0.11\linewidth]{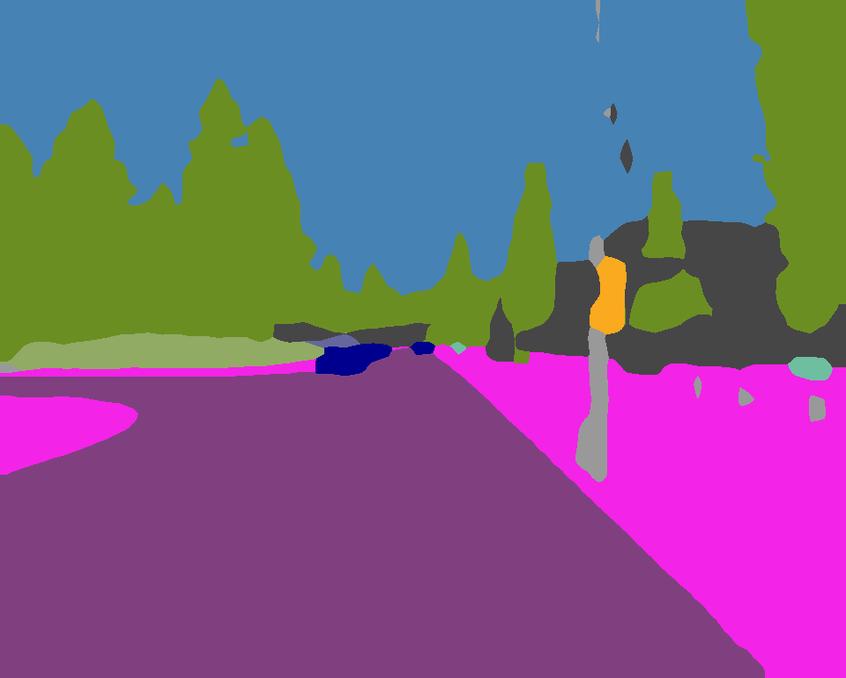} &\hspace{-0.47cm}
\includegraphics[width=0.162\linewidth, height=0.11\linewidth]{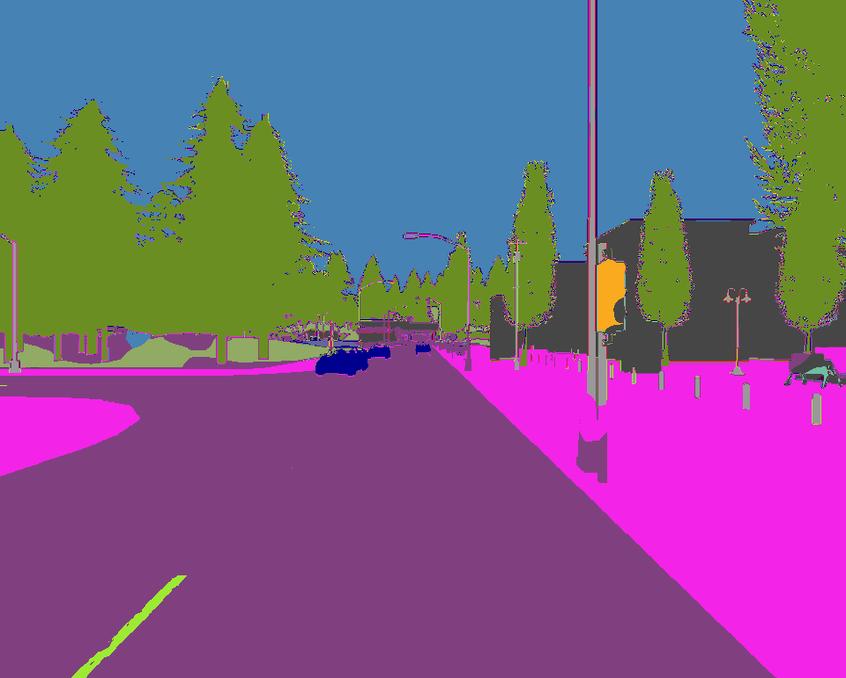}\\
\hline

\includegraphics[width=0.162\linewidth, height=0.11\linewidth]{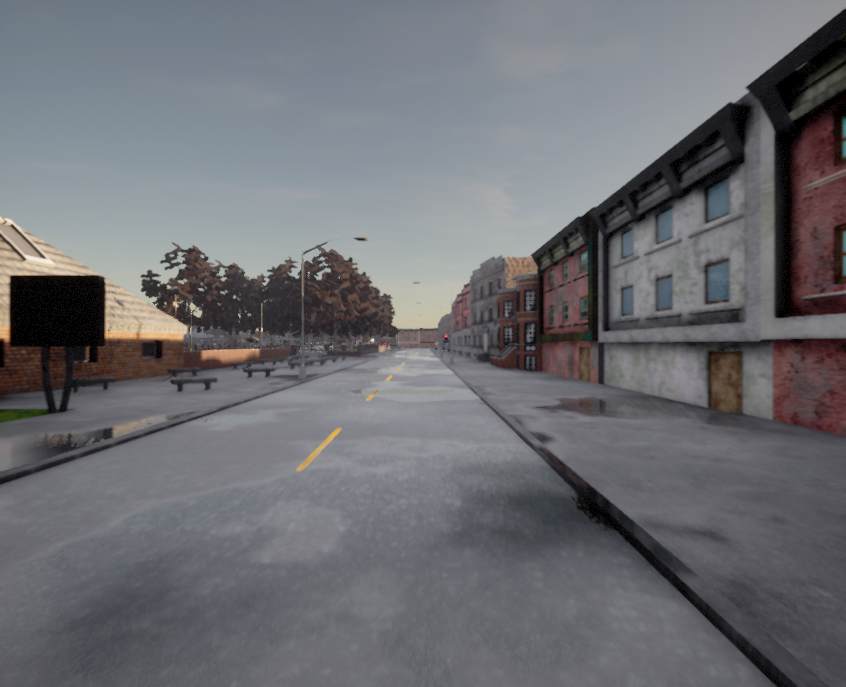} &\hspace{-0.47cm}
\includegraphics[width=0.162\linewidth, height=0.11\linewidth]{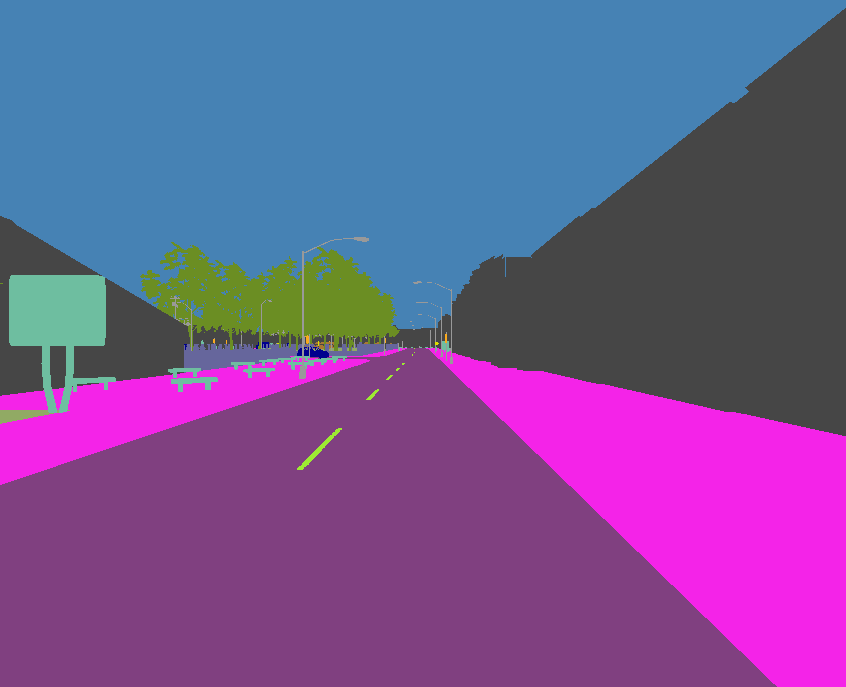} &\hspace{-0.47cm}
\includegraphics[width=0.162\linewidth, height=0.11\linewidth]{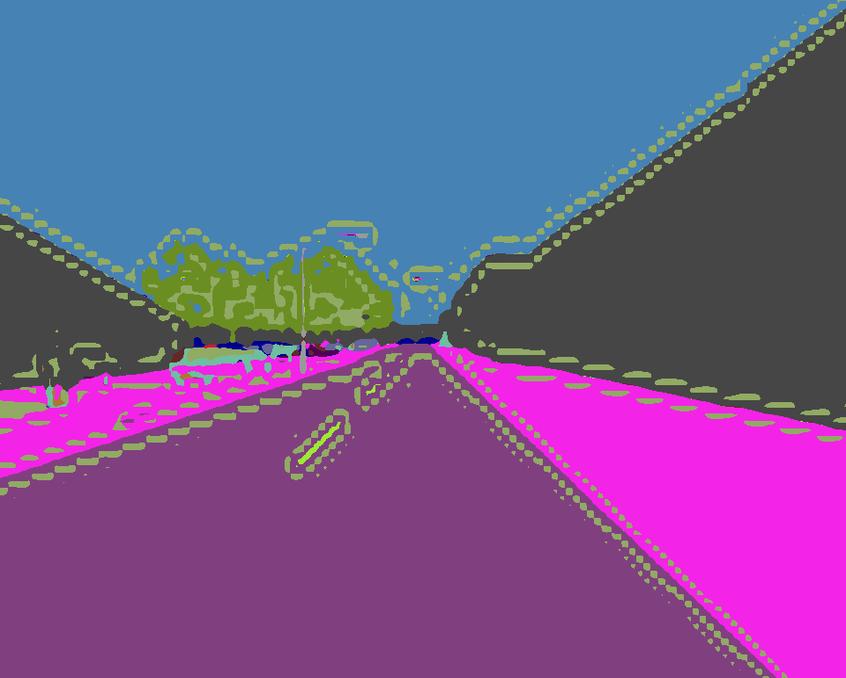} &\hspace{-0.47cm}
\includegraphics[width=0.162\linewidth, height=0.11\linewidth]{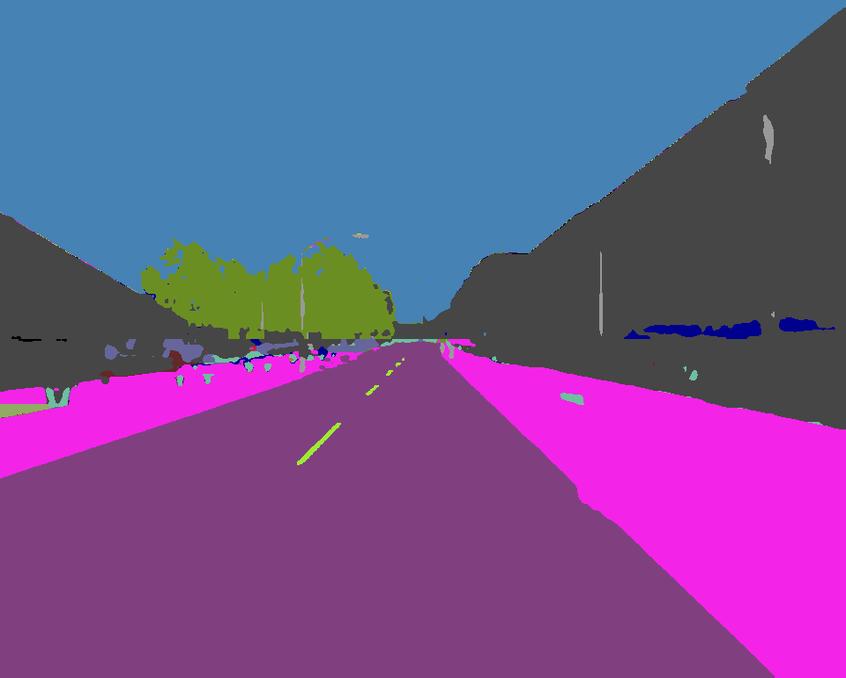} &\hspace{-0.47cm}
\includegraphics[width=0.162\linewidth, height=0.11\linewidth]{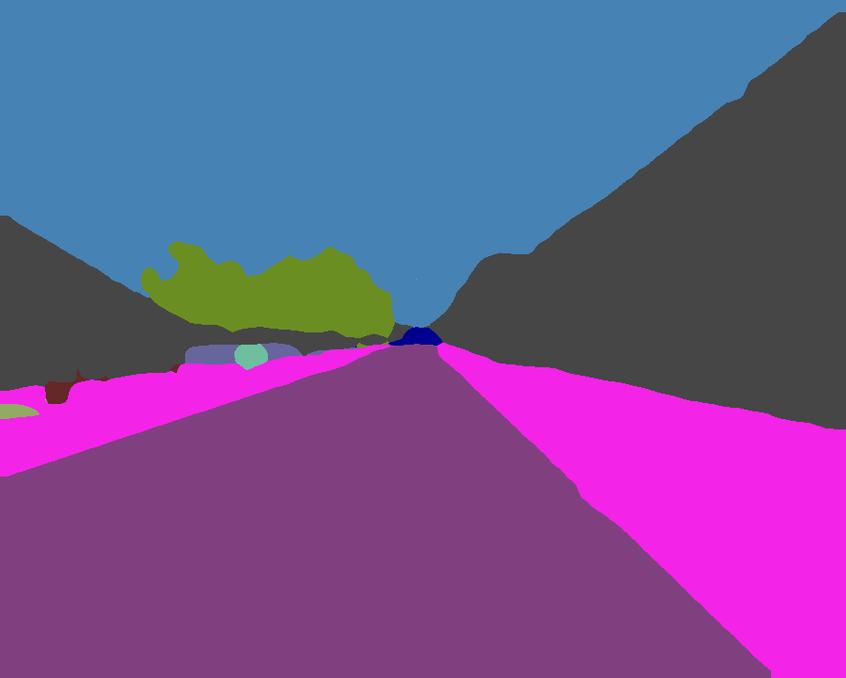} &\hspace{-0.47cm}
\includegraphics[width=0.162\linewidth, height=0.11\linewidth]{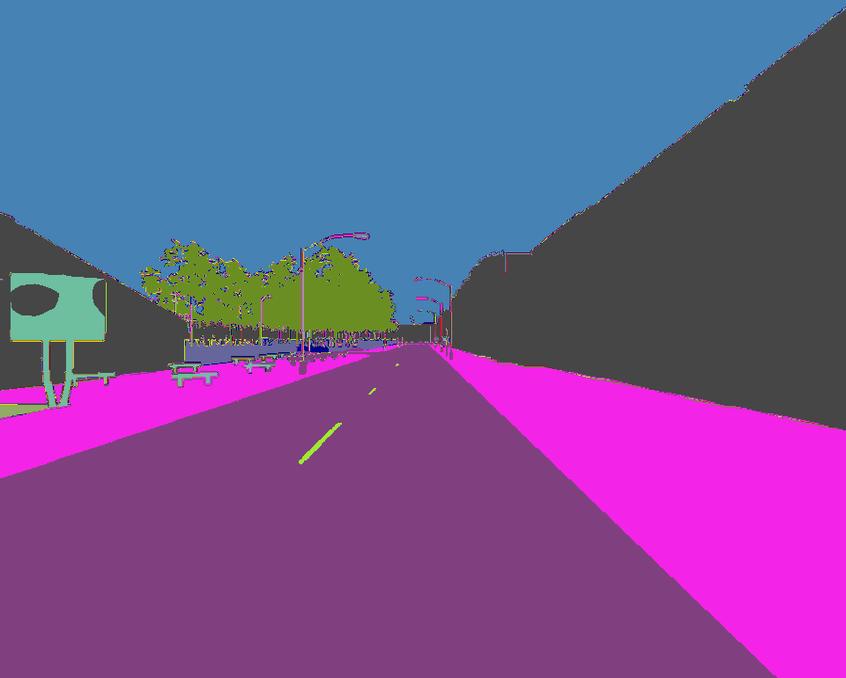}\\
\hline
\end{tabularx}
\label{tab:semantic_pred}
\vspace{-0.4cm}
\end{table*}

\subsubsection{Implementation Details}
The proposed AdvImmu model is implemented using the Pytorch framework, with experiments conducted on two NVIDIA GeForce 4090 GPUs. For the architecture, we implement the proposed AdvImmu based on ResNet \cite{He_2016_CVPR} and ASSP perception head \cite{chen2018encoderdecoder}. We compare the proposed AdvImmu with other 8 baselines, \ie, DeepLabv3+ \cite{chen2018encoderdecoder}, BiSeNetV2 \cite{yu2021bisenet}, SegNet \cite{badrinarayanan2017segnet}, TopFormer \cite{zhang2022topformer}, SeaFormer \cite{wan2023seaformer}, BASeg \cite{xiao2023baseg}, HRDA \cite{hoyer2023domain}, and AttaNet \cite{song2021attanet}. All these baselines are also implemented using Pytorch. We train the proposed AdvImmu and such baselines based on aforementioned datasets from scratch. For the optimization, we select the Adam optimizer for training, configuring it with Betas values of 0.9 and 0.999 and setting the weight decay to 1e-4. The training is executed with a batch size of 8 and a learning rate of 3e-4. For the InfoNCE regularizer, the hyperparameter $\tau$ is set to 0.5. For URs, the number of unfolded layers $K$ is set to 5.

\subsection{AdvImmu's Overall Evaluation}
\label{AdvImmu_evaluation}
In this section, we will conduct extensive expriments to explore the proposed AdvImmu from following aspects: performance, convergence, compleixty (measured by Floating Point Operations (FLOPs)), and inference delay (represented by Frames Per Second (FPS)).

\subsubsection{AdvImmu's Performance}
\Cref{Tab.perf_comp} compares the performance of the AdvImmu against DeepLabv3+, BiSeNetV2, SegNet, AttaNet, BASeg, HRDA, SeaFormer and TopFormer, across Apolloscapes, CARLA\_ADV, Shift, SynthiaSF, Cityscapes and GTA5 datasets quantitatively. Notably, in this table, bold font denotes the best performance model and underline denotes the second best performance model with respect to evaluation metrics. From \Cref{Tab.perf_comp}, we can observe that for almost all datasets, AdvImmu generally outperforms all other baselines across almost all metrics. Specifically, taking Apolloscapes dataset as example, AdvImmue outperforms DeepLabv3+, BiSeNetV2, SegNet, AttaNet, BASeg, HRDA, SeaFormer and TopFormer by (59.35 - 26.58) / 26.58 = 123.29\%, (59.35 - 22.92) / 22.92 = 158.94\%, (59.35 - 21.01) / 21.01 = 182.48\%, (59.35 - 20.59) / 20.59 = 188.25\%, (59.35 - 20.14) / 20.14 = 194.69\%, (59.35 - 21.55) / 21.55 = 175.41\%, (59.35 - 20.34) / 20.34 = 191.79\%, and (59.35 - 20.41) / 20.41 = 190.79\% in mIoU, respectively. This suggests that AdvImmu showcases significant performance improvement compared to other SOTA models.

\Cref{tab:semantic_pred} compares the inference performance of the proposed AdvImmu against other models qualitatively under various adverse weather conditions, such rainy, dark, foggy, cloudy, etc. From \Cref{tab:semantic_pred}, we can observe following patterns: \textbf{(I)} The proposed AdvImmu achieves the best performance across various adverse weather conditions and its prediction can approximately approach to the ground truth. For example, under the rainy condition (the third row), AdvImmu's predition is quite similar to the ground truth, while others' prediction have various weaknesses, such as class confusion (\eg, HRDA), boundary uncertainty (\eg, DeepLabv3+), small object errors (\eg, BASeg), etc. \textbf{(II)} Some models (\eg, HRDA) can predict well in clear weather (\eg, the first row), however, suffers from a significant performance decline under adverse conditions (\eg, the third row). 

\begin{figure*}[tp]
\centering 
\subfigure[Raw Image]{
\label{Fig.baseg_2774}
\includegraphics[width=0.32\linewidth,height=0.2\linewidth]{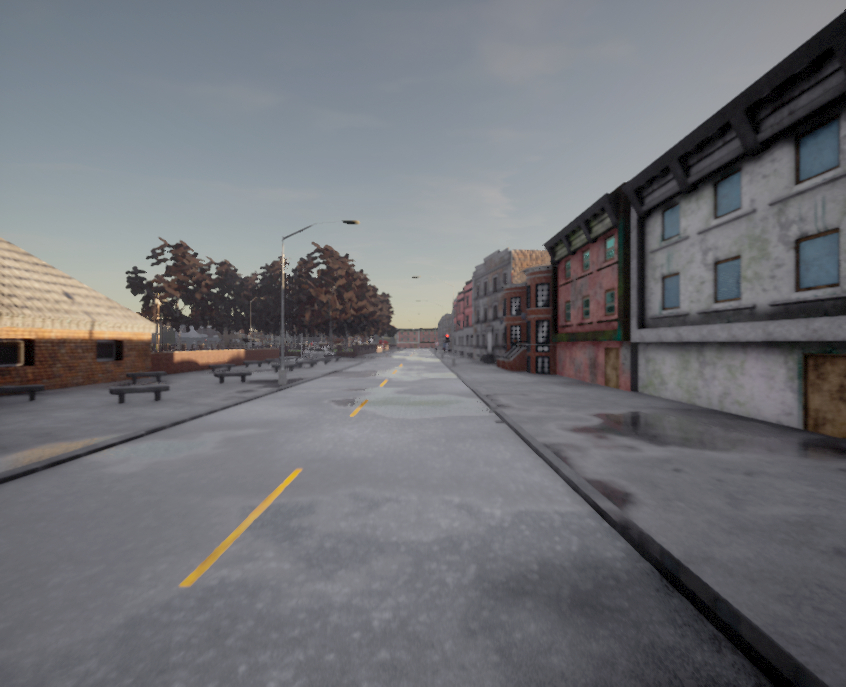}}
\hspace{-0.3cm}
\subfigure[Boundary]{
\label{Fig.baseg_2774_bounds}
\includegraphics[width=0.32\linewidth,height=0.2\linewidth]{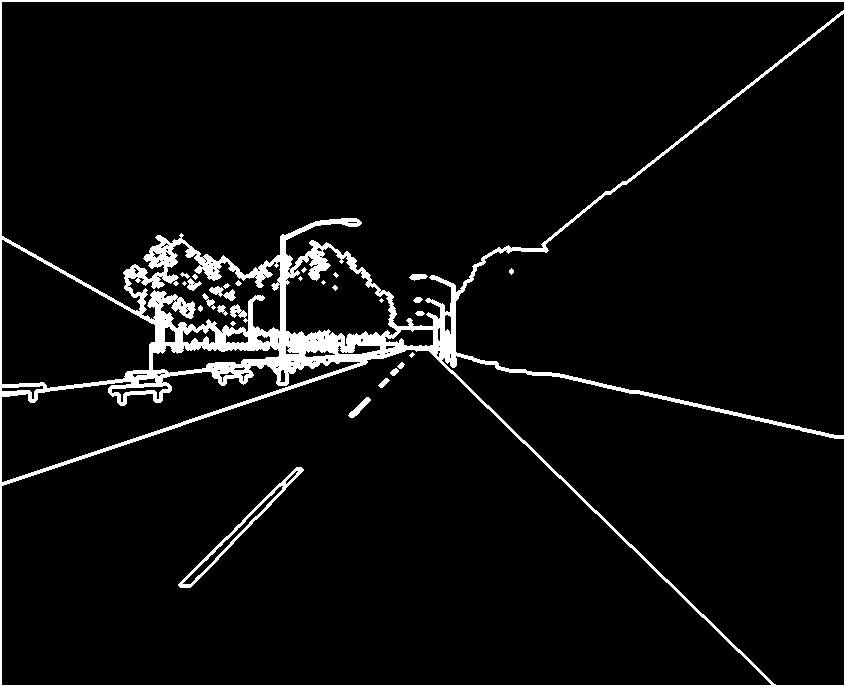}}
\hspace{-0.3cm}
\subfigure[Prediction]{
\label{Fig.baseg_2774_prediction}
\includegraphics[width=0.32\linewidth,height=0.2\linewidth]{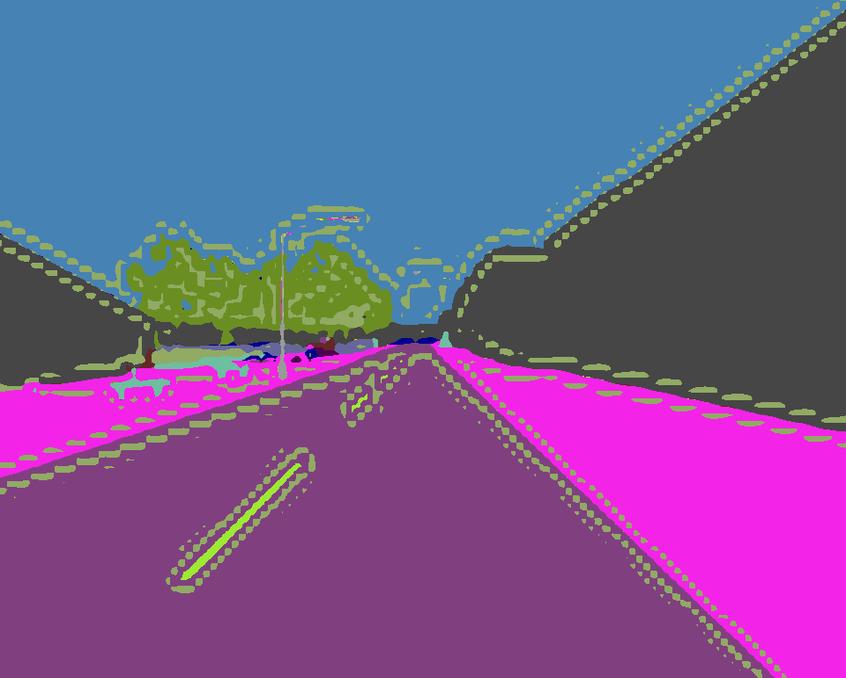}}
\vspace{-0.2cm}
\caption{Illustration of extra boundary information for BASeg.}
\label{Fig.baseg_boundary}
\vspace{-0.3cm}
\end{figure*}

From both \Cref{Tab.perf_comp} and \Cref{tab:semantic_pred}, we can find that BASeg achieves relatively competitive performance to the proposed AdvImmu and shows overwhelming advantages over other baselines. This is attributed to the consideration of extra boundary information in the training of BASeg, which can be illustrated in \Cref{Fig.baseg_boundary}. \Cref{Fig.baseg_2774} indicates an exmaple of image and \Cref{Fig.baseg_2774_bounds} denotes the corresponding boundary information. Obviously, these boundary information can assist BASeg model to overcome weaknesses, such as boundary uncertainty, class confision, etc. Thus, BASeg can achieve competitive performance underer various weather conditions. However, incorporation of extra boundary information is a double-edged sword and leads to obvious prediction noise along boundaries, which is demonstrated in \Cref{Fig.baseg_2774_prediction} and the third column of \Cref{tab:semantic_pred}.

\subsubsection{AdvImmu's Convergence}
\begin{figure*}[tp]
\hspace{-0.2cm}
\vspace{-0.4cm}
\subfigure[mIoU]{
\label{Fig.convergence_mIoU}
\includegraphics[width=0.24\linewidth, height=0.18\linewidth]{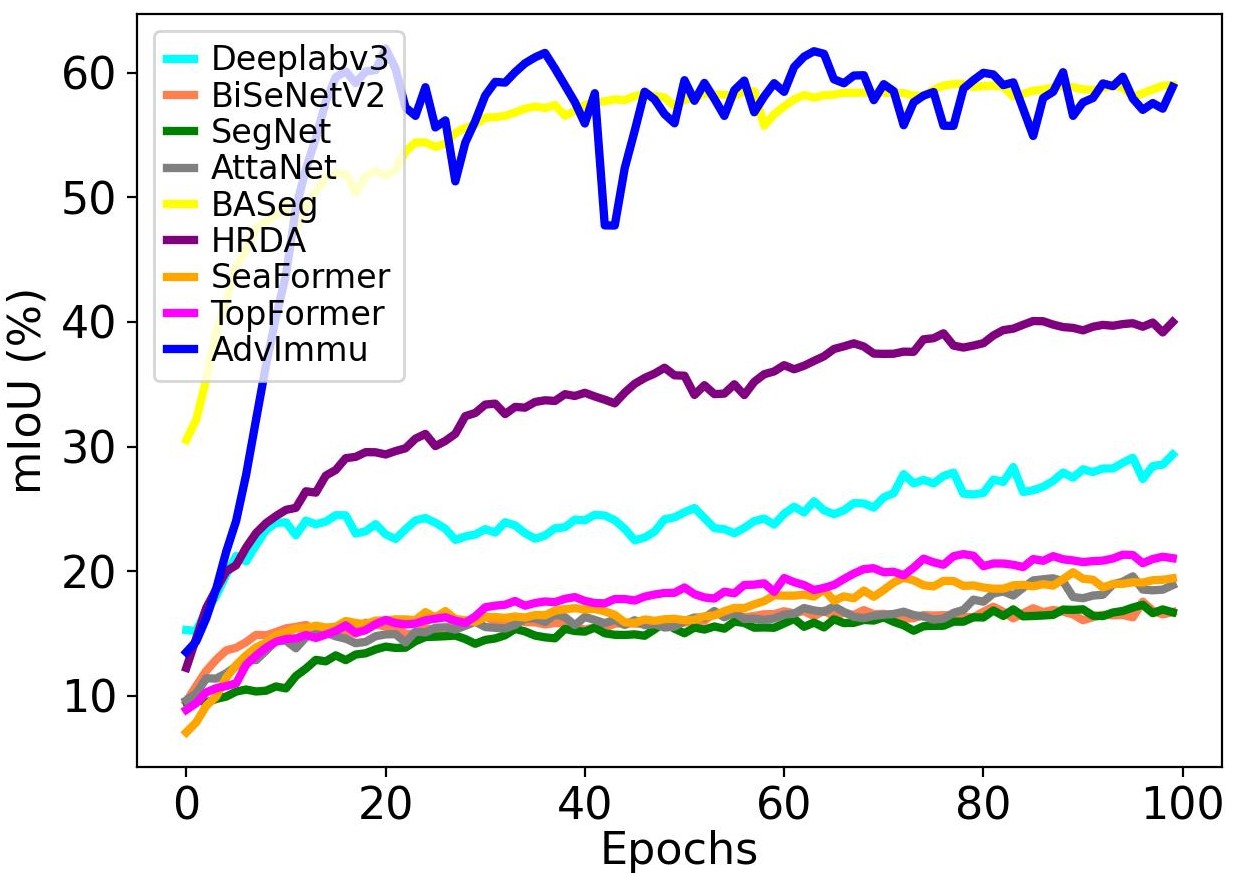}}
\hspace{-0.25cm}
\subfigure[mPre]{
\label{Fig.convergence_mPre}
\includegraphics[width=0.24\linewidth, height=0.18\linewidth]{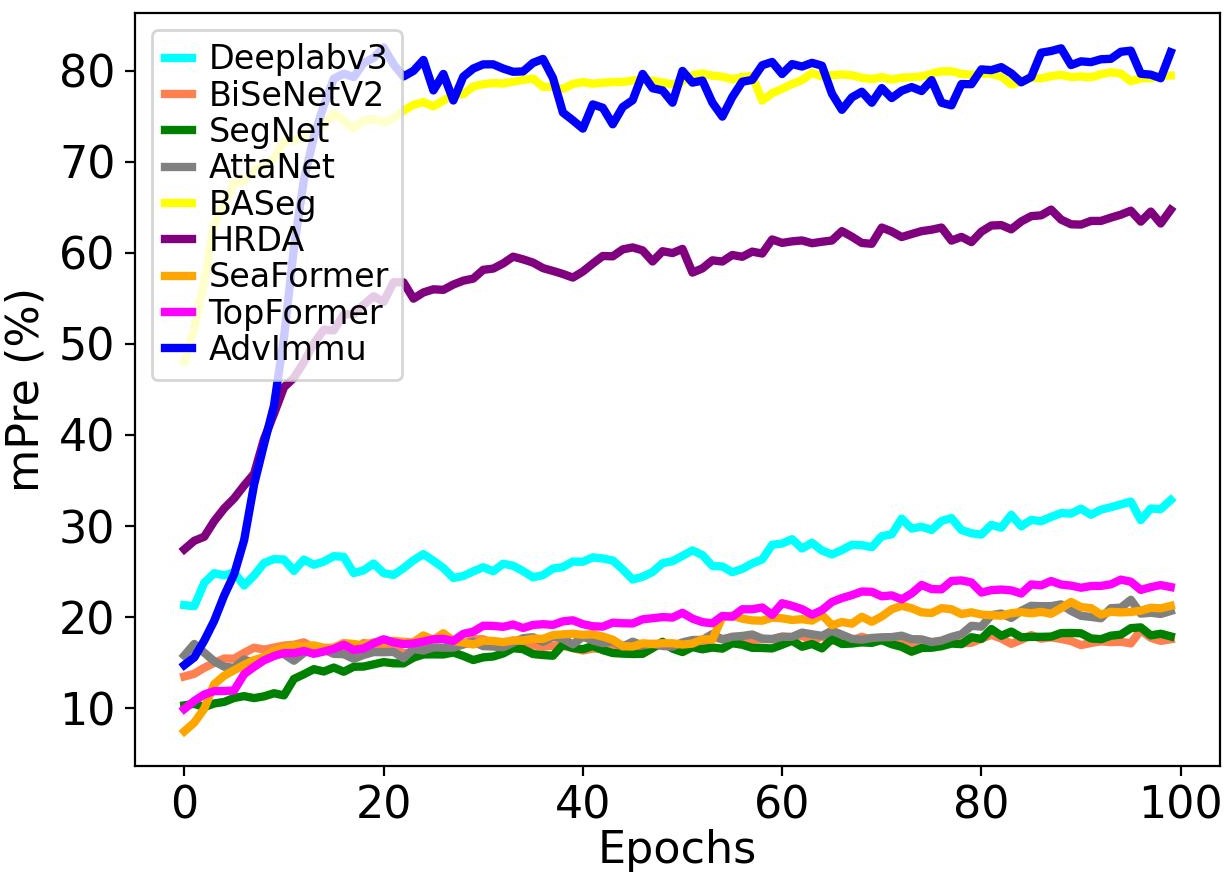}}
\hspace{-0.25cm}
\subfigure[mRec]{
\label{Fig.convergence_mRec}
\includegraphics[width=0.24\linewidth, height=0.18\linewidth]{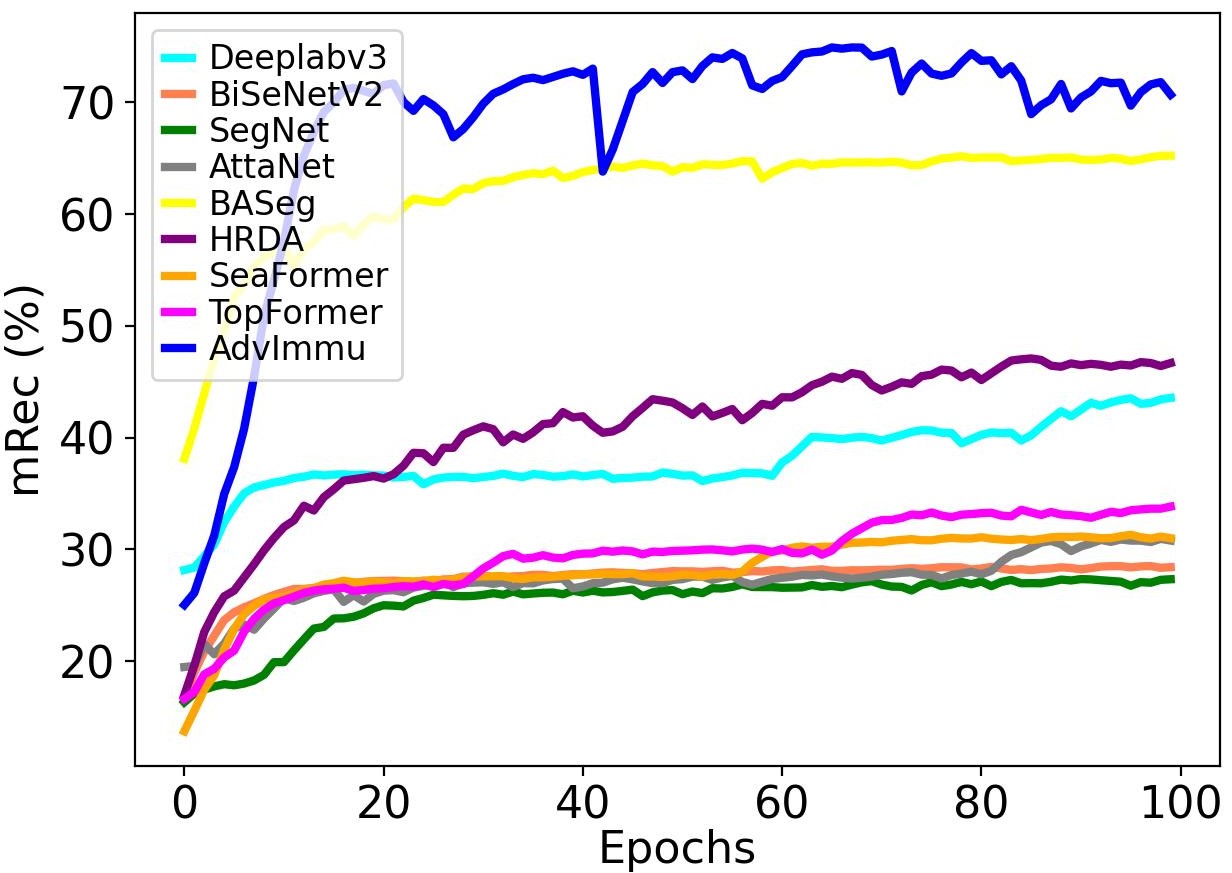}}
\hspace{-0.25cm}
\subfigure[mF1]{
\label{Fig.convergence_mF1}
\includegraphics[width=0.24\linewidth, height=0.18\linewidth]{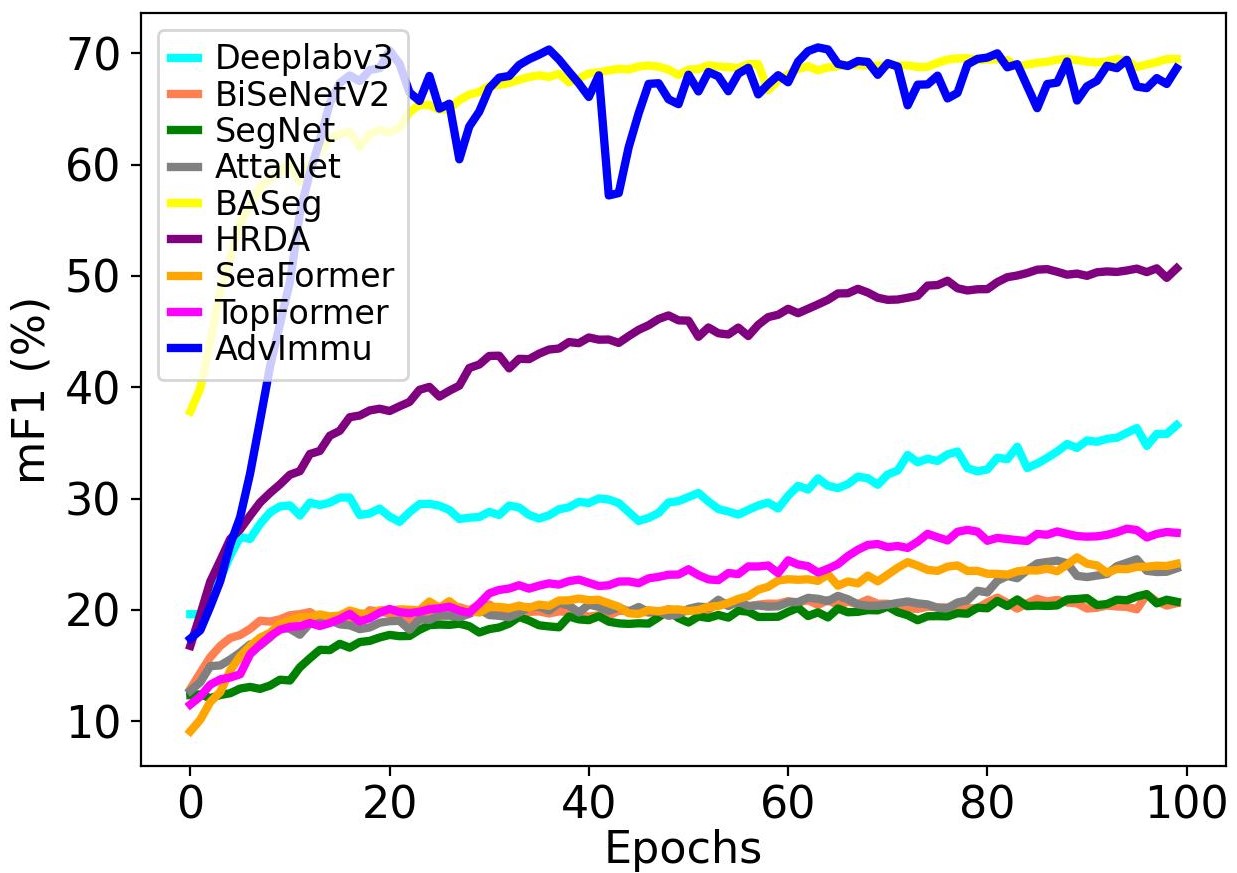}}
\vspace{0.15cm}
\caption{Convergence comparison of AdvImmu against other baselines on GTA5 dataset.}
\label{Fig.Convergence_comp}
\vspace{-0.3cm}
\end{figure*}

\Cref{Fig.Convergence_comp} compares the convergence of the proposed AdvImmu against all other baselines. From \Cref{Fig.Convergence_comp}, we can conclude following patterns: \textbf{(I)} Across almost all metrics, the proposed AdvImmu converges fastest compared to all the other baselines. For instance, taking mIoU as example, AdvImmu reaches to the maximum mIoU at around the 18-th epochs whereas other baselines reach their maximum value later than the 18-th epochs. \textbf{(II)} DL network architecture imposes crucial impact on convergence. Specifically, CNN-based models (\eg, AdvImmu, BASeg, DeepLabv3+, etc.) generally convergence faster than Transformer-based models (\eg, HRDA, SeaFormer, TopFormer, etc.). 

\subsubsection{AdvImmu's Complexity and Inference Delay}
\Cref{Tab:flops_and_fps} compares the FLOPs (using GFLOPs instead for being straightforward) and FPS of AdvImmu against other baselines. We can figure out following insights: \textbf{(I)} Models like TopFormer and BiSeNetV2 achieve high FPS with lower computational demands, making them suitable for real-time applications. \textbf{(II)} HRDA and BASeg have the highest GFLOPs (823.57 and 562.31, respectively) but suffer from low FPS, indicating that they are computationally intensive and slower. \textbf{(III)} AdvImmu has a GFLOPs similar to DeepLabv3+ but a significantly lower FPS (28.11), which is caused by AdvImmu's loading heavy input and unfolding regularizers.

\subsection{In-depth Insights of AdvImmu} \label{abl_study}
In this section, we will conduct comprehensive experiments and empirical analyses to investigate how LSM, GSM, URs, and feature fusion type impact the performance of the proposed AdvImmu.

\begin{table}[tp]
\vspace{0.2cm}
\caption{The comparison of GFLOPs and FPS of AdvImmu against other competitors}
\setlength{\tabcolsep}{20.5pt}
\begin{tabularx}{\linewidth}{ccc}
\hline
Methods    & GFLOPs & FPS \\ \hline
DeepLabv3+ & 50.58     & 297.06    \\
BiSeNetV2  & 36.46     & 545.84    \\
SegNet     & 327.93    & 65.54    \\
AttaNet    & 23.91     & 506.40    \\
BASeg      & 562.31    & 35.60    \\
HRDA       & 823.57    & 9.69     \\
SeaFormer  & 13.77     & 234.00    \\
TopFormer  & 3.56      & 490.38    \\
\textbf{AdvImmu (Ours)}    & 50.02     & 28.11     \\ \hline
\end{tabularx}
\label{Tab:flops_and_fps}
\end{table}

\begin{table}[tp]
\caption{The effect of LSM depth on AdvImmu performance}
\setlength{\tabcolsep}{5.5pt}
\begin{tabularx}{\linewidth}{ccccc}
\hline
LSM Depth & mIoU       & mPre       & mRec       & mF1        \\ \hline
Depth-1   & 41.52$\pm$0.77 & \textbf{57.83$\pm$0.92} & 46.90$\pm$0.67  & 49.96$\pm$0.78 \\
Depth-2   & 41.29$\pm$0.80  & 57.43$\pm$0.60  & 46.68$\pm$0.84 & 49.76$\pm$0.80  \\
Depth-3   & 41.58$\pm$0.78 & 57.52$\pm$0.85 & \textbf{47.08$\pm$0.72} & \textbf{50.09$\pm$0.75} \\
Depth-4   & \textbf{41.81$\pm$0.88} & 56.78$\pm$0.70  & 46.23$\pm$0.81 & 49.13$\pm$0.84 \\ \hline
\end{tabularx}
\label{Tab.LSM_depth_comp}
\vspace{0.3cm}
\end{table}

\subsubsection{LSM Depth on AdvImmu Performance}
\begin{figure*}[tp]
\hspace{-0.2cm}
\vspace{-0.4cm}
\subfigure[mIoU]{
\label{Fig.LSM_depth_mIoU}
\includegraphics[width=0.24\linewidth, height=0.18\linewidth]{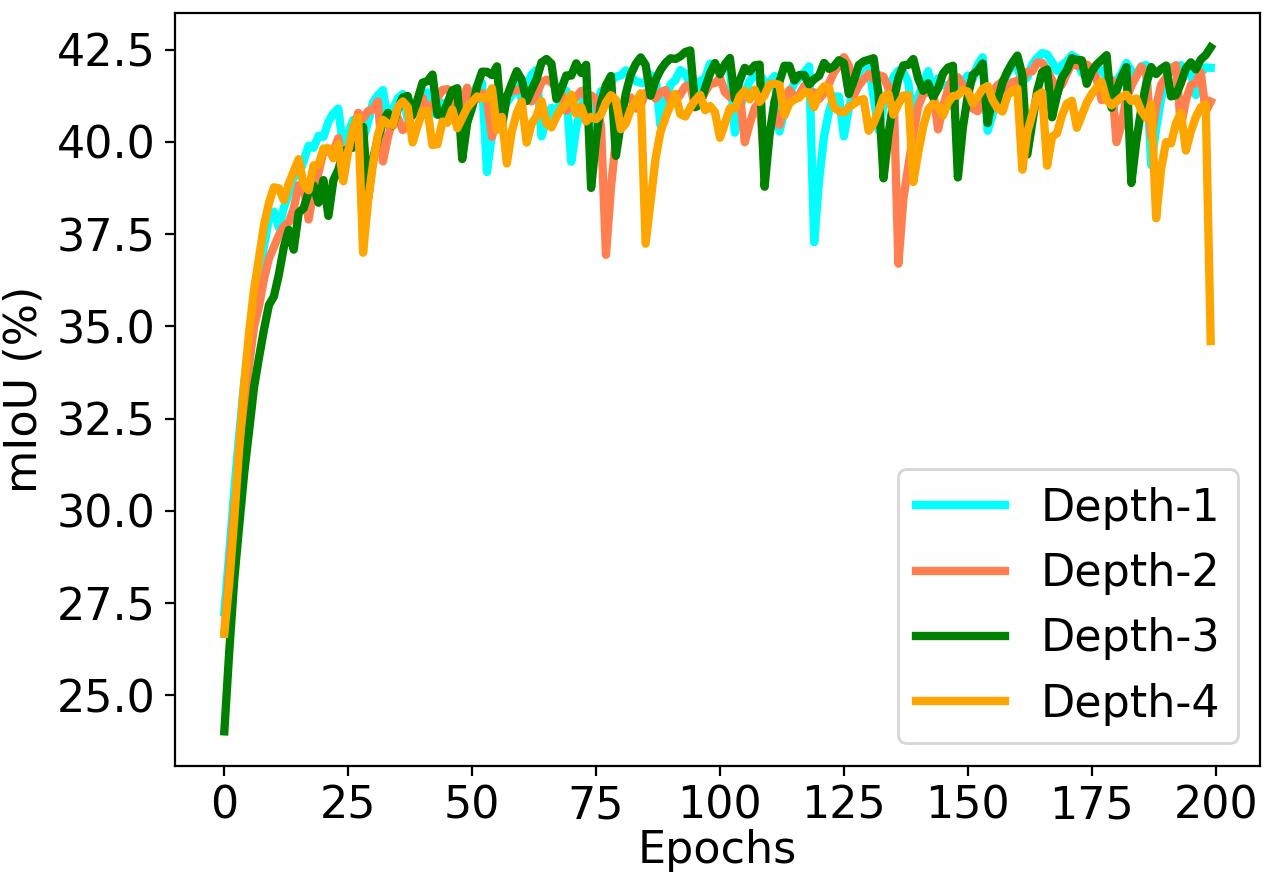}}
\hspace{-0.25cm}
\subfigure[mPre]{
\label{Fig.LSM_depth_mPre}
\includegraphics[width=0.24\linewidth, height=0.18\linewidth]{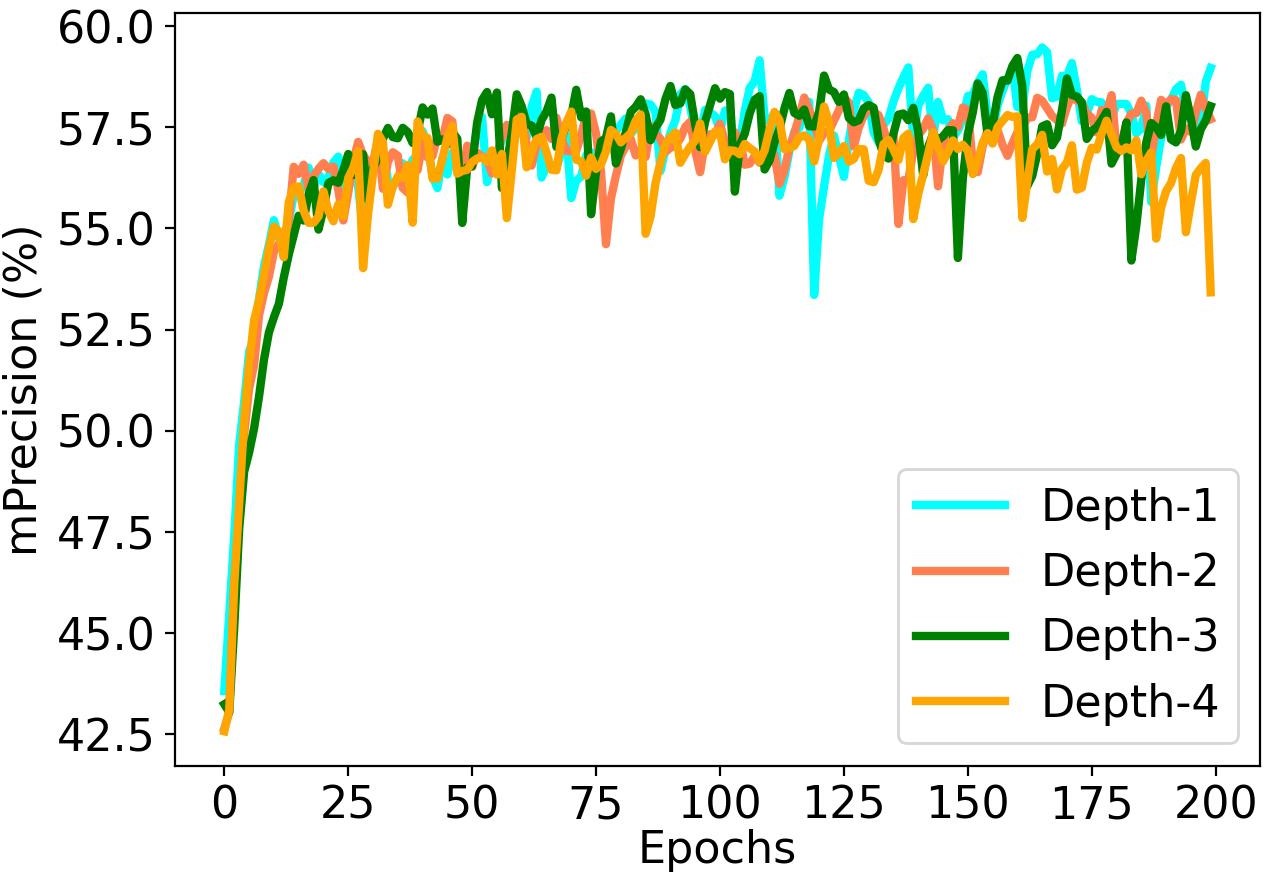}}
\hspace{-0.25cm}
\subfigure[mRec]{
\label{Fig.LSM_depth_mRec}
\includegraphics[width=0.24\linewidth, height=0.18\linewidth]{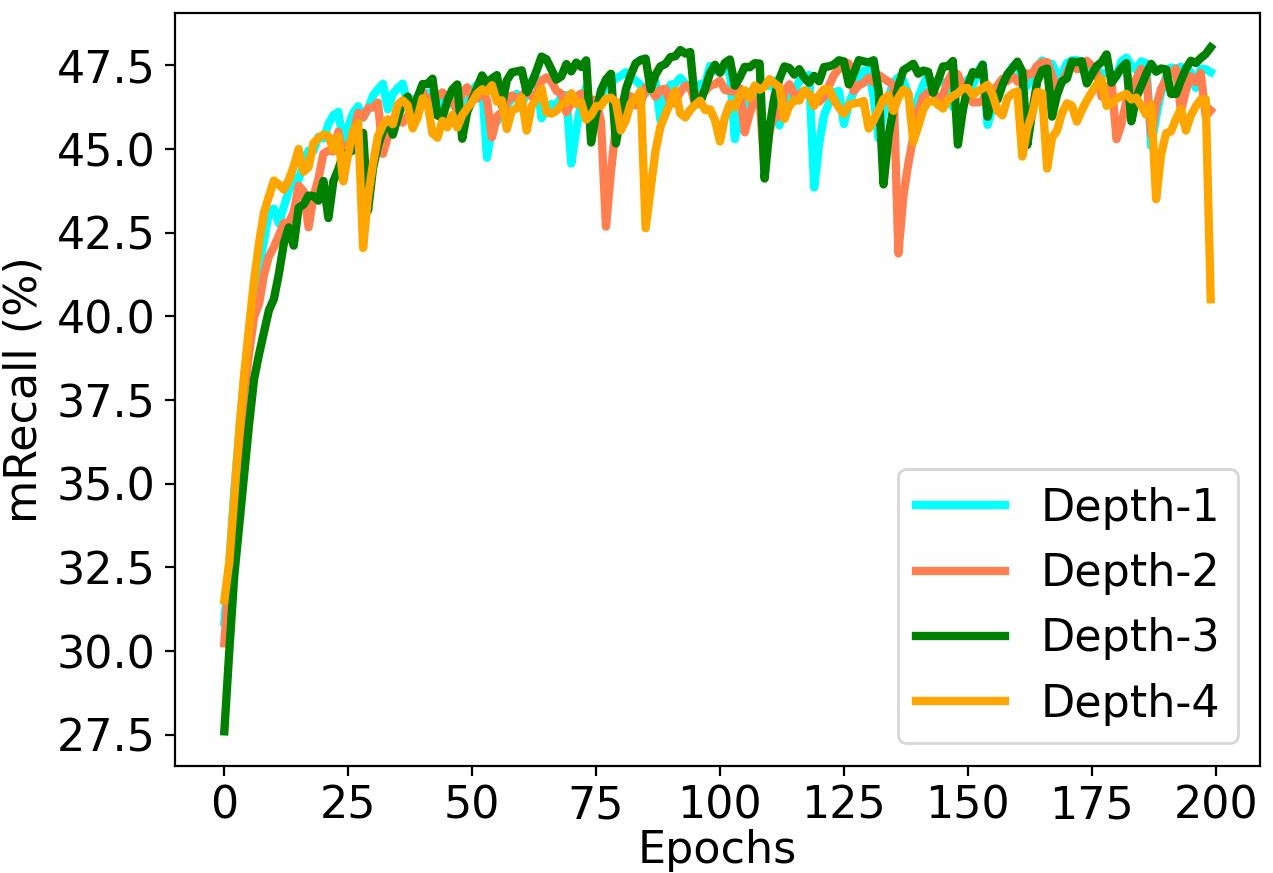}}
\hspace{-0.25cm}
\subfigure[mF1]{
\label{Fig.LSM_depth_mF1}
\includegraphics[width=0.24\linewidth, height=0.18\linewidth]{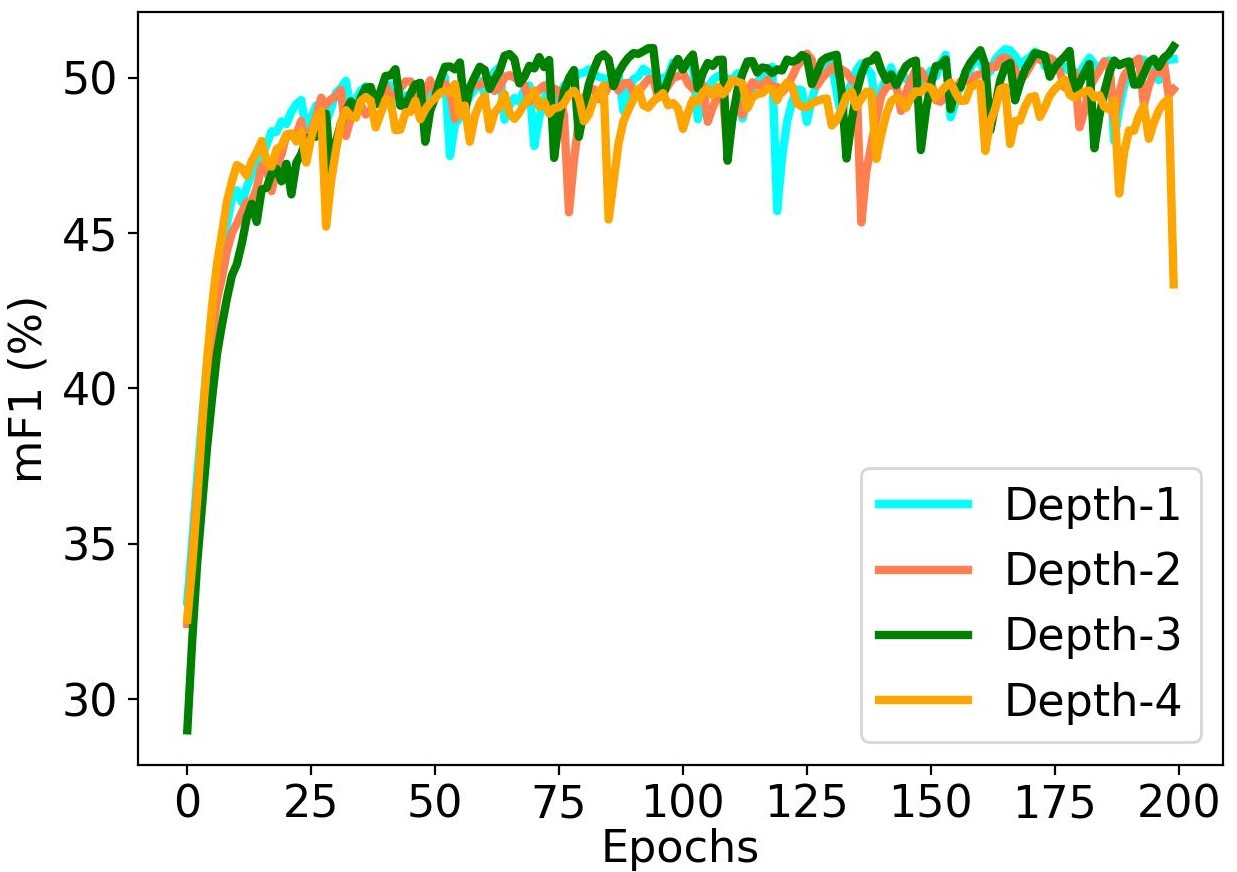}}
\vspace{0.15cm}
\caption{The comparison among different LSM depths.}
\label{Fig.LSM_depth_comp}
\vspace{-0.3cm}
\end{figure*}

LSM depth represents the number of consecutive frames prior to current frame. It can impose effect on IntU and DU, and further affects LSM and AdvImmu. In this experiment, we compare four cases with setting LSM depth to 1, 2, 3 and 4, and denote them as Depth-1, Depth-2, Depth-3, and Depth-4, respectively. \Cref{Tab.LSM_depth_comp} compares the effect of the LSM depth on AdvImmu performance. From \Cref{Tab.LSM_depth_comp}, we can find that across different LSM depths (Depth-1 through Depth-4), the performance metrics show certain variation, indicating that changes in LSM depth have impact on AdvImmu's performance. \Cref{Fig.LSM_depth_comp} corroborates this finding visually, showing that all depths reach different performance levels with slight fluctuations. However, the subtle variations in performance across different LSM depths indicate that neither too shallow (Depth-1) nor too deep (Depth-4) configurations do not consistently yield the best results. Depth-3 generally shows more stable and slightly improved performances, suggesting that moderate depths could be optimal. This implies that an intermediate LSM depth could potentially offer a better compromise between computation complexity and performance.

\subsubsection{EF vs LF}
\begin{figure*}[tp]
\hspace{-0.6cm}
\subfigure[EF architecture]{
\label{Fig.EF_arch}
\includegraphics[width=0.35\linewidth, height=0.22\linewidth]{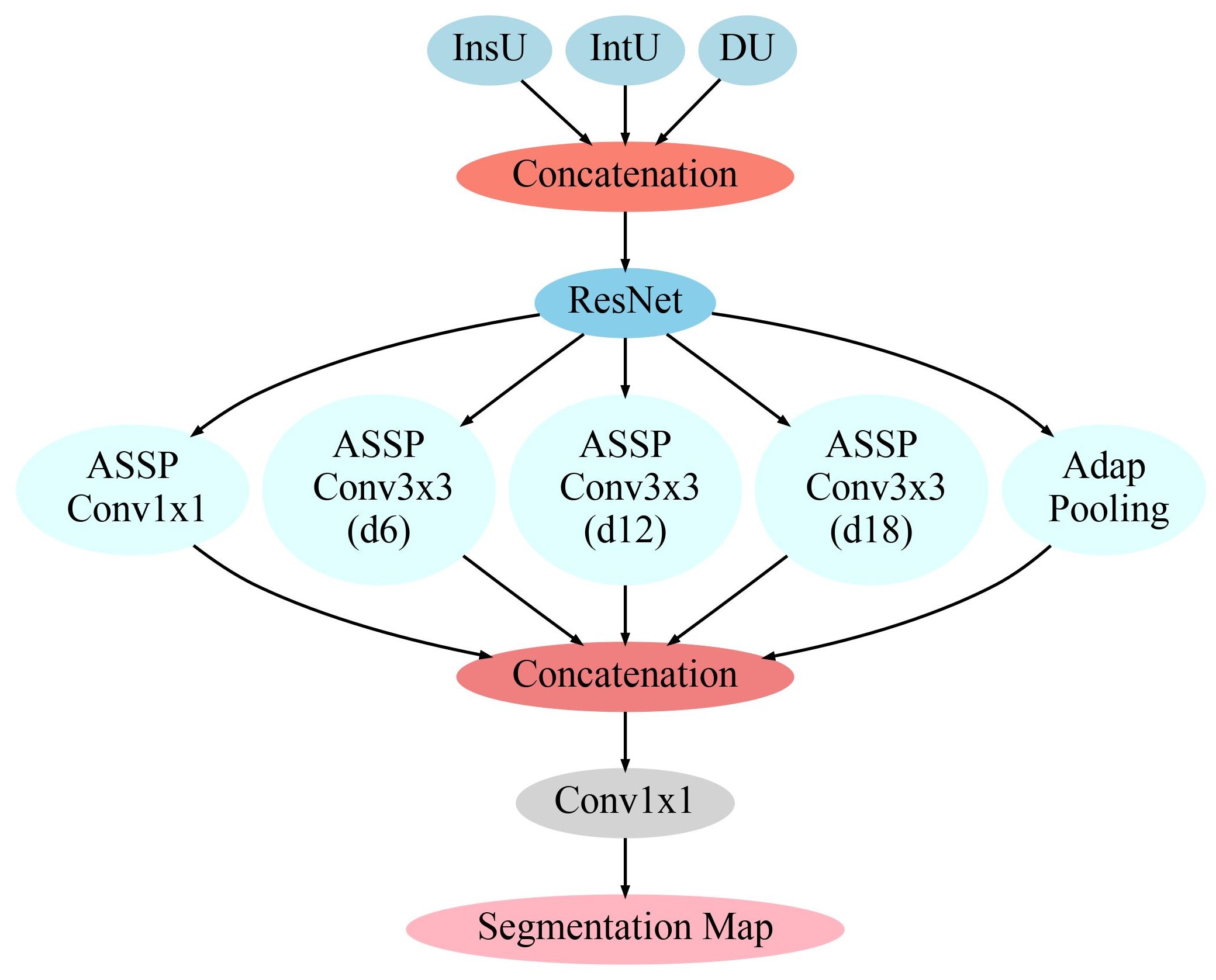}}
\hspace{-0.6cm}
\subfigure[LF architecture]{
\label{Fig.LF_arch}
\includegraphics[width=0.35\linewidth, height=0.22\linewidth]{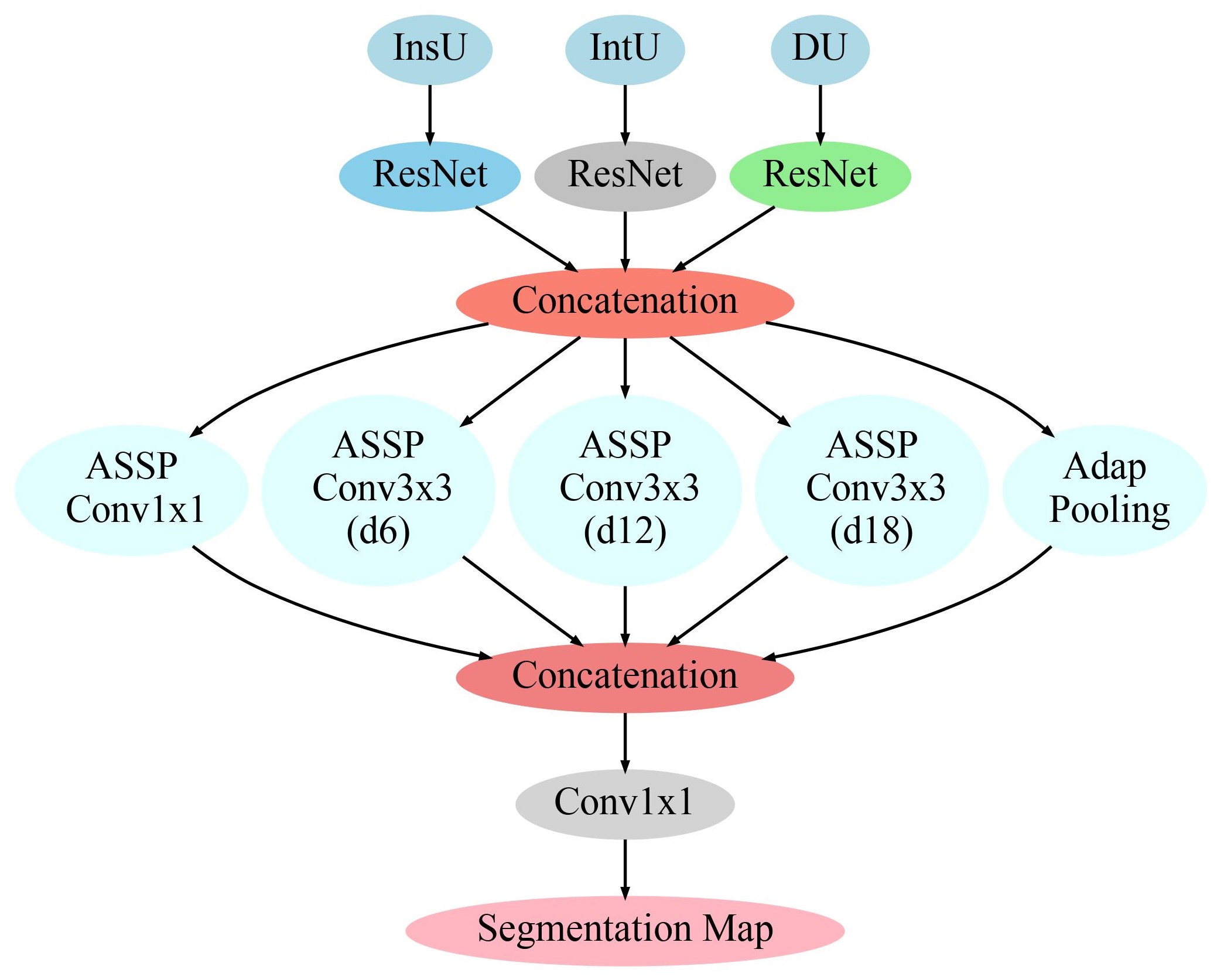}}
\hspace{-0.6cm}
\subfigure[Performance comparison]{
\label{Fig.EF_vs_LF_perf}
\includegraphics[width=0.34\linewidth, height=0.22\linewidth]{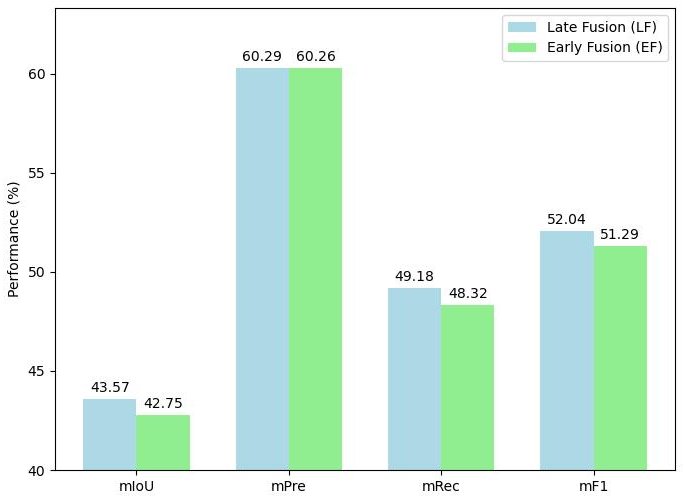}}
\caption{The comparison between EF and LF.}
\label{Fig.EF_vs_LF}
\vspace{-0.3cm}
\end{figure*}

EF and LF are two policies for fusing features from InsU, IntU, and DU. They both have respective pros and cons. Specifically, EF generally combines features at an early stage, allowing the model to learn interactions between InsU, IntU, and DU from the start. However, EF maybe suffers from overfitting due to the high dimensionality of combined features. In contrast, LF can preserve unique features of InsU, IntU, and DU, allowing specialized processing, whereas LF involves intensive computation and has less opportunity for learning interactions. In order to investigate how EF and LF affect the AdvImmu performance, we conduct experiments to compare their performance.  \Cref{Fig.EF_vs_LF} presents the comparison between EF and LF. Specifically, \Cref{Fig.EF_arch} and \Cref{Fig.LF_arch} illustrate the model architecture of EF and LF of AdvImmu, respectively. \Cref{Fig.EF_vs_LF_perf} compares the inference performance of EF and LF across four metrics, where LF performs better than EF in all adopted metrics. Overall, although LF generally outperforms EF, LF has more learnable parameters and thus requires more powerful computational device and more training time. Therefore, we should take both model performance and training cost into account to trade off them when we choose which one actually best fits our needs.

\subsubsection{Enabling GSM vs Disabling GSM}
\begin{figure*}[tp]
\hspace{-0.25cm}
\subfigure[mIoU]{
\label{Fig.GSM_mIoU}
\includegraphics[width=0.24\linewidth, height=0.18\linewidth]{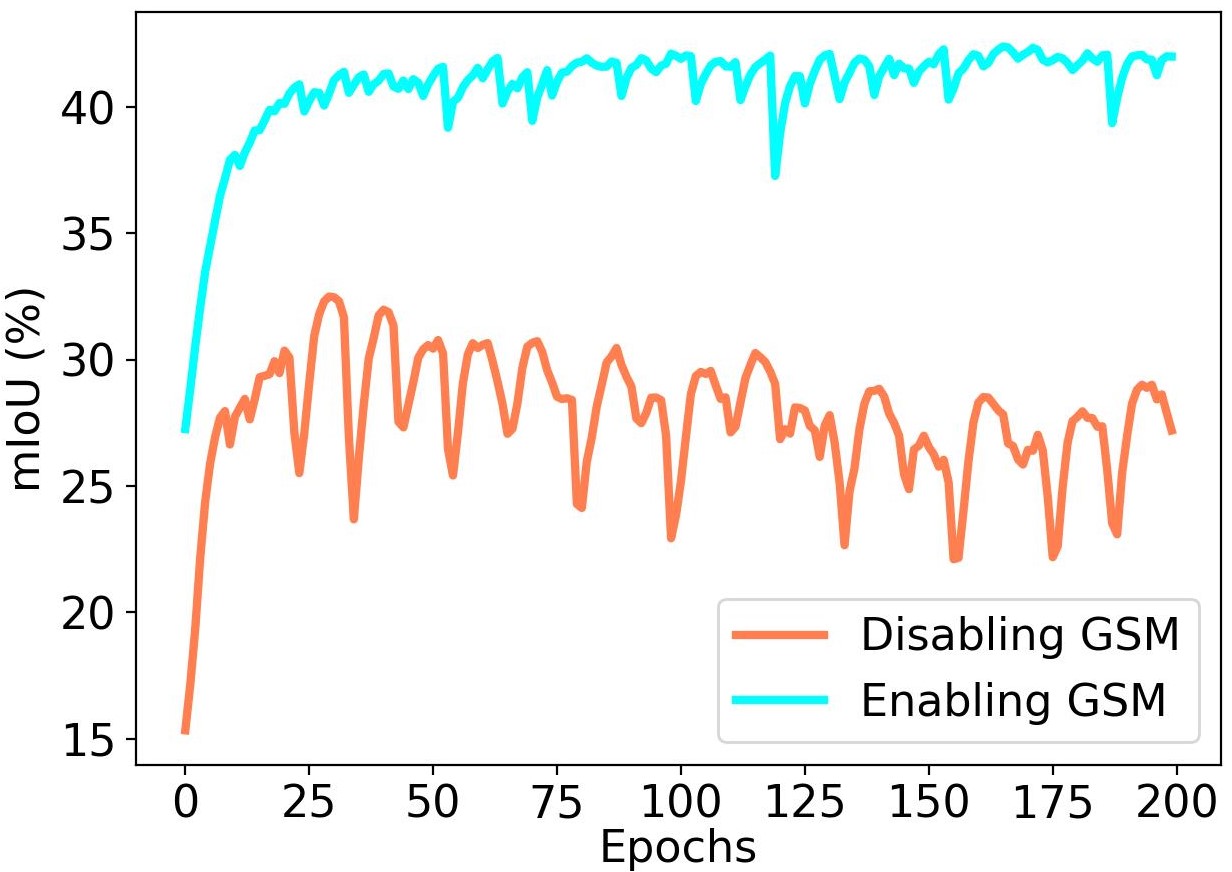}}
\hspace{-0.25cm}
\subfigure[mPre]{
\label{Fig.GSM_mPre}
\includegraphics[width=0.24\linewidth, height=0.18\linewidth]{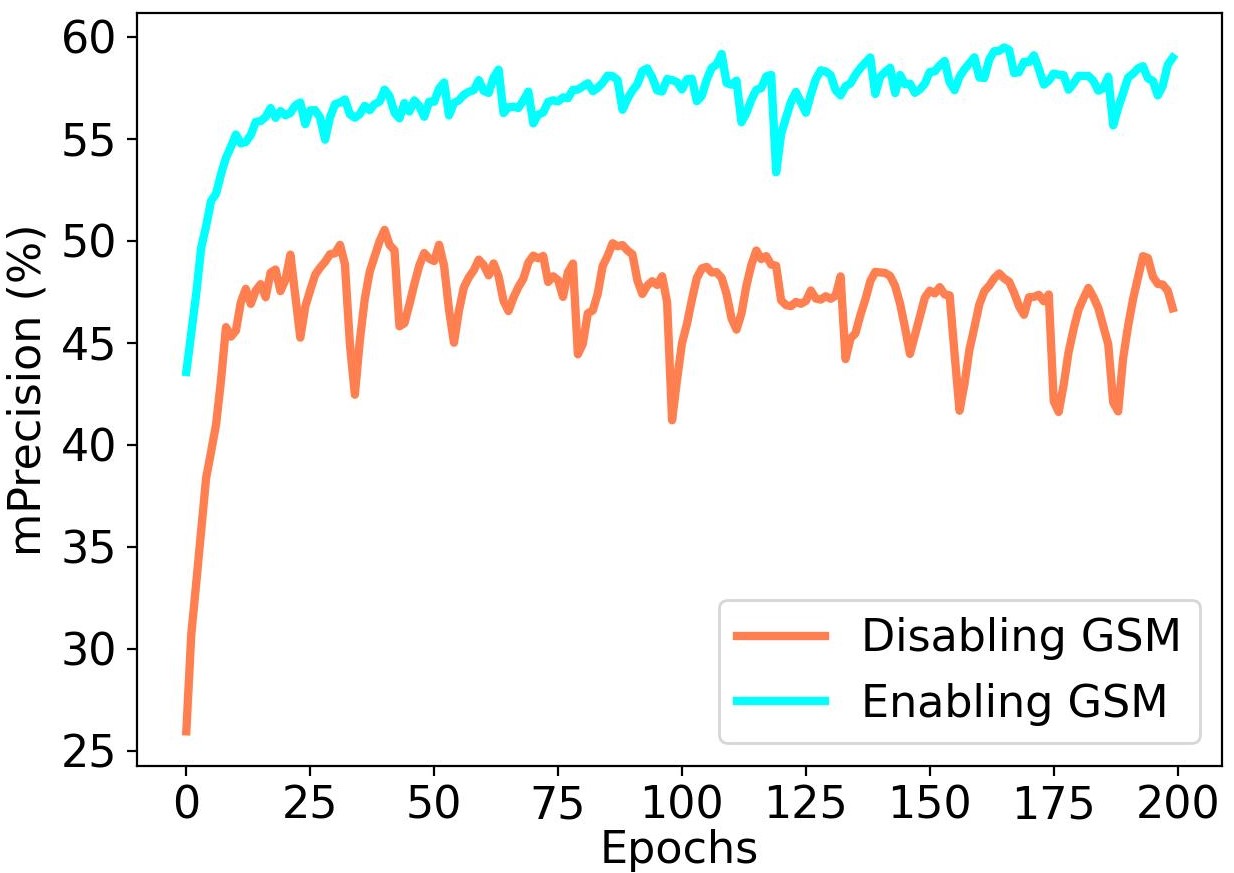}}
\hspace{-0.25cm}
\subfigure[mRec]{
\label{Fig.GSM_mRec}
\includegraphics[width=0.24\linewidth, height=0.18\linewidth]{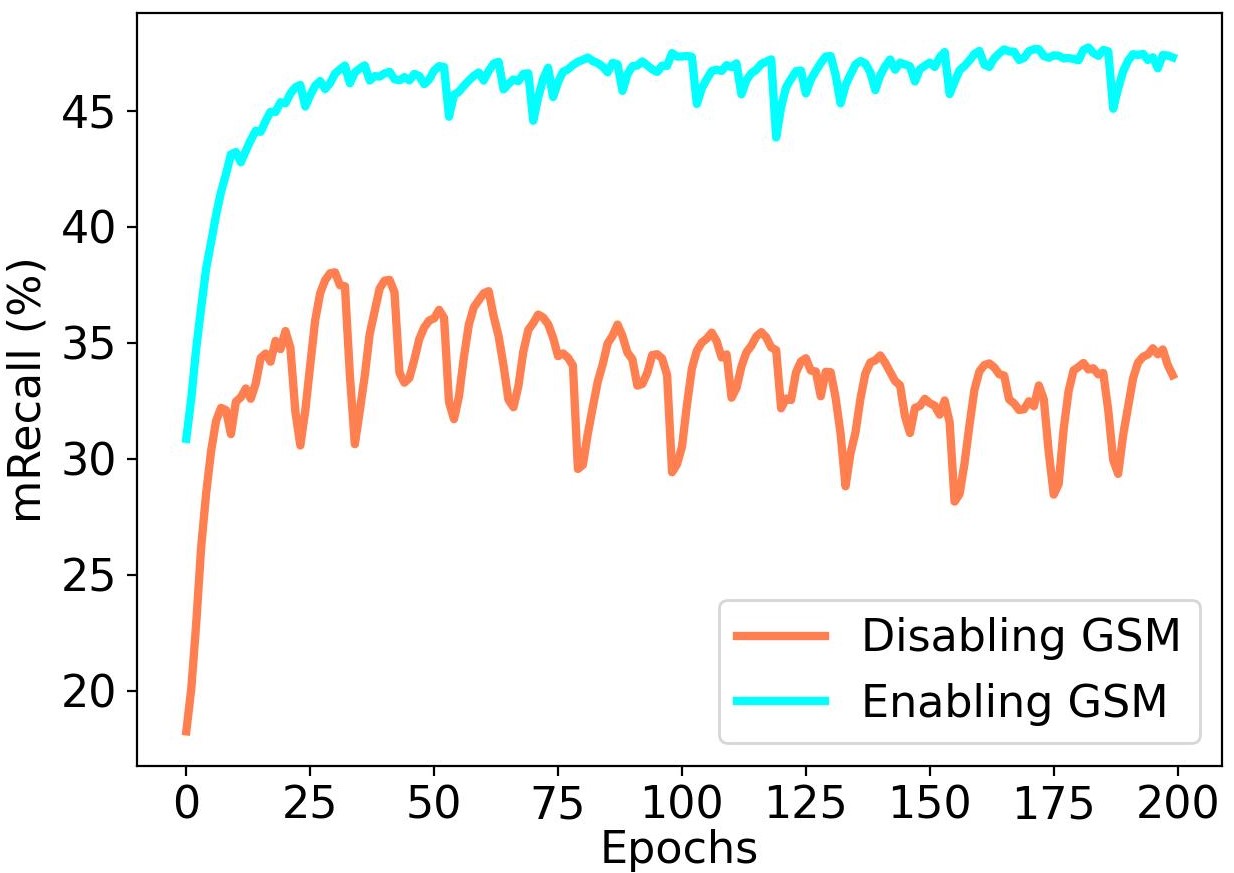}}
\hspace{-0.25cm}
\subfigure[mF1]{
\label{Fig.GSM_mF1}
\includegraphics[width=0.24\linewidth, height=0.18\linewidth]{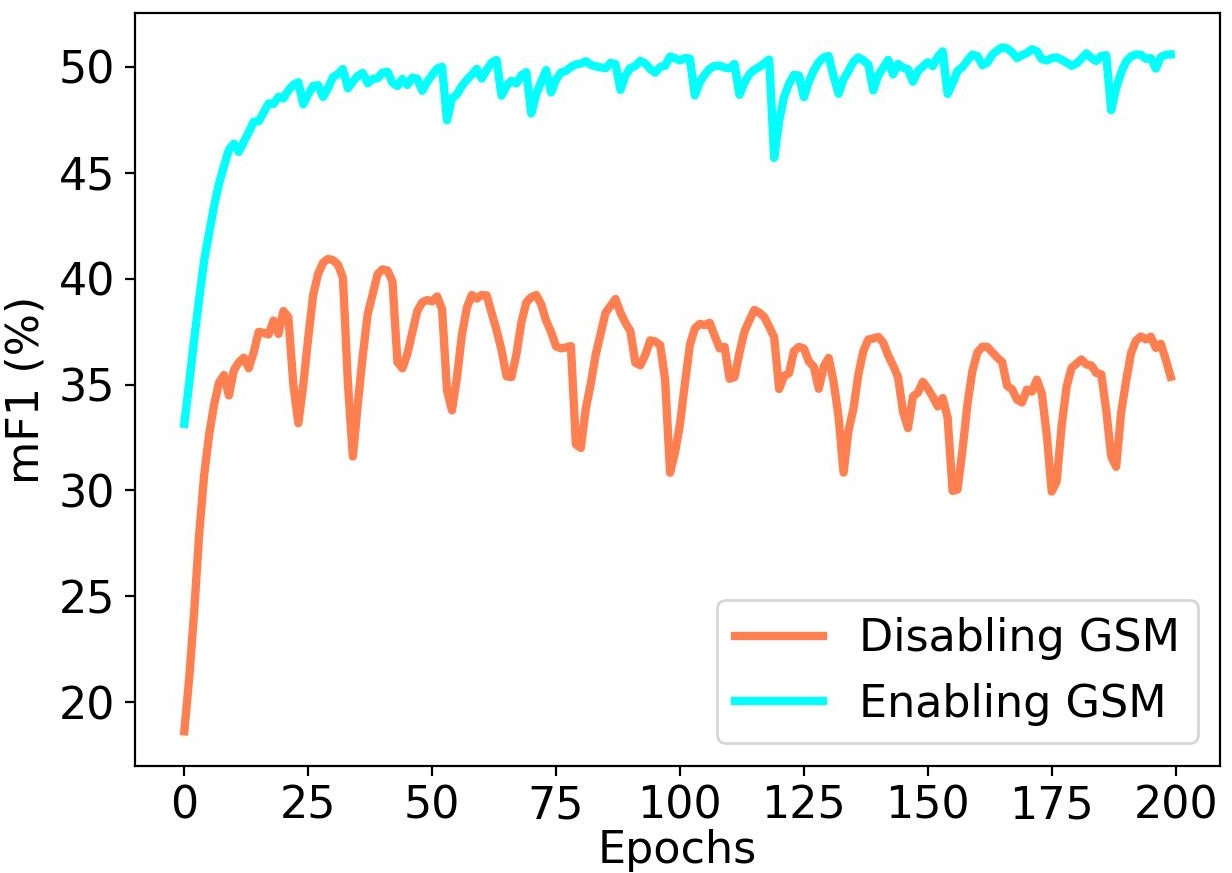}}
\vspace{-0.2cm}
\caption{The performance comparison between enabling and disabling GSM.}
\label{Fig.GSM_comp}
\vspace{-0.3cm}
\end{figure*}

GSM is designated to shuffle LSM segments to prevent the overfitting to specific temporal patterns. This is a key element to improve the across-weather performance of AdvImmu under various adverse weather conditions. To figure out how much GSM influences the performance of the proposed AdvImmu, we conduct experiments to compare the performance of enabling and disabling GSM. \Cref{Tab.GSM_comp} demonstrates that enabling GSM significantly improves AdvImmu's inference performance across all measured metrics. Specifically, we can consolidate the performance improvement from following two aspects: \textbf{(I)} Compared to disabling GSM, enabling GSM enhances all performance metrics with a large margin. Taking the mIoU as example, the performance improvement is (41.52 - 27.12) / 27.12 = 53.10\%. \textbf{(II)} Compared to disabling GSM, enabling GSM improves the performance stability across multiple adverse weather conditions. Still taking mIoU as example, the standard variance reduces with the ratio of (1.84 - 0.77) / 1.84 = 58.15\%. Such findings can further be confirmed visually by \Cref{Fig.GSM_comp}. Concretely, \Cref{Fig.GSM_comp} shows that with GSM enabled, the performance metrics are consistently higher and more stable across 200 epochs, compared to the lower and more fluctuating metrics when GSM is disabled. Overall, enabling GSM leads to better and more reliable performance.

\begin{table}[tp]
\caption{The effect of GSM on AdvImmu performance}
\setlength{\tabcolsep}{4.0pt}
\begin{tabularx}{\linewidth}{ccccc}
\hline
Loss          & mIoU       & mPre       & mRec       & mF1        \\ \hline
Disabling GSM & 27.12$\pm$1.84 & 46.81$\pm$1.76 & 32.96$\pm$1.69 & 35.36$\pm$1.94 \\
Enabling GSM  & \textbf{41.52$\pm$0.77} & \textbf{57.83$\pm$0.92} & \textbf{46.90$\pm$0.67} & \textbf{49.96$\pm$0.78} \\  \hline
\end{tabularx}
\label{Tab.GSM_comp}
\vspace{-0.2cm}
\end{table}

\begin{table}[tp]
\caption{The regularizer comparison}
\setlength{\tabcolsep}{3.5pt}
\begin{tabularx}{\linewidth}{ccccc}
\hline
Loss                    & mIoU       & mPre       & mRec       & mF1        \\ \hline
CE            & 26.77$\pm$0.77 & 38.35$\pm$0.78 & 31.09$\pm$0.64 & 32.54$\pm$0.87 \\
VRs & 31.49$\pm$0.34 & 50.23$\pm$0.89 & 35.56$\pm$0.40 & 37.84$\pm$0.43 \\
URs ($L=2$) & 52.13$\pm$2.94 & 68.06$\pm$1.21 & 53.96$\pm$2.86 & 58.02$\pm$2.57 \\
URs ($L=5$) & \textbf{59.35$\pm$0.57} & \textbf{81.38$\pm$2.22} & \textbf{60.76$\pm$0.49} & \textbf{65.28$\pm$0.55} \\ \hline
\end{tabularx}
\label{Tab.regularizer_comp}
\end{table}

\subsubsection{Vanilla regurlarizers (VRs) vs URs}
\begin{figure*}[tp]
\hspace{-0.25cm}
\subfigure[mIoU]{
\label{Fig.regularizer_mIoU}
\includegraphics[width=0.24\linewidth, height=0.18\linewidth]{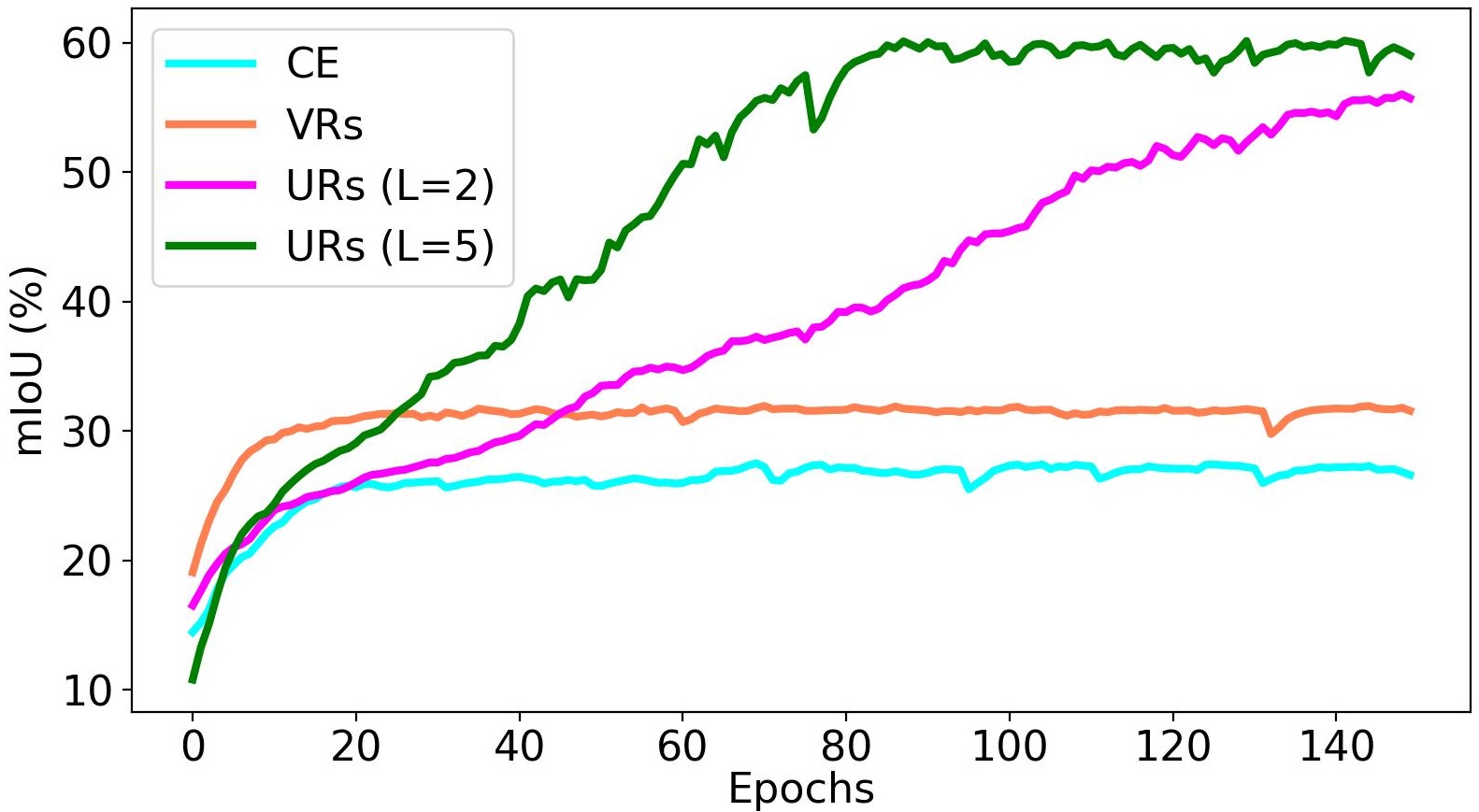}}
\hspace{-0.25cm}
\subfigure[mPre]{
\label{Fig.regularizer_mPre}
\includegraphics[width=0.24\linewidth, height=0.18\linewidth]{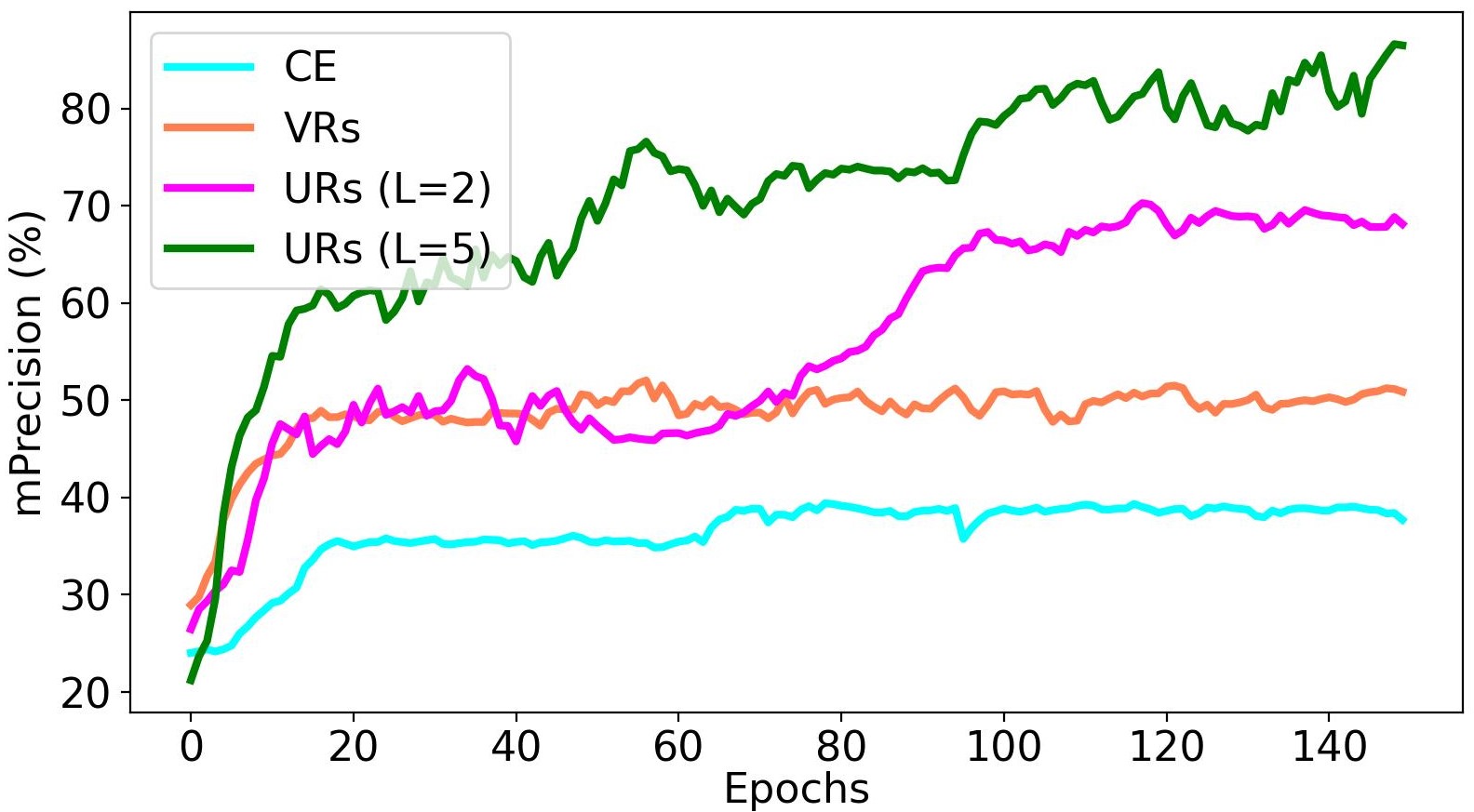}}
\hspace{-0.25cm}
\subfigure[mRec]{
\label{Fig.regularizer_mRec}
\includegraphics[width=0.24\linewidth, height=0.18\linewidth]{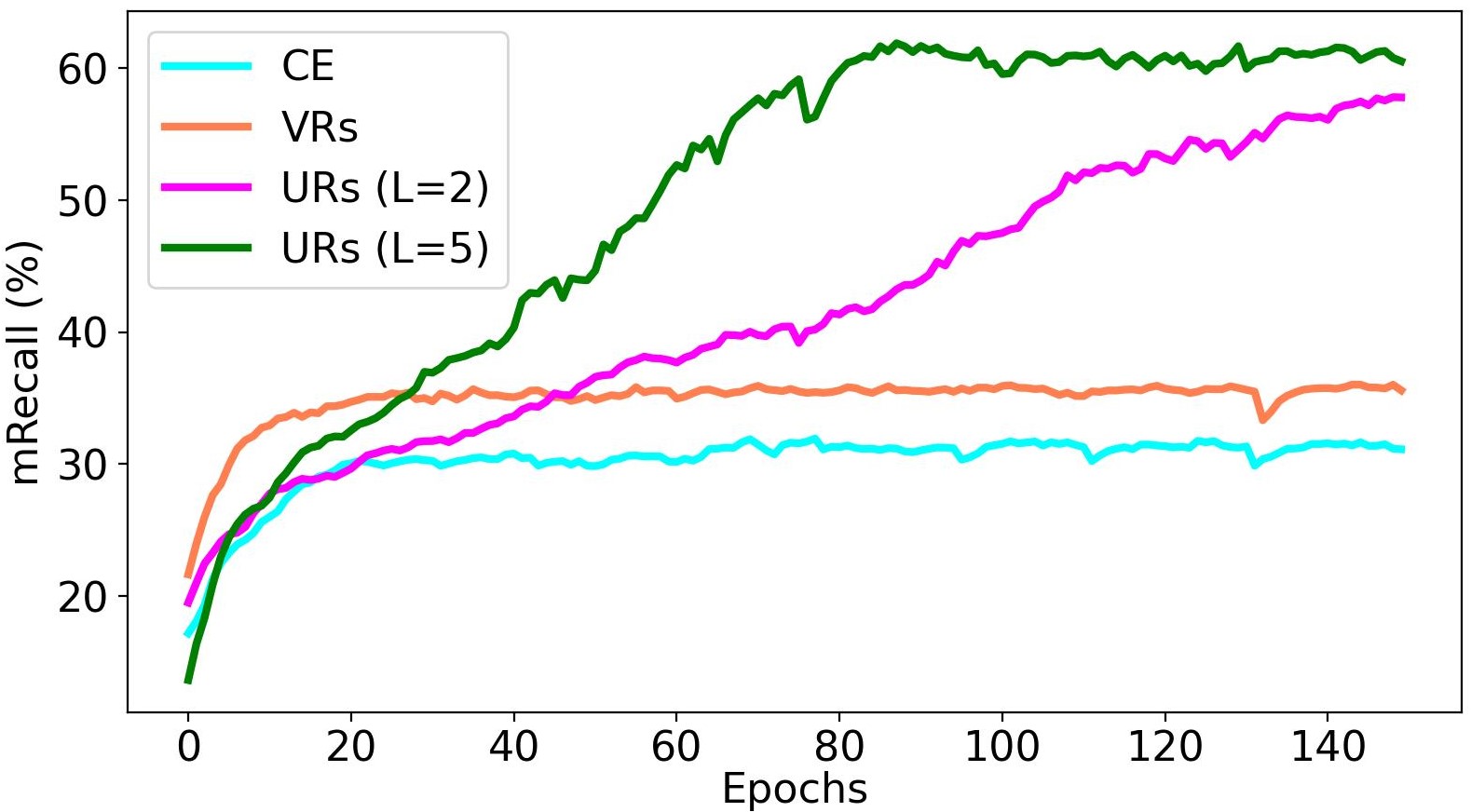}}
\hspace{-0.25cm}
\subfigure[mF1]{
\label{Fig.regularizer_mF1}
\includegraphics[width=0.24\linewidth, height=0.18\linewidth]{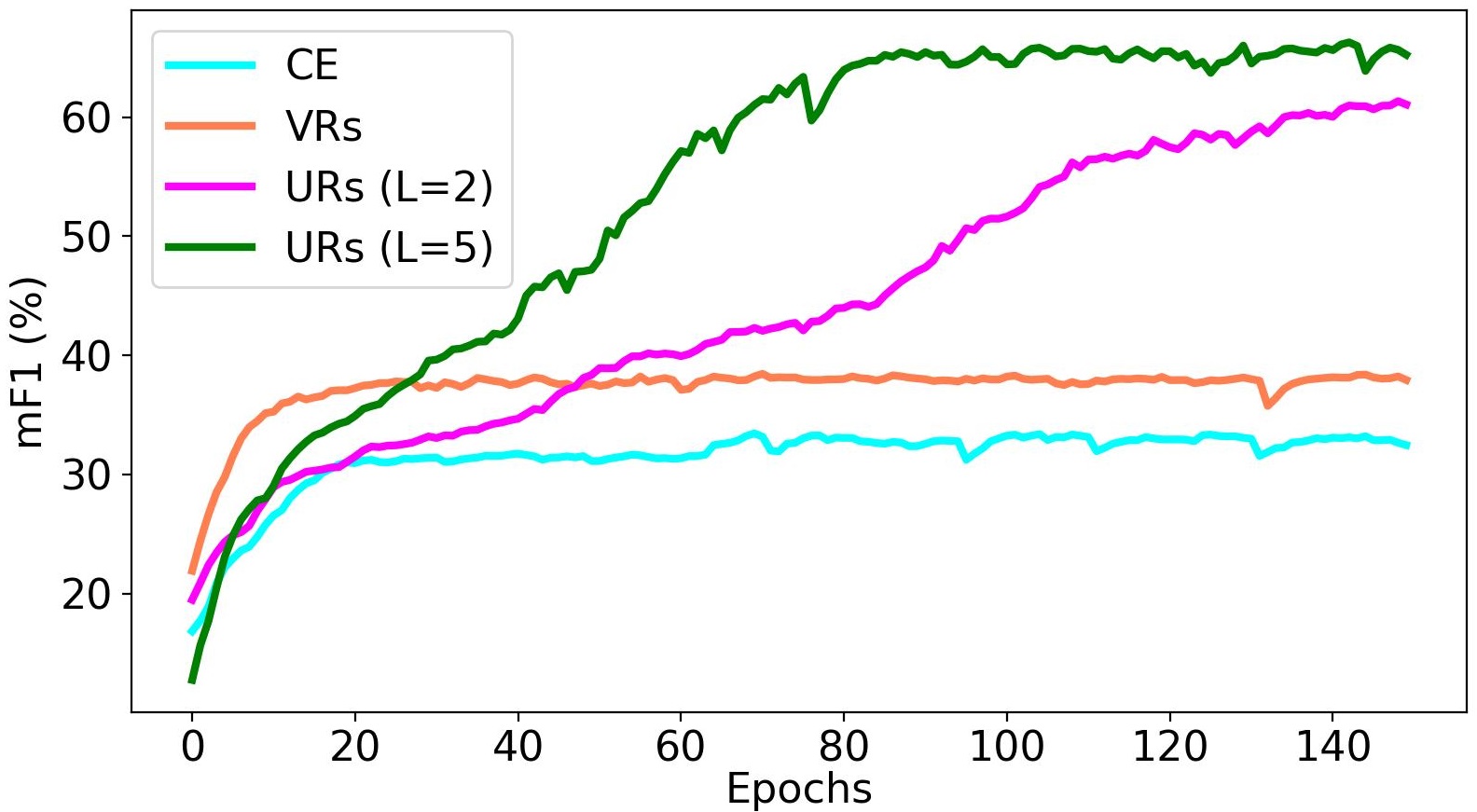}}
\vspace{-0.2cm}
\caption{The regularizer comparison.}
\label{Fig.regularizer_comp}
\end{figure*}

\Cref{Tab.regularizer_comp} and \Cref{Fig.regularizer_comp} compare the performance of the following cases: \textbf{(I)} only cross entropy loss (denoted as CE); \textbf{(II)} cross entropy plus vanilla regularizers (denoted as VRs); \textbf{(III)} cross entropy plus 2-layer unfolded regularizers (denoted as URs ($L=2$)); \textbf{(IV)} cross entropy plus 5-layer unfolded regularizers (denoted as URs ($L=5$)). By comparing these four cases, we can observe the following patterns: \textbf{(I)} \Cref{Tab.regularizer_comp} shows a consistent improvement in performance when the proposed image-level regularizer and inter-class contrastive regularizers are added to the basic CE loss, leading to better scores in mIoU, mPre, mRec, and mF1. \textbf{(II)} \Cref{Tab.regularizer_comp} also demonstrates that URs outperform corresponding VRs in almost all metrics. For example, URs ($L=2$) and URs ($L=5$) achieve mIoUs of 52.13 and 59.35, respectively, compared to 31.49 achieved by VRs. \textbf{(III)} \Cref{Tab.regularizer_comp} additionally shows that the more layers the regularizers are unfolded, the better the performance obtained. For instance, URs ($L=5$) achieves an mIoU of 59.35, better than the 52.13 achieved by URs ($L=2$). These patterns can be visually confirmed by \Cref{Fig.regularizer_comp}. In addition, \Cref{Fig.regularizer_comp} illustrates that URs ($L=5$) converges faster than URs ($L=2$). 

\begin{figure*}[tp]
    \centering
    \begin{minipage}{0.2\linewidth}
        \centering
        \subfigure[Raw Image]{
            \label{Fig.chan_raw}
            \includegraphics[width=\linewidth, height=0.575\linewidth]{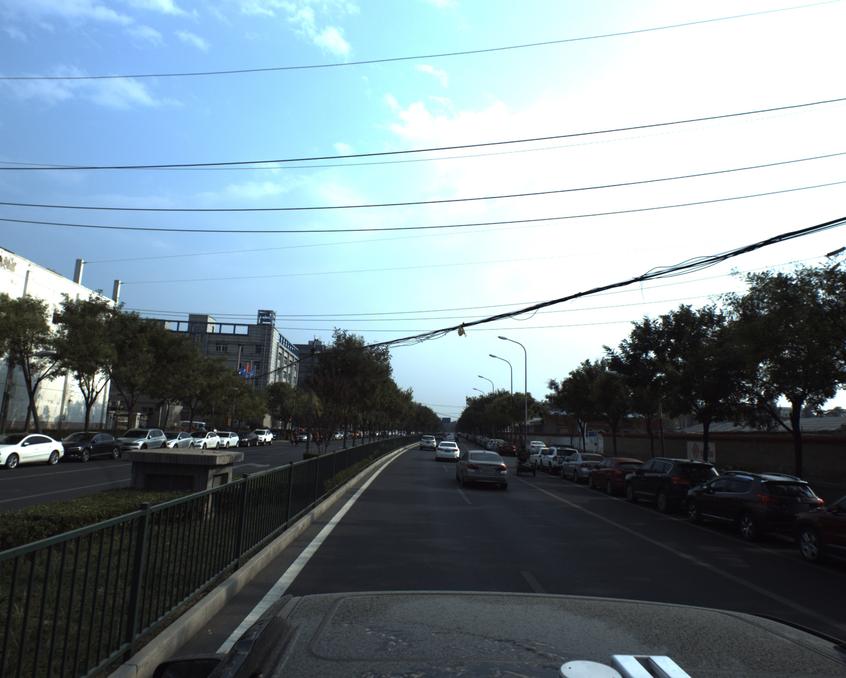}}

        \vspace{-0.3cm}
        \subfigure[Ground Truth]{
            \label{Fig.chan_gt}
            \includegraphics[width=\linewidth, height=0.575\linewidth]{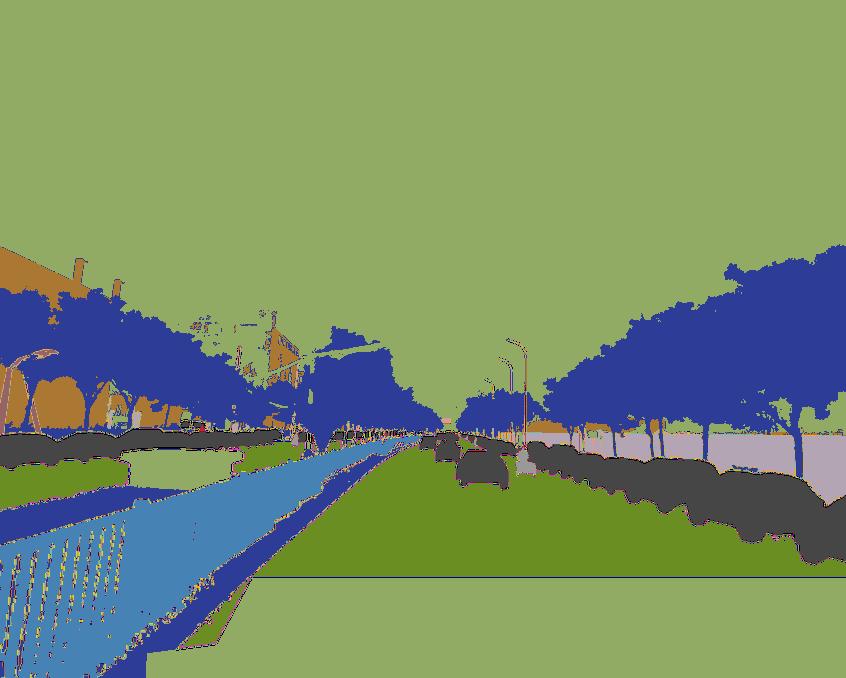}}
        \vspace{-0.3cm}
        \label{Fig.URs_example}
        \caption{One example.}
    \end{minipage}%
    \begin{minipage}{0.85\linewidth}
    \vspace{-0.3cm}
    \captionof{table}{Output comparison among CE, VRs, URs (L=2) and URs (L=5) for specific channels}
    \centering
    \begin{tabular}{|c|@{}c@{}@{}c@{}@{}c@{}@{}c@{}|}
    \hline
    \multirow{1}{*}{} & \multicolumn{1}{c|}{\multirow{1}{*}{CE}} & \multicolumn{1}{c|}{VRs} & \multicolumn{1}{c|}{URs (L=2)} & URs (L=5) \\ \hline
   \rotatebox[origin=c]{90}{\parbox[c]{2.0cm}{\small\textbf{~~Chan1(Car)}}}  &
   \parbox[c]{0.21\linewidth}{\centering\includegraphics[width=0.99\linewidth, height=0.63\linewidth]{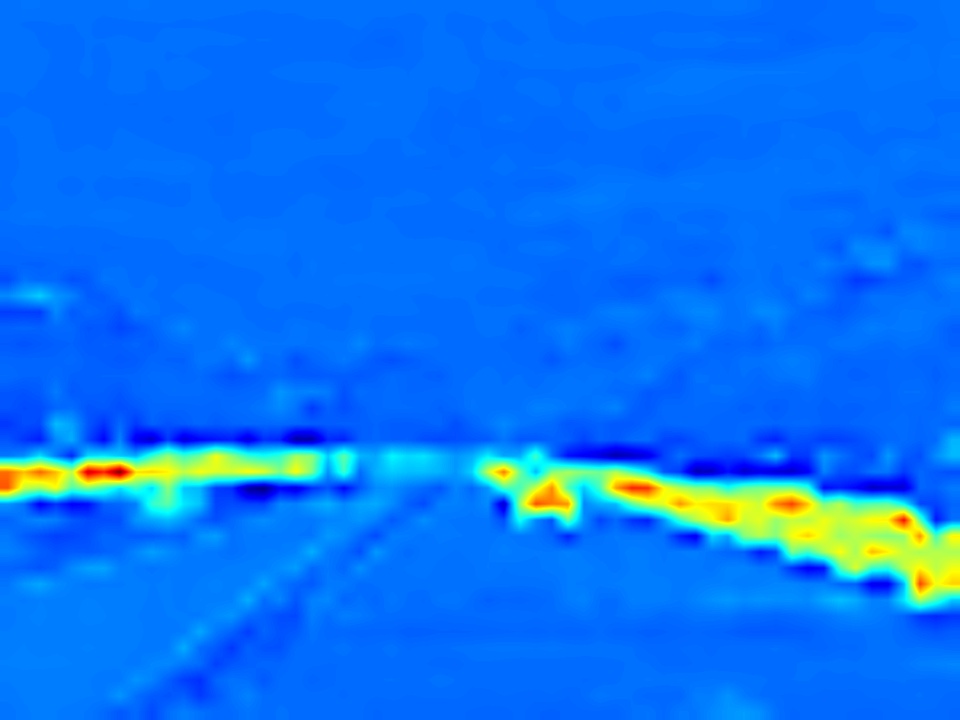}} &
   \parbox[c]{0.21\linewidth}{\centering\includegraphics[width=0.99\linewidth, height=0.65\linewidth]{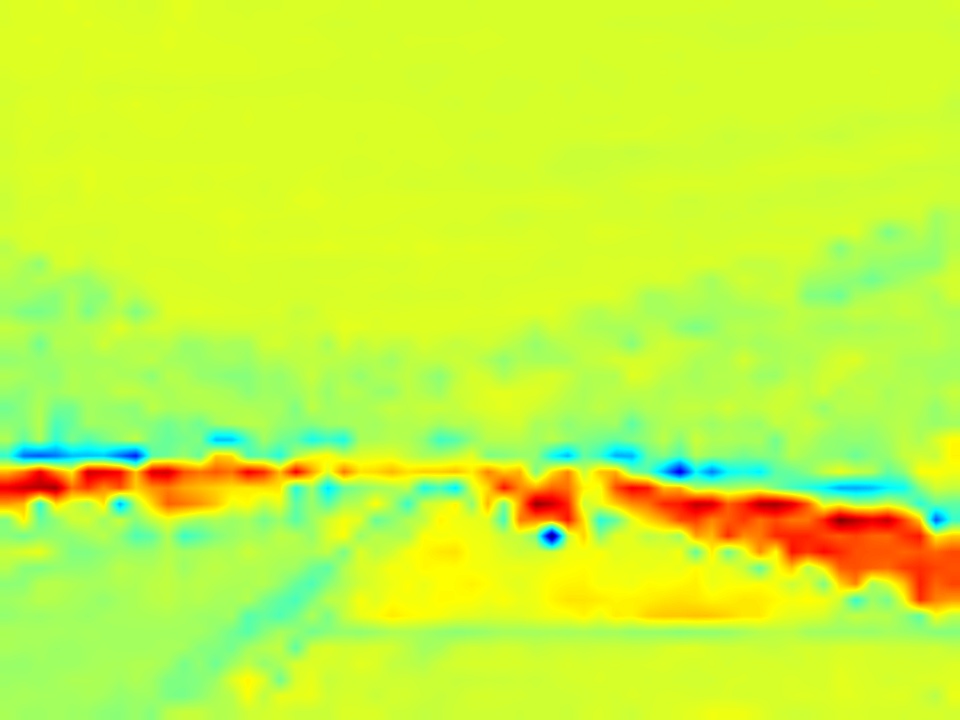}} &
   \parbox[c]{0.21\linewidth}{\centering\includegraphics[width=0.99\linewidth, height=0.63\linewidth]{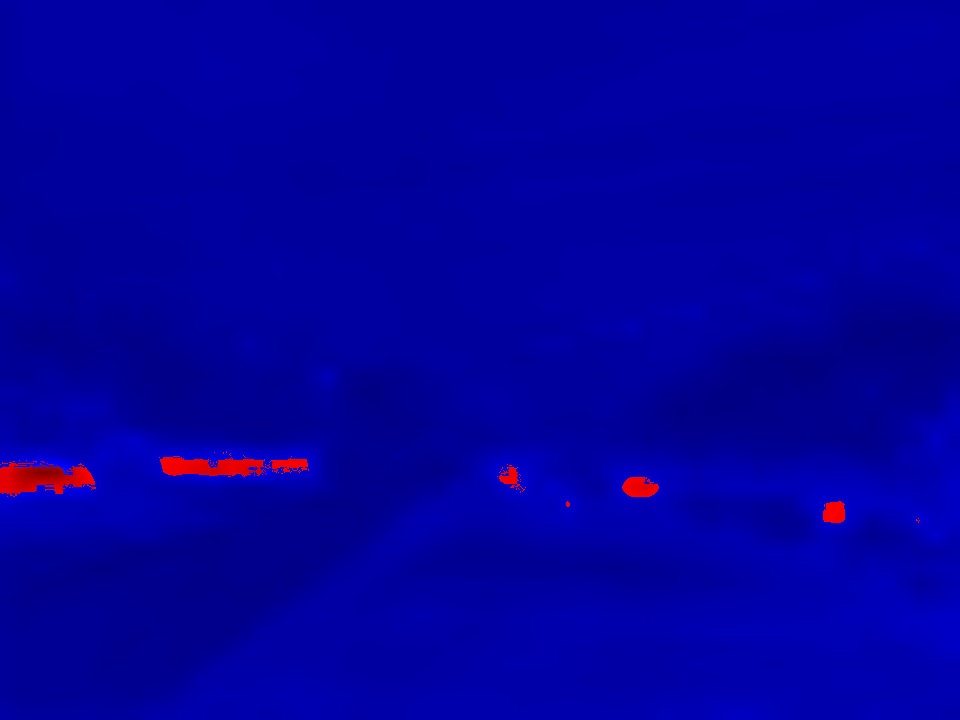}} &
   \parbox[c]{0.21\linewidth}{\centering\includegraphics[width=0.99\linewidth, height=0.65\linewidth]{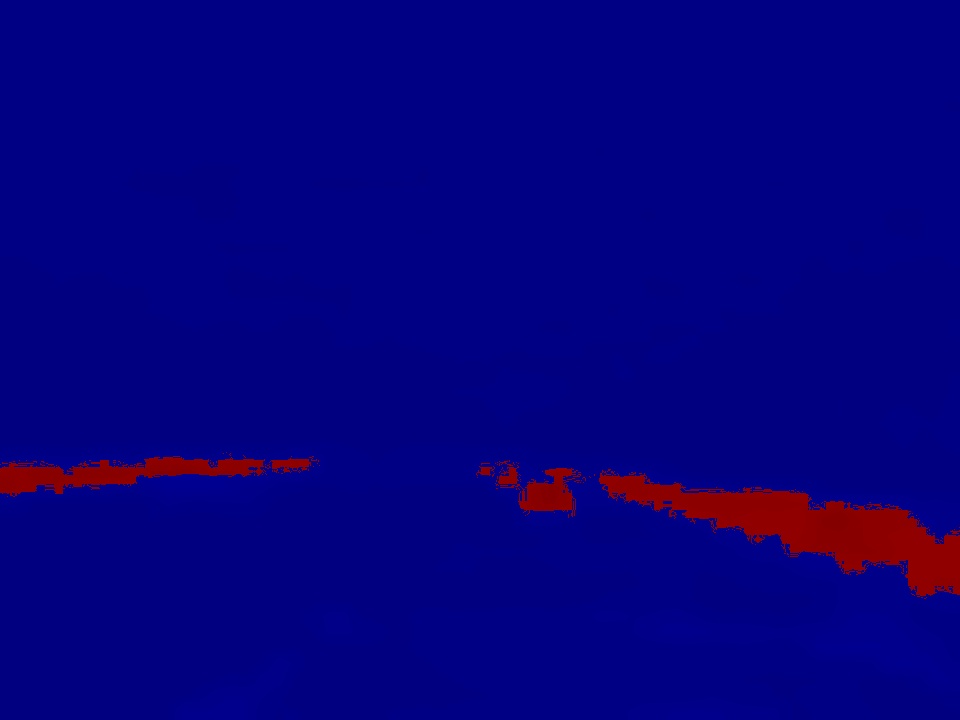}}   \\ \cline{1-1}
   \rotatebox[origin=c]{90}{\parbox[c]{2.0cm}{\small\textbf{Chan13(Fence)}}} &
   \parbox[c]{0.21\linewidth}{\centering\includegraphics[width=0.99\linewidth, height=0.65\linewidth]{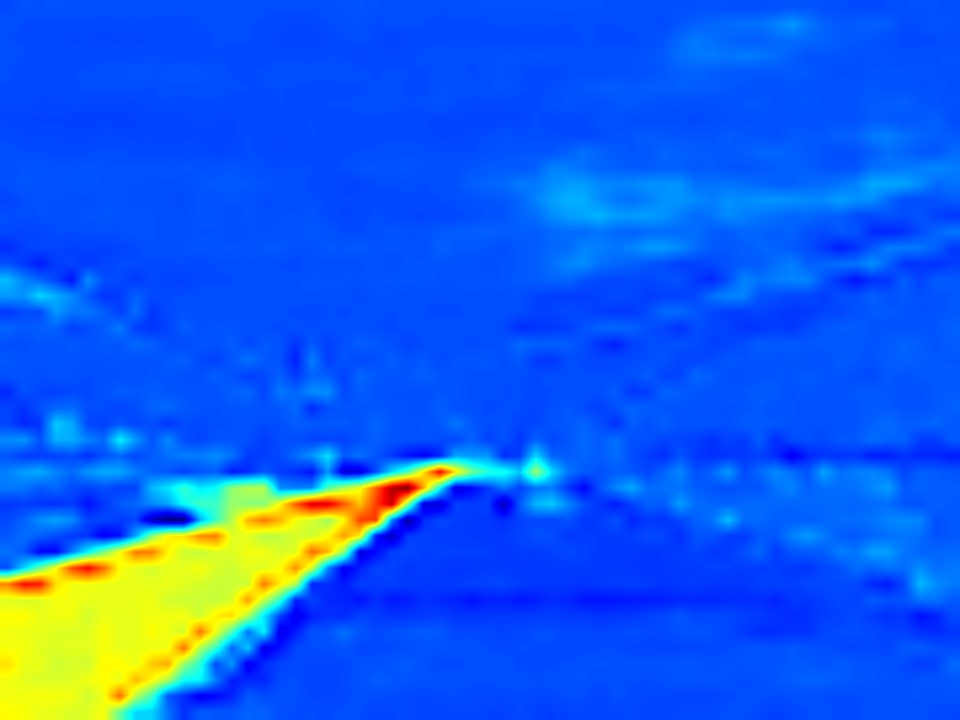}} &
   \parbox[c]{0.21\linewidth}{\centering\includegraphics[width=0.99\linewidth, height=0.63\linewidth]{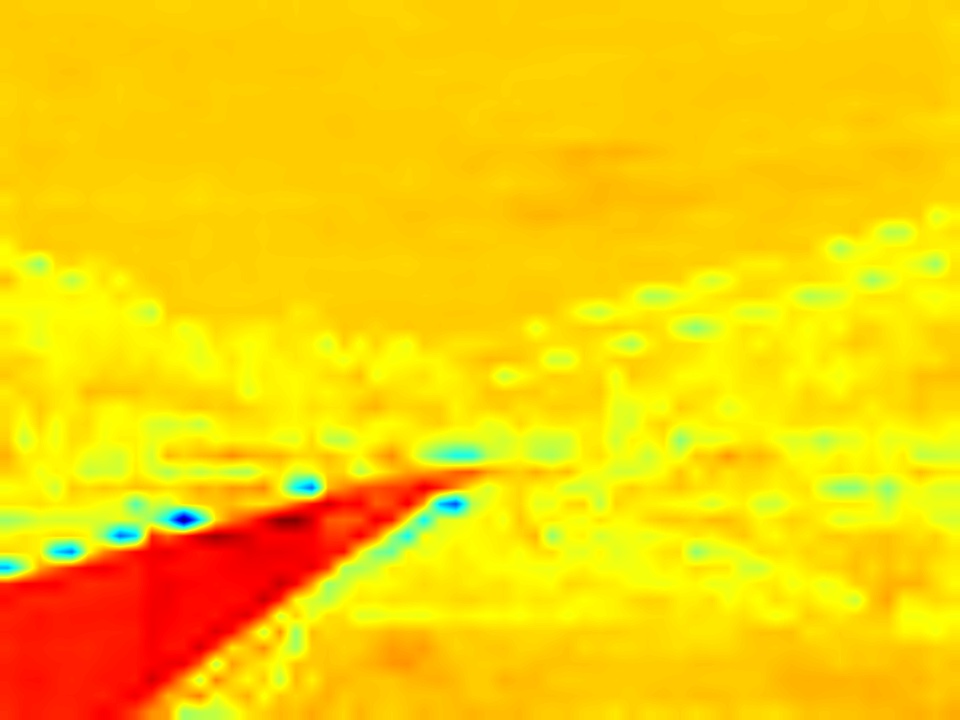}} &
   \parbox[c]{0.21\linewidth}{\centering\includegraphics[width=0.99\linewidth, height=0.65\linewidth]{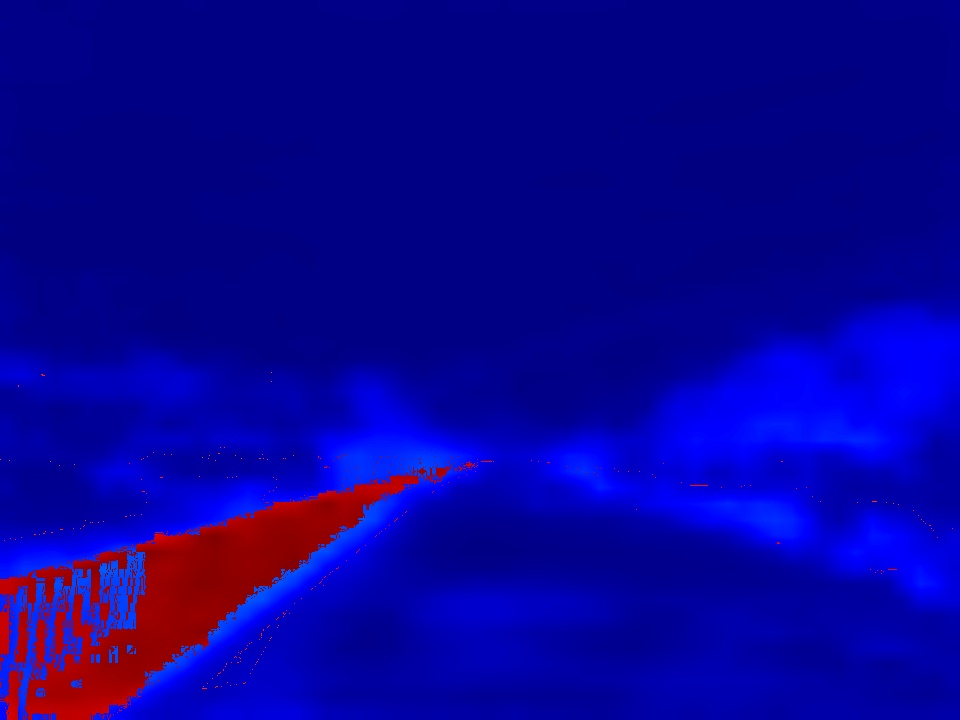}} &
   \parbox[c]{0.21\linewidth}{\centering\includegraphics[width=0.99\linewidth, height=0.63\linewidth]{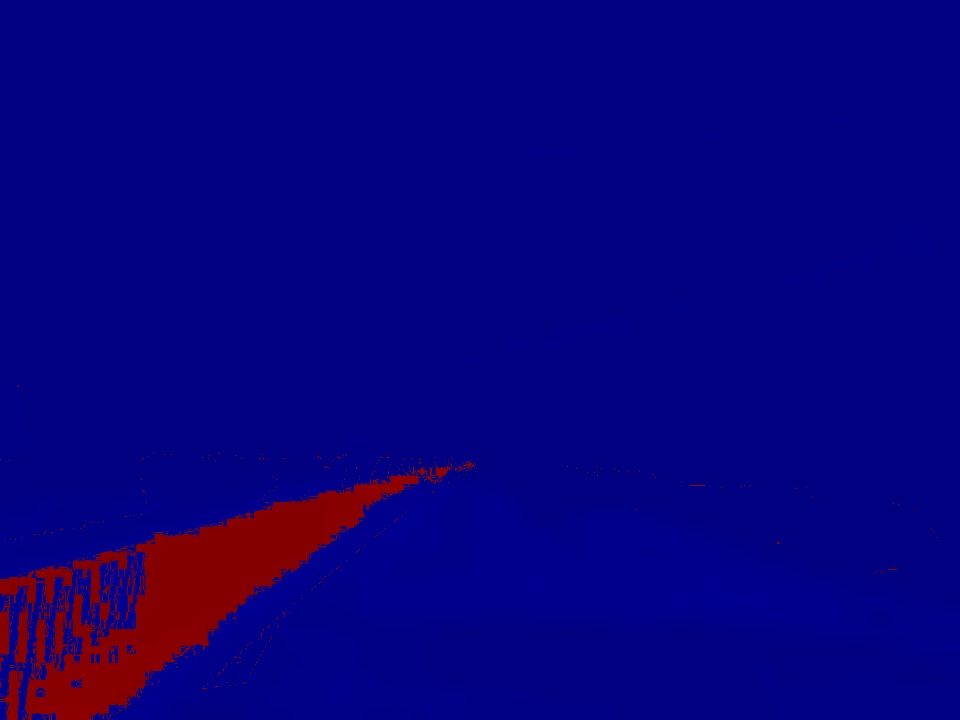}} \\ \hline
   \end{tabular}
   \label{Tab:indepth_exploration_URs}
\end{minipage} 
\vspace{-0.4cm}
\end{figure*}

Subsequently, in order to explore how URs improve the prediction performance, we compare the output of above four cases in a channel-wise manner.  \Cref{Fig.chan_raw} and \Cref{Fig.chan_gt} denote an example image and the corresponding ground truth, respectively. \Cref{Tab:indepth_exploration_URs} showcases the channel-wise output of above four cases, taking the example image as input. Notably, \Cref{Tab:indepth_exploration_URs} only exemplifies the first channel (\ie, Car) and 13-th channel (\ie, Fence) out of 23 channels. From \Cref{Tab:indepth_exploration_URs}, we can find that: \textbf{(I)} For both channels, pure CE can generally supervise the model to learn the most features of Car and Fence, but still mix other channels' features (\ie, noises) with current channel features. For example, the light blue areas actually represent features of other channels and are mixed up together. \textbf{(II)} VRs can effectively draw features of the same class closer while push features of different classes further apart. Specifically, for both channels, the features of Car (or Fence) become more contrastive relative to features of remaining parts. However, VRs' supervised features still exist noises. \textbf{(III)} URs enhance the learned features significantly. Precisely, the features of the Car (or Fence) are nearly identical, while those of the other areas are almost uniform, making the contrast between these two parts much clearer. Notably, as the number of the unfolded layers of URs increases, this trend becomes more pronounced, which can be demonstrated by comparing the outputs of URs (L = 2) and URs (L = 5).

\begin{figure}[tp]
    \centering
    \includegraphics[width=\linewidth, height=0.53\linewidth]{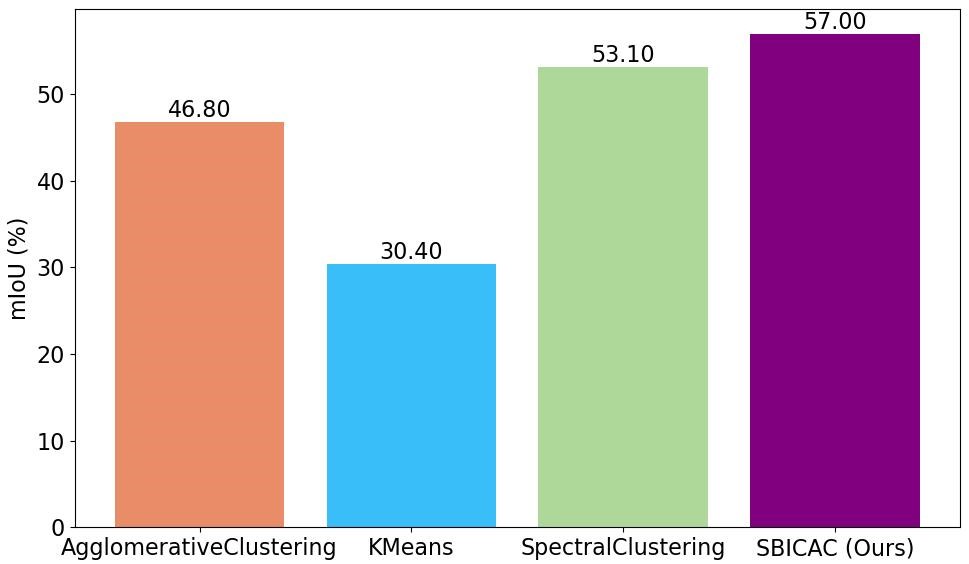}
    \caption{mIoU between ground truth and various clustering algorithm-generated pseudo-labels.}
    \label{Fig.clutering_perf_mIoU}
\vspace{0.2cm}
\end{figure}

\subsection{SBICAC Evaluation} \label{SBICAC_eval}
\Cref{Fig.cluster_comp} presents a qualitative comparison of generated pseudo-labels between the proposed SBICAC and other off-the-shelf clustering algorithms.
To quantitatively evaluate the proposed SBICAC against such baselines, we compare the mIoU between the ground truth and the respective masks generated by such baselines. However, in general, the semantic class IDs within the generated masks by those clustering algorithms are not consistent with those within the ground truth. To overcome the mismatch of semantic class IDs for measuring the aforementioned mIoU, an approach is structured as follows:
\begin{enumerate}
    \item  Enumeration of ID Mappings: The approach initiates by enumerating all possible mappings of cluster IDs between ground truth and the generated masks. This ensures that every potential permutation of cluster assignments is considered.
    \item  Computation of IoU: For each hypothesized mapping, the IoU metric is calculated. 
    \item  Aggregation of IoU: To derive a correspondence, the highest IoU values for each class are identified, and their average is computed, resulting in the mIoU. This mIoU measures the overall alignment between the clustering output and the ground truth.
\end{enumerate}

\Cref{Fig.clutering_perf_mIoU} shows a quantitative comparison of four clustering algorithms. The proposed SBICAC algorithm performs the best with an mIoU of 57.00\%. It is followed by SpectralClustering at 53.10\%, AgglomerativeClustering at 46.80\%, and KMeans with the lowest mIoU of 30.40\%. This demonstrates that SBICAC outperforms the other three clustering algorithms. 

On the other hand, we utilize both the ground truth and SBICAC-generated pseudo-labels to train the proposed AdvImmu. \Cref{Tab:training_comp_with_SBICAC_labels} compares the inference performance of models trained by using ground truth and SBICAC-generated pseudo-labels, respectively. We can find that the models trained with SBICAC-generated labels can achieve performance comparable to, or even surpassing, the models trained with the ground truth. This superior performance is attributed to the fact that the generated labels are coarser than the ground truth (\ie, easier to classify). Thus, we can conclude that the proposed SAM+SBICAC can effectively generate pseudo-labels for the proposed AdvImmu method.

\begin{table*}[tp]
\caption{The performance comparison of models trained using ground truth and SBICAC-generated pseudo-labels}
\setlength{\tabcolsep}{3.9pt}
\begin{tabularx}{\linewidth}{cccccccccccc}
\hline
\multirow{2}{*}{Method} & \multicolumn{2}{c}{mIoU}  &  & \multicolumn{2}{c}{mPre}  &  & \multicolumn{2}{c}{mRec}  &  & \multicolumn{2}{c}{mF1}   \\ \cline{2-3} \cline{5-6} \cline{8-9} \cline{11-12} 
                        & Ground Truth & SAM+SBICAC &  & Ground Truth & SAM+SBICAC &  & Ground Truth & SAM+SBICAC &  & Ground Truth & SAM+SBICAC \\ \hline
DeepLabv3+ & 26.58$\pm$0.49          &24.62$\pm$0.06          & &30.23$\pm$0.89          &26.88$\pm$0.05 & &31.14$\pm$0.57          &32.17$\pm$0.13 &  &32.32$\pm$0.59  &28.98$\pm$0.07        \\
BiSeNetV2  & 22.92$\pm$0.86          &23.02$\pm$0.25          & &27.85$\pm$1.20          &24.93$\pm$0.19 & &27.10$\pm$1.11          &30.20$\pm$0.36 &  &27.12$\pm$1.57  &26.95$\pm$0.26        \\
SegNet     & 21.01$\pm$0.51          &24.51$\pm$1.15          & &24.80$\pm$1.11          &27.89$\pm$1.71 & &24.96$\pm$0.54          &31.69$\pm$1.22 &  &24.60$\pm$0.70  &29.23$\pm$1.48        \\
AttaNet    & 20.59$\pm$0.18          &25.09$\pm$0.23          & &25.88$\pm$0.31          &28.66$\pm$0.19 & &25.38$\pm$0.22          &32.46$\pm$0.27 &  &25.55$\pm$0.19  &29.99$\pm$0.24        \\
BASeg      & 20.14$\pm$0.18          &32.10$\pm$0.15          & &31.72$\pm$0.29          &43.05$\pm$0.42 & &25.26$\pm$0.20          &39.46$\pm$0.21 &  &25.75$\pm$0.31  &40.33$\pm$0.16        \\
HRDA       & 21.55$\pm$0.27          &16.86$\pm$0.69          & &34.13$\pm$0.31          &29.40$\pm$1.40 & &26.09$\pm$0.34          &23.12$\pm$0.66 &  &26.40$\pm$0.31  &21.88$\pm$0.70        \\
SeaFormer  & 20.34$\pm$0.16          &22.04$\pm$0.12          & &25.33$\pm$0.21          &23.85$\pm$0.09 & &25.03$\pm$0.13          &28.91$\pm$0.22 &  &24.26$\pm$0.16  &25.84$\pm$0.11        \\
TopFormer  & 20.41$\pm$0.00          &23.03$\pm$0.30          & &25.19$\pm$0.45          &24.86$\pm$0.40 & &25.20$\pm$0.22          &30.33$\pm$0.40 &  &24.28$\pm$0.20  &26.95$\pm$0.44        \\
AdvImmu    & \textbf{59.35$\pm$0.57} &\textbf{63.97$\pm$2.20} & &\textbf{81.38$\pm$2.22} &\textbf{86.60$\pm$0.99} & &\textbf{60.76$\pm$0.49} &\textbf{65.63$\pm$1.77} &  &\textbf{65.28$\pm$0.55} &\textbf{70.78$\pm$1.57} \\ \hline
\end{tabularx}
\label{Tab:training_comp_with_SBICAC_labels}
\end{table*}

\section{Conclusion}
\label{conclusion}
AD street scene semantic understanding in various adverse weather conditions is challenging. Widely used domain adaption solution suffered from either being dependent on tough-to-collect reference images or significant performance drop in the scenario of mixup of multiple adverse weather conditions. To tackle such weaknesses of domain adaption, this work proposed AdvImmu to enhance the model generalization across multiple adverse weather conditions by considering local temporal correlation. It process\rev{es} frame sequences using InsU, IntU and DU to capture key information, consistent background and dynamic changes, and then shuffl\rev{es} these processed features to avoid overfitting to specific temporal patterns and to improve generalization. To overcome \rev{the difficult-to-obtain frame annotations}, we integrated the SAM and SBICAC clustering algorithm to support training. Extensive experiments showed that AdvImmu outperform\rev{s} existing methods significantly in adopted metrics. The future works include following aspects: \textbf{(I)} We plan to extent the research to other tasks in the AD scenario, such as object detection, \rev{and} end-to-end driving. \textbf{(II)} We also plan to extend the research to multi-vehicle system by incorporating federated learning paradigm.

\end{document}